\documentclass{article}

\usepackage{microtype}
\usepackage{graphicx}
\usepackage{subfigure}
\usepackage{booktabs} 
\usepackage{multirow}
\usepackage{colortbl}
\usepackage{hyperref}

\usepackage[accepted]{icml2021}

\icmltitlerunning{Progressive-Scale Blackbox Attack via Projective Gradient Estimation}

\usepackage{amsmath,mathtools,amssymb,amsthm,dsfont,cleveref,footmisc,booktabs,afterpage,diagbox}

\usepackage{amsthm,amsmath,bm}
\usepackage{dsfont}
\usepackage{nicefrac}

\def\1{\bm{1}}

\def\mI{{\bm{I}}}

\def\mM{{\bm{M}}}

\def\mR{{\bm{R}}}
\def\mS{{\bm{S}}}
\def\mT{{\bm{T}}}
\def\mU{{\bm{U}}}
\def\mV{{\bm{V}}}
\def\mW{{\bm{W}}}

\def\mSigma{{\bm{\Sigma}}}

\DeclareMathAlphabet{\mathsfit}{\encodingdefault}{\sfdefault}{m}{sl}
\SetMathAlphabet{\mathsfit}{bold}{\encodingdefault}{\sfdefault}{bx}{n}

\def\gF{{\mathcal{F}}}

\def\gO{{\mathcal{O}}}

\def\sR{{\mathbb{R}}}

\newcommand{\E}{\mathbb{E}}

\newcommand{\R}{\mathbb{R}}

\newcommand{\Var}{\mathrm{Var}}

\DeclareMathOperator*{\argmax}{arg\,max}

\def\rank{\textup{rank}}

\def\T{{\scriptscriptstyle\mathsf{T}}}

\def\diag{\mathrm{diag}}
\def\proj{\mathrm{proj}}

\def\dif{\mathrm{d}}
\def\sgn{\textup{sgn}}
\def\dom{\textup{dom}}
\def\im{\textup{im}}
\def\proj{\textup{proj}}
\def\Span{\textup{span}}
\def\Unif{\textup{Unif}}
\def\Beta{\textup{Beta}}
\newcommand{\half}{\nicefrac{1}{2}}
\def\opt{\textup{opt}}

\newtheorem{lemma}{Lemma}[section]

\newtheorem{theorem}{Theorem}

\newtheorem{innercustomthm}{Theorem}
\newenvironment{customthm}[1]
  {\renewcommand\theinnercustomthm{#1}\innercustomthm}
  {\endinnercustomthm}

\newtheorem{innercustomlem}{Lemma}
\newenvironment{customlem}[1]
  {\renewcommand\theinnercustomlem{#1}\innercustomlem}
  {\endinnercustomlem}

\theoremstyle{definition}
\newtheorem{definition}{Definition}

\Crefname{adxproposition}{Proposition}{Propositions}
\Crefname{adxcorollary}{Corollary}{Corollaries}
\Crefname{adxdefinition}{Definition}{Definitions}
\Crefname{adxtheorem}{Theorem}{Theorems}

\theoremstyle{remark}
\newtheorem*{remark}{Remark}

\usepackage{scalerel,stackengine}
\stackMath
\renewcommand\hat[1]{%
\savestack{\tmpbox}{\stretchto{%
  \scaleto{%
    \scalerel*[\widthof{\ensuremath{#1}}]{\kern.1pt\mathchar"0362\kern.1pt}%
    {\rule{0ex}{\textheight}}
  }{\textheight}%
}{2.4ex}}%
\stackon[-6.9pt]{#1}{\tmpbox}%
}

\usepackage{xspace}
\usepackage{enumitem}
\usepackage{version}

\newcommand{\ourapproach}{PSBA\xspace}

\newcommand{\ourvariant}{-PGAN\xspace}

\newcommand{\xt}{x_t}
\newcommand{\f}{\bm{f}}
\newcommand{\tnablaS}{\widetilde{\nabla S}}

\definecolor{eqbrown}{rgb}{0.561,0.502,0.451}
\definecolor{eqpink}{rgb}{0.816,0.192,0.847}
\definecolor{eqblue}{rgb}{0.392,0.314,0.784}
\definecolor{eqgreen}{rgb}{0.227,0.494,0.278}

\crefname{equation}{Eq.}{Eqs.}
\crefname{figure}{Fig.}{Figs.}

\newcount\Final  
\definecolor{darkgreen}{rgb}{0,0.5,0}
\definecolor{darkblue}{rgb}{0,0,0.5}
\definecolor{purple}{rgb}{1,0,1}
\definecolor{gray}{rgb}{0.5,0.5,0.5}
\newcommand{\kibitz}[2]{\ifnum\Final=0\textcolor{#1}{#2}\fi}

\ifnum\Final=0\else\excludeversion{old}\fi
\ifnum\Final=0\newenvironment{new}{\color{blue}}{}\else\fi

\begin{document}

\twocolumn[
\icmltitle{Progressive-Scale Boundary Blackbox Attack via Projective Gradient Estimation}



\icmlsetsymbol{equal}{*}

\begin{icmlauthorlist}
\icmlauthor{Jiawei Zhang}{equal,zju}
\icmlauthor{Linyi Li}{equal,uiuc}
\icmlauthor{Huichen Li}{uiuc}
\icmlauthor{Xiaolu Zhang}{antfin}
\icmlauthor{Shuang Yang}{ali}
\icmlauthor{Bo Li}{uiuc}
\end{icmlauthorlist}

\icmlaffiliation{zju}{Zhejiang University, China (work done during remote internship at UIUC)}
\icmlaffiliation{uiuc}{UIUC, USA}
\icmlaffiliation{antfin}{Ant Financial, China}
\icmlaffiliation{ali}{Alibaba Group US, USA}

\icmlcorrespondingauthor{Linyi Li}{linyi2@illinois.edu}
\icmlcorrespondingauthor{Bo Li}{lbo@illinois.edu}

\icmlkeywords{adversarial machine learning, blackbox attack}

\vskip 0.3in
]



\printAffiliationsAndNotice{\icmlEqualContribution} 

\begin{abstract}

   Boundary based blackbox attack has been recognized as practical and effective, given that an attacker only needs to access the final model prediction. However, the query efficiency of it is in general high especially for high dimensional image data.
   In this paper, we show that such efficiency highly depends on the scale at which the attack is applied, 
    and attacking at the optimal scale significantly improves the efficiency.
    In particular, we propose a theoretical framework to analyze and show three key characteristics to improve the query efficiency.
    We prove that there exists an \textit{optimal} scale for projective gradient estimation.
    Our framework also explains the satisfactory performance achieved by existing boundary blackbox attacks.
    Based on our theoretical framework, we propose \textbf{P}rogressive-\textbf{S}cale enabled projective \textbf{B}oundary \textbf{A}ttack~(\ourapproach)
   to improve the query efficiency via
    progressive scaling techniques.
    In particular, we employ Progressive-GAN to optimize the scale of projections, which we call \ourapproach\ourvariant. We evaluate our approach on both spatial and frequency scales.
    Extensive experiments on MNIST, CIFAR-10, CelebA, and ImageNet against different models including a real-world face recognition API show that \ourapproach\ourvariant significantly outperforms existing baseline attacks in terms of query efficiency and attack success rate. We also observe relatively stable optimal scales for different models and datasets. The code is publicly available at \url{https://github.com/AI-secure/PSBA}.
\end{abstract}

\section{Introduction}

    \label{sec:intro}
    
    
    Blackbox attacks against machine learning (ML) models have raised great concerns recently given the wide application of ML~\cite{krizhevsky2009learning,he2016deep,vaswani2017attention}.
     Among these blackbox attacks, boundary blackbox attack~\cite{brendel2018decision,chen2020hopskipjumpattack} has shown to be effective.  However, it usually requires a large number of queries against the target model given the high-dimensional search space. Recent research shows that it is possible to sample the queries from a lower dimensional sampling space first and project them back to the original space for gradient estimation to reduce the  query complexity~\cite{Li_2020_CVPR,li2020nolinear}. These observations raise additional questions: \textit{What is the ``optimal'' projection subspace that we can sample from? How effective such projective attack would be? What is the query complexity for such projective gradient estimation approach?} To answer these questions, in this paper we analyze the general Projective Boundary blackbox Attack framework~(PBA), 
    for which only the decision boundary information (i.e. label) is revealed, and propose Progressive-Scale enabled projective Boundary blackbox Attack~(\ourapproach) to gradually search for the optimal sampling space, which can boost the query efficiency of PBA both theoretically and empirically.
    
\begin{figure}[t]
        \centering
        \includegraphics[width=\linewidth]{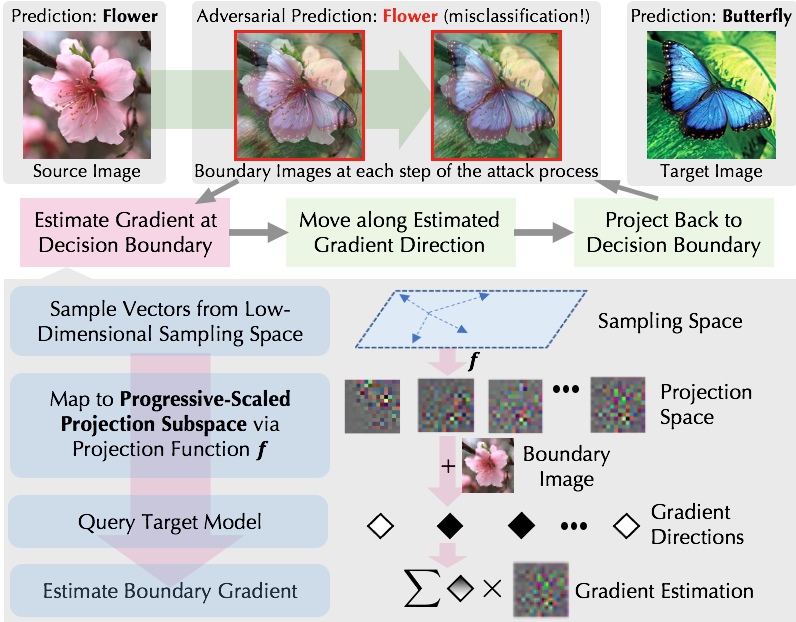}
        \vspace{-7mm}
        \caption{\small An overview of the progressive-scale boundary blackbox attack (\ourapproach) via projective gradient estimation.}
        \label{fig:theory}
    \end{figure}

    The overall pipeline of \ourapproach and the corresponding analysis framework are shown in \Cref{fig:theory}, where an attacker starts with a \textit{source image} and manipulate it to be ``visually close" with a \textit{target image} while preserving its label and therefore fool a ML model via \textit{progressive-scale} based projective gradient estimation. In particular, an attacker first conducts binary search to find a boundary point based on the source image; then samples several perturbation vectors from a low-dimensional sampling space and project them back to the \textit{projection subspace} via a {projection function} $\f$ to estimate the gradient.
    Finally, the attacker will move along the estimated gradient direction to construct the adversarial instance.
    The main goal of \ourapproach is to search for the optimal \textit{projection subspace} for attack efficiency and effectiveness purpose, and we explore such optimal projection subspace both theoretically and empirically.

    
    Theoretically, we develop a \textit{general framework} to analyze the query efficiency of gradient estimation for PBA.
    With the framework, we 
    1)~provide the expectation and concentration bounds for  cosine similarity between the estimated and  true gradients based on nonlinear projection functions, while  previous work only considers identical projection~\cite{chen2020hopskipjumpattack} or orthogonal sampling~\cite{Li_2020_CVPR,li2020nolinear};
    2)~discover several key characteristics that contribute to tighter gradient estimation, including small dimensionality of the projection subspace, large projected length of the true gradient onto the projection subspace, and high sensitivity on the projected true gradient direction;
    3)~analyze the trade-off between small dimensionality  and large projected length, and prove the existence of an optimal subspace dimensionality, i.e., an optimal scale;
    4)~prove that choosing a subspace with large projected length of true gradient as the projection subspace can improve the query efficiency of gradient estimation.
    This framework not only provides theoretical justification for existing PBAs~\cite{brendel2018decision,chen2020hopskipjumpattack,Li_2020_CVPR,li2020nolinear},
    but also enables the design of more efficient blackbox attacks, where the proposed \ourapproach is an example.
    
    Inspired by our theoretical analysis, we design PSBA to progressively search for the optimal scale to perform  projective gradient estimation for boundary blackbox attacks.
    We first consider the spatial scale (i.e., resolution of images), and apply PSBA to search over different spatial scales. 
    %
    We then extend \ourapproach to the frequency scale~(i.e., threshold of low-pass filter) and spectrum scale~(i.e., dimensionality of PCA).
    In particular, as a demonstration, we instantiate \ourapproach with Progressive-GAN~(PGAN), and we conduct extensive experiments to 1)~justify our theoretical analysis on key characteristics that contribute to tighter gradient estimation and 2)~show that  \ourapproach\ourvariant \emph{consistently and significantly} outperforms existing boundary attacks such as HSJA~\cite{chen2020hopskipjumpattack}, QEBA~\cite{Li_2020_CVPR}, NonLinear-BA~\cite{li2020nolinear}, EA~\cite{dong2019efficient}, and Sign-OPT~\cite{cheng2019sign} on various datasets including MNIST~\cite{lecun1998gradient}, CIFAR-10~\cite{krizhevsky2009learning}, CelebA~\cite{liu2015faceattributes}, and ImageNet~\cite{imagenet_cvpr09} against different models including a real-world face recognition API. 
    
{\bf \underline{Technical Contributions.}}
In this paper, we take the \textit{first} step
towards exploring the impacts of different projection scales of projection space in boundary blackbox attacks.
We make contributions on both theoretical
and empirical fronts.   
\begin{itemize}[leftmargin=*,itemsep=-0.5mm]
        \vspace{-0.8em}
        \item 
       We propose the first theoretical framework to analyze boundary blackbox attacks with general projection functions.
        Using this framework, we derive tight expectation and concentration bounds for the cosine similarity between estimated and true gradients.
        \item 
       We characterize the key characteristics and trade-offs for a good projective gradient estimator.
        In particular, we theoretically prove the existence of the optimal scale of the projection space.
        \item 
       We propose Progressive-Scale based projective Boundary Attack~(\ourapproach) via progressively searching for the optimal scale in a self-adaptive way under spatial, frequency, and spectrum scales.
        \item 
       We instantiate \ourapproach by PGAN for empirical evaluation.
        Extensive experiments show that \ourapproach\ourvariant outperforms the state-of-the-art boundary attacks on MNIST, CIFAR-10, CelebA, and ImageNet against different blackbox models and a real-world face recognition API.
    \end{itemize}
    
    \noindent\textbf{Related Work.}
    Adversarial attacks against ML have been conducted to explore the vulnerabilities of learning models and therefore improve their robustness ~\cite{szegedy2013intriguing,kurakin2016adversarial}.
    Most existing attacks~\cite{goodfellow2014explaining,madry2018towards} assume the whitebox setting, where the attacker has full access to the target model including its structure and weights.
    However, in practice, such as commercial face recognition APIs~\cite{facepp_main}, we cannot access the full model.
    Thus, several blackbox attacks have been proposed, which mainly fall into three categories: transfer-based, query-based, and hybrid blackbox attacks:
    1)~The \emph{transfer-based attacks} usually train a surrogate model, attack the surrogate model using whitebox attacks, and exploit the adversarial transferability~\cite{papernot2016transferability,tramer2017space} to use the generated adversarial examples to attack the blackbox model.
    2)~The \emph{query-based attacks} can be further divided into score-based and decision-based.
    The \emph{score-based attacks} assume that we know the confidence score of the target model's prediction and therefore estimate the gradient based on the prediction scores~\cite{chen2017zoo,bhagoji2018practical,tu2019autozoom,ilyas2018black,cheng2019improving,chen2017zoo,li2020projection}.
    The \emph{decision-based attacks} assume that we only know the final decision itself which is more practical, such as the boundary attack~\cite{brendel2018decision},  EA~\cite{dong2019efficient} and Sign-OPT~\cite{cheng2019sign}. 
    HSJA~\cite{chen2020hopskipjumpattack} extends the boundary attack by adopting a sampling-based gradient estimation component to guide the search direction.
    Later on, QEBA~\cite{Li_2020_CVPR} and NonLinear-BA~\cite{li2020nolinear} are proposed to use a projection function to sample from a lower dimensional space to improve the sampling efficiency.
    Our work focuses on decision-based attacks.
    Specifically, our work systematically studies the projective gradient estimation for boundary attacks and reveals that \textit{progressive-scale enabled projection} could improve the query efficiency both theoretically and empirically.
    3)~The \emph{hybrid attacks} usually train one or multiple surrogate models and leverage the gradient information~\cite{guo2019subspace,tashiro2020diversity} or adversarial examples~\cite{suya2020hybrid} to guide the generation of queries for the target model.
    
    \textit{Progressive scaling} has long been an effective methodology for different tasks, such as pyramidal-structured objection detection~\cite{lin2017feature,zhang2020cascaded}, high-resolution generative neural networks~\cite{progressive_gan,zhang2019progressive,wu2020cascade}, and deep feature extraction~\cite{cai2016unified,ma2020mdfn}.
    In this work, we aim to explore whether it is possible to progressively conduct queries from different subspaces (e.g., spatial and frequency subspaces) against a blackbox machine learning model to perform query efficient blackbox attacks.
    

    
\section{Preliminaries}
    \label{sec:prelim}
    In this section, we introduce the related notations for our projective gradient estimation.
    Let $[n]$ denote the set $\{1,2,\dots,n\}$.
    For arbitrary two vectors $a$ and $b$, let $\langle a,b \rangle$ and $\cos\langle a,b \rangle$ denote their dot product and cosine similarity respectively.
    For a matrix $\mW \in \sR^{n\times m}$ and a vector $v\in\sR^n$, we denote its projection on $\Span(\mW)$ by $\proj_{\mW} v$.\footnote{From linear algebra, $\proj_{\mW} v = \mW ( \mW^\T \mW)^{-1} \mW^\T v$.}
    Without loss of generality, we focus on the adversarial attack on an image classifier $G: \sR^n \to \sR^C$ where $C$ is the number of classes.
    For given $x\in\sR^n$,  $G$ predicts the label with highest confidence: $\argmax_{i\in [C]} G(x)_i$.
    As for the \emph{threat model}, we consider the practical setting where only the \emph{decision} of the classification model $G$ is accessible for attackers.

    Following the literature~\cite{chen2020hopskipjumpattack,Li_2020_CVPR}, we define difference function $S(\cdot)$ and sign function $\phi(\cdot)$.
    \begin{definition}
        \label{def:S-and-phi}
        Let label $y_0$ be model $G$'s prediction for input $x^*$.
        For the targeted attack, the adversarial target is $y'\in [C]$.
        Define the difference function $S_{x^*}: \sR^n \to \sR$ as below:
        \begin{equation*}
            \small
            S_{x^*}(x) := \left\{
            \begin{aligned}
                & \max_{y \in [C]: y \neq y_0} G(x)_y - G(x)_{y_0}, & \text{(untargeted attack)} \\
                & G(x)_{y'} - \max_{y \in [C]: y \neq y'} G(x)_y. & \text{(targeted attack)} \\
            \end{aligned}
            \right.
        \end{equation*}
        The function $\phi_{x^*}: \sR^n \to \{\pm1\}$ is the sign function of $S_{x^*}$:
        \begin{equation*}
            \small
            \phi_{x^*}(x) := \left\{
            \begin{aligned}
                & +1 & \text{if } S_{x^*}(x) > 0, \\
                & -1 & \text{otherwise}.
            \end{aligned}
            \right.
        \end{equation*}
    \end{definition}
    When there is no ambiguity, we may abbreviate them as $S(x)$ and $\phi(x)$ respectively.
    For a target image $x^*$, the attacker crafts an image $x$, ensuring that the difference function $S_{x^*}(x) \ge 0$ to perform a success attack while minimizing the distance $\|x - x^*\|_2$.
    
    We call $x$ a \emph{boundary point} if $S_{x^*}(x) = 0$.
    We assume that $S_{x^*}$ is $\beta_S$-smooth.
    Formally, for any $x, z \in \sR^n$\footnote{
        \label{note1}
        For any non-differentiable point of $S_{x^*}$ or $\f$, the assumption generalizes to any Clarke's generalized gradient~\cite{clarke2008nonsmooth}. 
        For simplicity, we assume $S_{x^*}$ and $\f$ are differentiable hereinafter.
        The $\|\cdot\|_2$ operator stands for the $\ell_2$ norm for vector $\nabla S$ and the spectral norm for Jacobian matrix $\nabla \f$.
    },
    \begin{equation}
        \small
        \frac{\|\nabla S_{x^*}(x) - \nabla S_{x^*}(z)\|_2}{\|x - z\|_2} \le \beta_S.
        \label{eq:S-assumption}
    \end{equation}
    Note that we can only query the value of $\phi_{x^*}$ instead of $S_{x^*}$ according to the threat model.
    In boundary attack, we estimate the gradient of $S$ by querying $\phi$.

        We generalize existing gradient estimators for boundary attack in a projective form.
        Suppose we have a projection $\f:\sR^m \to \sR^n$, where $m \le n$.
        This projection can be obtained in various ways, such as PCA~\cite{Li_2020_CVPR}, VAE~\cite{li2020nolinear}, or just an identical projection~\cite{chen2020hopskipjumpattack}.
        Similar to \Cref{eq:S-assumption}, we assume $\f$ is $\beta_{\f}$-smooth\footref{note1}.
        With the projection, given an input $x_t$ that is at or close to the boundary of $S_{x^*}$, with pre-determined step size $\delta_t$, we can estimate gradient $\nabla S_{x^*}(\xt)$ as such:
        \begin{definition}[Boundary Gradient Estimator]
            \label{def:gradient-estimator}
            On $\xt \in \sR^n$, with pre-determined step size $\delta_t$, let $\{u_b\}_{b=1}^B$ be $B$ vectors that are uniformly sampled from the $m$-dimensional unit sphere $S^{m-1}$.
            The gradient $\nabla S_{x^*}(\xt)$ is estimated by
            \vspace{-2mm}
            \begin{equation}
                \small
                \tnablaS_{x^*,\delta_t}(\xt) := \dfrac{1}{B} \sum_{b=1}^B \phi_{x^*}\left(\xt + \Delta \f(\delta_t u_b) \right)
                \Delta \f(\delta_t u_b),
                \vspace{-1mm}
                \label{eq:gradient-estimator}
            \end{equation}
            where $\Delta \f(x) := \f(x) - \f(0)$.
            When there is no ambiguity, we may omit subscript $x^*$~or~$\delta_t$.
            Note that each query of $\phi_{x^*}$ is a query to the blackbox model, so computing $\tnablaS(\xt)$ requires $B$ queries to the blackbox model.
            \vspace{-0.15cm}
        \end{definition}
        As previous work shows, the expected cosine similarity between the estimated and true gradients can be bounded.
        For example, with \emph{identical} projection $\f$~\cite{chen2020hopskipjumpattack},
        \begin{equation}
            \label{eq:hsja-bound}
            \small
            \cos \langle \tnablaS(x_t), \nabla S (\xt) \rangle \ge 1 - \frac{9\beta_S^2 \delta_t^2 n^2}{8\|\nabla S(\xt)\|_2^2}.
        \end{equation}
        
        \begin{definition}[Sampling Space and Projection Space]
            For the given projection $\f: \sR^m \to \sR^n$. We call the domain $\dom(\f) = \sR^m$ the \emph{sampling space}, and the subspace consisting of projected images the \emph{projection subspace}.
            \label{def:sampling-space-proj-space}
        \end{definition}
        
        \begin{remark}
            The definition reflects the sample-and-project process in gradient estimation.
            The projection subspace is a subspace of the original input space $\sR^n$.
            Typically, the step size $\delta_t$ is small, and $\Delta \f(\delta u) \approx \nabla \f(0) \cdot (\delta u)$ by linear approximation.
            Therefore, we can view $\f$ as a non-singular linear projection where $m$ is the dimensionality of the projection subspace. 
        \end{remark}
    
        
        
        
\section{Progressive-Scale Blackbox Attack: \ourapproach}
\begin{figure}[t]
    \centering
    \includegraphics[width=1.03\linewidth]{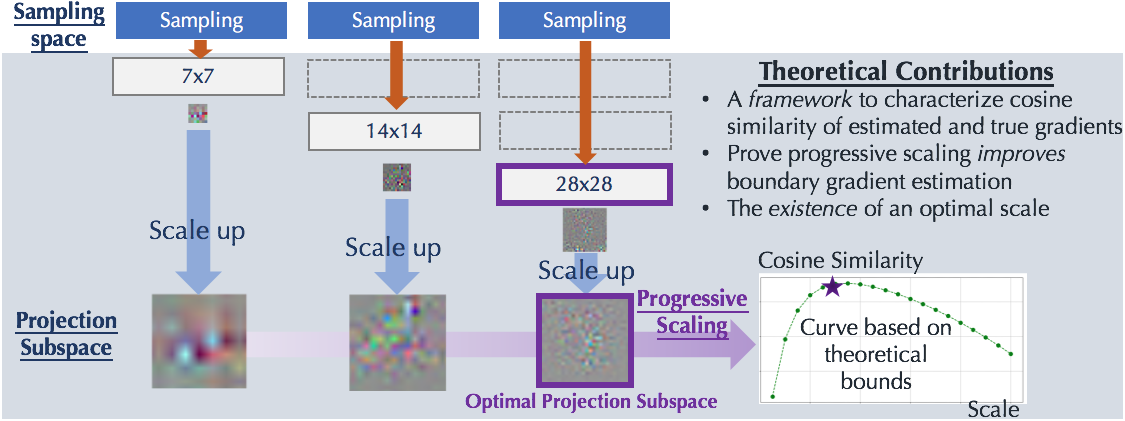}
    \vspace{-7.5mm}
    \caption{\small  \ourapproach with progressive scalling on spatial domain.}
    \label{fig:scale}
    \vspace{-5mm}
\end{figure}
    \label{sec:approach}
    
    In this section we will introduce the general pipeline of proposed \textbf{P}rogressive-\textbf{S}cale projective \textbf{B}oundary \textbf{A}ttack~(\ourapproach), followed by detailed analysis of it in Section~\ref{sec:analysis}.
    On the high level, \ourapproach progressively increases the scale of\textit{ projection subspace} where the perturbation vectors will be chosen from, until it reaches the ``optimal" scale as shown in Figure~\ref{fig:scale}. The scales are sampled from domains such as spatial, frequency, and spectral domain.
    At each scale, \ourapproach is composed of two stages: \textit{training stage} and \textit{attack stage}.
    
    At the \textit{training stage}, we train a generative model $M$  (e.g., GAN), with gradient images obtained from any model.
    The input space of $M$ is a low-dimensional space ($m$).
    For instance, from the spatial domain, we can sample from different resolutions such as $7 \times 7$, $14 \times 14$, and $28 \times 28$. 
   Then we leverage  $M$ to project the sampled vectors back to the original space with interpolation. These projected images form the \textit{projection subspace}.

    At the \textit{attack stage}, we adapt the boundary attack pipeline, and first select a source image $\hat{x_0}$~(drawn from images from the adversarial target class). In each iteration $t$, the attack reduces the distance from current adversarial sample $\xt$ to the target image $x^*$ via three steps:
    (1)~binary search for the boundary point $\xt$ where $S_{x^*}(\xt)=0$ on the line connecting $\hat{x_{t-1}}$ and $x^*$, with pre-determined precision threshold $\theta$;
    (2)~estimate the gradient at $\xt$ using the boundary gradient estimator~(\Cref{def:gradient-estimator}) with step size $\delta_t$;
    (3)~normalize the estimated gradient, and perform a step of gradient ascent to get $\hat \xt$.
    Note that in (1), the binary search for boundary point requires $\gO(\log \nicefrac{1}{\theta})$ queries to the blackbox model, and in (2), the gradient estimation requries $B$ samples and queries to the blackbox model. 
    In \Cref{sec:analysis} we will analyze the relation between $B$ and the quality of estimated gradients in terms of cosine similarity.
    
    We select the optimal scale for projection subspace based on a validation set.
    If with current scale, after $1,000$ queries, the average distance to the target image $x^*$ is smaller than that of the previous scale, we try a new increased scale.
    Otherwise, i.e., the average distance is larger than that of the previous one, we select the previous scale as the optimal scale, and use it as the scale of projection subspace.
    This process can be viewed as climbing to find the maximum point on the curve in \Cref{fig:scale} from left to right. Detailed pseudocode can be found in Appendix~\ref{sec:alg_pseudocode}.
    
    In particular, we mainly consider different scales in the spatial, frequency, and spectrum domains.
    For spatial domain, $M$ on the progressively grown scale (i.e., resolution) can be effectively trained via Progressive-GAN. 
    The frequency scales correspond to thresholds for the low-pass frequency filter, and the spectrum scales correspond to dimensionalities of PCA.
   For frequency and spectrum, $M$ can be trained on full-scale and trimmed to fit in the required lower scale as discussed in Section~\ref{subsec:projection-on-selected-subspace}.
    
\section{Analysis of Gradient Estimation}
\label{sec:analysis}

    In this section, we analyze the similarity between the estimated and true gradients for general projective boundary attack frameworks.
    These attacks all follow \Cref{def:gradient-estimator} to estimate the gradient.
    Our goal is to improve the gradient similarity while reducing the number of queries~($B$).
    
    In \Cref{subsec:general-bound}, we present cosine similarity bounds between the estimated and true gradients for the gradient estimator with general \textbf{nonlinear projection function}, and analyze the key characteristics that improve such gradient estimation.
    In \Cref{subsec:projection-on-selected-subspace}, we analyze the \textbf{bounds of cosine similarity} when the output of projection $\f$ is constrained on selected projection subspace.
    We show how constraining on a representative subspace improves gradient estimation compared with performing gradient estimation on the original space, which explains why \ourapproach outperforms existing methods.
    \subsection{General Cosine Similarity Bounds}
        \label{subsec:general-bound}
        \label{sec:assumption}
        
        Let $\nabla \f(0) \in \sR^{n \times m}$ be the Jacobian matrix of the projection $\f$ at the origin.
        {Throughout the paper},
        we assume that $\nabla \f(0)$ has full-rank since $\nabla\f$ is non-singular in general case.
        We further assume that there exists a column vector $\nabla \f(0)_{:,c}$ that is \emph{aligned with} the projected true gradient $\proj_{\nabla \f(0)} \nabla S(\xt)$, and other column vectors are orthogonal to it.
        Formally, there exists $c \in [m]$, such that $\nabla \f(0)_{:,c} = k\proj_{\nabla \f(0)} \nabla S(\xt)$ with $k \neq 0$,
        and for any $i \neq c$, $\langle \nabla \f(0)_{:,i}, \nabla \f(0)_{:,c} \rangle = 0$.
        This assumption guarantees that the projection model $\f$ produces \emph{no} directional sampling bias for true gradient estimation~(\Cref{lem:a-2}) following the standard setting.
        We remark that since vectors tend to be orthogonal to each other in high-dimensional case~\cite{orthogonalvectors}, this assumption holds with higher confidence in high-dimensional cases.
        In \Cref{subsec:analysis-verify}, we empirically verify this assumption.

        \begin{lemma}[$\nabla \f$ Decomposition]
            \label{lem:nabla-f-decomposition}
            Under the above assumption,
            there exists a singular value decomposition of $\nabla\f(0) = \mU \mSigma \mV^\T$ such that
            \begin{align*}
                \small
                & \mU_{:,1} = \proj_{\nabla \f(0)} \nabla S(\xt) / \|\proj_{\nabla \f(0)}\nabla S(\xt)\|_2 \\
                \text{ or } &
                \mU_{:,1} = -\proj_{\nabla \f(0)}\nabla S(\xt) / \|\proj_{\nabla \f(0)}\nabla S(\xt)\|_2
            \end{align*}
            where $\mU \in \sR^{n\times n}$, $\mV \in \sR^{m\times m}$ are orthogonal matrices;
            $\mSigma = \diag(\alpha_1,\alpha_2,\dots,\alpha_m) \in \sR^{n\times m}_{\ge 0}$ is a rectangular diagonal matrix with $\alpha_1 > 0$.
            Denote $\max_{i\in [m]} \alpha_i$ as $\alpha_{\max}$.
        \end{lemma}
        The proof can be found in \Cref{adxsec:proof}.
        Compared with the standard SVD decomposition, now the first column vector of $\mU$ can be fixed to the normalized projected true gradient vector or its opposite.
        
        \noindent\textbf{Definition of Sensitivity.}
            In \Cref{lem:nabla-f-decomposition}, for projection $\f$, we can view the resulting $\alpha_1$ as the \emph{sensitivity} of the projection model on the (projected) true gradient direction;
            and $\{\alpha_i\}_{i=2}^m$ as the \emph{sensitivity} of the projection model on directions orthogonal to the true-gradient.
            With higher sensitivity on projected true gradient direction~($\alpha_1$) and smaller sensitivity on other orthogonal directions~($\{\alpha_i\}_{i=2}^m$), the gradient estimation becomes better as we will show later.
    \subsubsection*{Main Theorems}
        Here we will present our main theorems for the expectation~(\Cref{thm:approach-boundary-expectation}) and concentration bound~(\Cref{thm:approach-boundary-concentration}) of cosine similarity between the estimated and true gradients.
    
        \begin{theorem}[Expected cosine similarity]
            \label{thm:approach-boundary-expectation}
            The difference function $S$ and the projection $\f$ are as defined before.
            For a point $\xt$ that is $\theta$-close to the boundary, i.e., there exists $\theta' \in [-\theta,\theta]$ such that $S(\xt + \theta' S(\xt) / \|\nabla S(\xt)\|_2) = 0$,
            let estimated gradient $\tnablaS(\xt)$ be as computed by \Cref{def:gradient-estimator} with step size $\delta$ and sampling size $B$.
            Over the randomness of the sampled vectors $\{u_b\}_{i=1}^B$,
            \begin{equation*}
                \small
                \label{eq:thm3-1}
                \resizebox{0.45\textwidth}{!}{
                $
                \begin{aligned}
                    & \cos\langle \E\tnablaS(\xt), \nabla S(\xt) \rangle \ge \dfrac{\|\proj_{\nabla\f(0)} \nabla S(\xt)\|_2}{\|\nabla S(\xt)\|_2} \cdot \\
                    & \left(1 - \dfrac{(m-1)^2 \delta^2}{8\alpha_1^2}
                    \left(
                        \dfrac{\delta \gamma^2}{\alpha_1}
                        +
                        \dfrac{\gamma}{\alpha_1} \sqrt{\dfrac{\sum_{i=2}^m \alpha_i^2}{m-1}}
                        +
                        \dfrac{1.58\beta_{\f}}{\sqrt{m-1}}
                    \right.\right. \\
                    & \hspace{2em} \left.\left.
                        +
                        \dfrac{\gamma\theta}{\alpha_1\delta}
                        \cdot 
                        \dfrac{\|\nabla S(\xt)\|_2}{\|\proj_{\nabla\f(0)} \nabla S(\xt)\|_2}
                    \right)^2\right),
                \end{aligned}
                $
                }
            \end{equation*}
            where
            \begin{equation}
                \small
                \label{eq:thm3-2}
                \gamma := \beta_{\f} +
                \dfrac{\beta_S \left(\max_{i\in [m]} \alpha_i + \half \delta \beta_{\f}\right)^2 + \beta_S \theta^2 / \delta^2 }{\|\proj_{\nabla\f(0)}\nabla S(\xt)\|_2}.
            \end{equation}
        \end{theorem}
        
        \begin{proof}[Proof Sketch.]
            The high-level 
            idea is using Taylor expansion with Lagrange 
            remainder to control both the first-order and higher-order errors, and plug the error terms into the distribution of dot product between $\nabla S(\xt)$ and $\nabla \f(0) \cdot u_b$.
            This dot product follows a linearly transformed Beta distribution~\cite{chen2020hopskipjumpattack}.
            The error 
            terms are 
            separately controlled for the projected gradient direction~(i.e., direction of $\proj_{\nabla \f(0)} \nabla S(\xt)$) and other orthogonal directions.
            Then the controlled directional errors are combined as an $\ell_2$-bounded error vector.
            The complete proof is deferred to \Cref{adxsubsec:proof-thm-approach-boundary-expectation}.
        \end{proof}
        \begin{remark}
            This bound characterizes the expected cosine similarity of the boundary gradient estimator.
            For an identical projection $\f$, if $\xt$ is exactly the boundary point, i.e., $S(\xt) = 0$, from the theorem we get
            $$
                \small
                \cos\langle \E\tnablaS(\xt), \nabla S(\xt) \rangle \ge 1 - \frac{(n-1)^2 \delta^2 \beta_S^2}{2 \|\nabla S(\xt)\|_2^2},
            $$
            where we leverage the fact that $\delta = O(1/n)$ is usually small.
            This bound is of the same order as the previous work~\cite{chen2020hopskipjumpattack} shown in \Cref{eq:hsja-bound}, while containing a tighter constant $1/2$ instead of $9/8$.
            
            Suppose both the difference function $S$ and projection $\f$ are linear, i.e., $\beta_S = \beta_{\f} = 0$, then we have $\cos\langle \E\tnablaS(\xt), \nabla S(\xt) \rangle \ge \frac{\|\proj_{\nabla\f(0)} \nabla S(\xt)\|_2}{\|\nabla S(\xt)\|_2}$.
            From \Cref{lem:cosine-similarity-projected-subspace}, this is the optimal cosine similarity obtainable with the projection $\f$.
            Furthermore, for identical projection, $\cos\langle \E\tnablaS(\xt), \nabla S(\xt) \rangle \ge 1$, which verifies the optimality of our bound over existing work~\cite{Li_2020_CVPR,li2020nolinear}.
        \end{remark}
        
        \begin{theorem}[Concentration of cosine similarity]
            \label{thm:approach-boundary-concentration}
            Under the same setting as \Cref{thm:approach-boundary-expectation}, over the randomness of the sampled vector $\{u_b\}_{i=1}^B$, with probability $1-p$,
            \begin{equation*}
                \small
                \label{eq:thm4-1}
                \resizebox{0.5\textwidth}{!}{
                $
                \begin{aligned}
                    & \cos\langle \tnablaS(\xt), \nabla S(\xt) \rangle \ge \dfrac{\|\proj_{\nabla\f(0)} \nabla S(\xt)\|_2}{\|\nabla S(\xt)\|_2} \cdot \\
                    & \left( 1 - \dfrac{(m-1)^2\delta^2}{8\alpha_1^2} \left( \dfrac{\delta\gamma^2}{\alpha_1} + \dfrac{\gamma}{\alpha_1} \sqrt{\dfrac{\sum_{i=2}^m \alpha_i^2}{m-1}} +
                    \dfrac{1.58\beta_{\f}}{\sqrt{m-1}} \right. \right.\\
                    & \left.\left. +  \dfrac{\gamma\theta}{\alpha_1\delta}
                    \cdot 
                    \dfrac{\|\nabla S(\xt)\|_2}{\|\proj_{\nabla\f(0)} \nabla S(\xt)\|_2}
                    + \dfrac{\frac{1}{\delta} \sqrt{\sum_{i=1}^m \alpha_i^2} \cdot \sqrt{\frac{2}{B} \ln(\frac{m}{p})}}{\sqrt{m-1}} \right)^2\right),
                \end{aligned}
                $
                }
            \end{equation*}
            where $\gamma$ is as defined in \Cref{eq:thm3-2}.
        \end{theorem}
    
        \begin{proof}[Proof Sketch.]
            Each $u_b$ is sampled independently, and on each axis the samples are averaged.
            Therefore, we apply Hoeffding's inequality on each axis, and use the union bound to bound the total $\ell_2$ length of the error vector.
            We propagate this extra error term throughout the proof of \Cref{thm:approach-boundary-expectation}.
            The detail proof is in \Cref{adxsubsec:proof-thm-approach-boundary-concentration}.
            %
        \end{proof}
        
        \begin{remark}
            This is the \textit{first} concentration bound for the boundary gradient estimator to our best knowledge.
            From it we quantitatively learn how increasing the number of queries $B$ increases the precision of the estimator, while the expectation bound cannot reflect it directly.
        \end{remark}
            We note that
            the above two theorems are general---they provide finer-grained bounds for \emph{all} existing boundary blackbox attacks, e.g., \cite{chen2020hopskipjumpattack,li2020nolinear,Li_2020_CVPR}, and the proposed \ourapproach, thus these bounds provide a principled framework to analyze boundary blackbox attacks.
            Next, based on this framework we will (1)~discover key characteristics that affect the query efficiency of gradient estimation; (2)~explain why some existing attacks are more efficient than others; (3)~show why \ourapproach improves upon these attacks.
        
        \label{subsec:result-interpretation}
        \label{subsec:key-characteristics}
        \begin{figure*}[!t]
            \small
            \centering
            \boxed{
            \resizebox{0.95\textwidth}{!}{
            $
            \cos\langle \tnablaS(\xt), \nabla S(\xt) \rangle \ge 
            \dfrac{
                \textcolor{eqbrown}{\bm{\|\proj_{\nabla\f(0)} \nabla S(\xt)\|_2}}
            }{\|\nabla S(\xt)\|_2}
            \cdot
                \left(
                    1 - \gO\left( \textcolor{eqpink}{\bm{m^2}} \cdot 
                        \textcolor{eqgreen}{\dfrac{\bm{\sum_{i=2}^m \alpha_i^2}}{\bm{m-1}}}
                        \left( 
                            \overbrace{
                            \dfrac{\delta^2 \beta_{\f}^2}{
                                \textcolor{eqblue}{\bm{\alpha_1^4}}
                            } + 
                            \dfrac{
                                \textcolor{eqgreen}{\bm{\alpha_{\max}^4}}
                            }{
                                \textcolor{eqblue}{\bm{\alpha_1^4}}
                            } \cdot \dfrac{\delta^2\beta_S^2}{
                            \textcolor{eqbrown}{\bm{\|\proj_{\nabla\f(0)} \nabla S(\xt)\|_2^2}}
                            }
                        }^{\text{expectation}}
                        + 
                        \overbrace{\dfrac{\ln(\textcolor{eqpink}{\bm{m}} / p )}{
                            B\textcolor{eqblue}{\bm{\alpha_1^2}}
                        } }^{\text{sampling error}}
                        \right)
                    \right)
                \right)
            $
            }
            }
            \caption{
                Cosine similarity bound in big-$\gO$ notation.
                The ``expectation'' reflects the expectation bound in \Cref{thm:approach-boundary-expectation}.
                The ``sampling error'' is the additional term in \Cref{thm:approach-boundary-concentration} that makes the bound hold with probability at least $1-p$.
                When projection $\f$ is constrained in selected linear subspace $V$~(see \Cref{subsec:projection-on-selected-subspace}), 
                the bound has the same form with all $\textcolor{eqbrown}{\bm{\|\proj_{\nabla \f(0)} \nabla S(\xt)\|_2^2}}$ replaced by $\textcolor{eqbrown}{\bm{\|\proj_{V} \nabla S(\xt)\|_2^2}}$.
            }
            \label{fig:order-figure}
            \vspace{-2mm}
        \end{figure*}

        \subsubsection*{Key Characteristics of Projective Gradient Estimation}
        Based on these two main theorems, we draw several observations for key characteristics that would help improve the gradient estimator. For simplification, we will leverage the big-$\gO$ notation for the above expectation and concentration bounds~(\Cref{thm:approach-boundary-expectation,thm:approach-boundary-concentration}), as shown in \Cref{fig:order-figure}.
        Compared with the expectation bound, the concentration bound adds a ``sampling error term'' that makes the bound hold with probability at least $1 - p$.
        In \Cref{fig:order-figure}, we label different terms in different colors to represent the key characteristics as optimization goals of a good gradient estimator:
        \begin{enumerate}[label=(\arabic*),leftmargin=*,itemsep=-0.3mm]
            \vspace{-1em}
            \item \emph{Reduce the dimensionality $m$ for the projection subspace}: 
            To increase the cosine similarity, 
            we can reduce the dimensionality $\textcolor{eqpink}{\bm{m}}$.
            \vspace{-0.1cm}
            \item \emph{Increase the projected length of true gradient on the projection subspace}: 
            To increase the cosine similarity,
            we can increase the \textcolor{eqbrown}{brown} term $\textcolor{eqbrown}{\bm{\|\proj_{\nabla\f(0)}\nabla S(\xt)\|_2}}$, which is the projected length of true gradient on the projection subspace of $\f$.
            \vspace{-0.1cm}
            \item \emph{Improve the sensitivity on the true gradient direction}:
            According to \Cref{lem:nabla-f-decomposition}, $\alpha_1$ is the sensitivity for the true gradient direction, and $\alpha_i$ for $i\neq 1$ is the sensitivity for orthogonal directions.
            To increase the cosine similarity, we can increase the \textcolor{eqblue}{blue} term $\textcolor{eqblue}{\bm{\alpha_1}}$, or reduce the \textcolor{eqgreen}{green} term $\textcolor{eqgreen}{\bm{\frac{\sum_{i=2}^m \alpha_i^2}{m-1}}}$ and $\textcolor{eqgreen}{\bm{\alpha_{\max}^4}}$.
            In \Cref{subsec:analysis-verify}, we empirically verify that PGAN achieves this, where $\alpha_1^2$ is consistently and significantly larger than $\frac{\sum_{i=2}^m \alpha_i^2}{m-1}$, and therefore we leverage PGAN in our implementation.
            Note that for identical projection~\cite{chen2020hopskipjumpattack} or orthogonal projection~\cite{Li_2020_CVPR}, all $\alpha_i$'s are equal.
            The performance gain in NonLinearBA~\cite{li2020nolinear} can be explained by 
            this characteristic.
            \vspace{-0.2em}
        \end{enumerate}
        We illustrate these key characteristics in \Cref{adxsec:key-characteristics}.
        Other factors that can improve the cosine similarity are also revealed, such as smaller step size $\delta$, and larger sampling~(i.e., query) numbers $B$. 
        However, they directly come at the cost of more queries as discussed in \cite{chen2020hopskipjumpattack}.

        \begin{figure}[!t]
            \centering
            \includegraphics[width=0.47\textwidth]{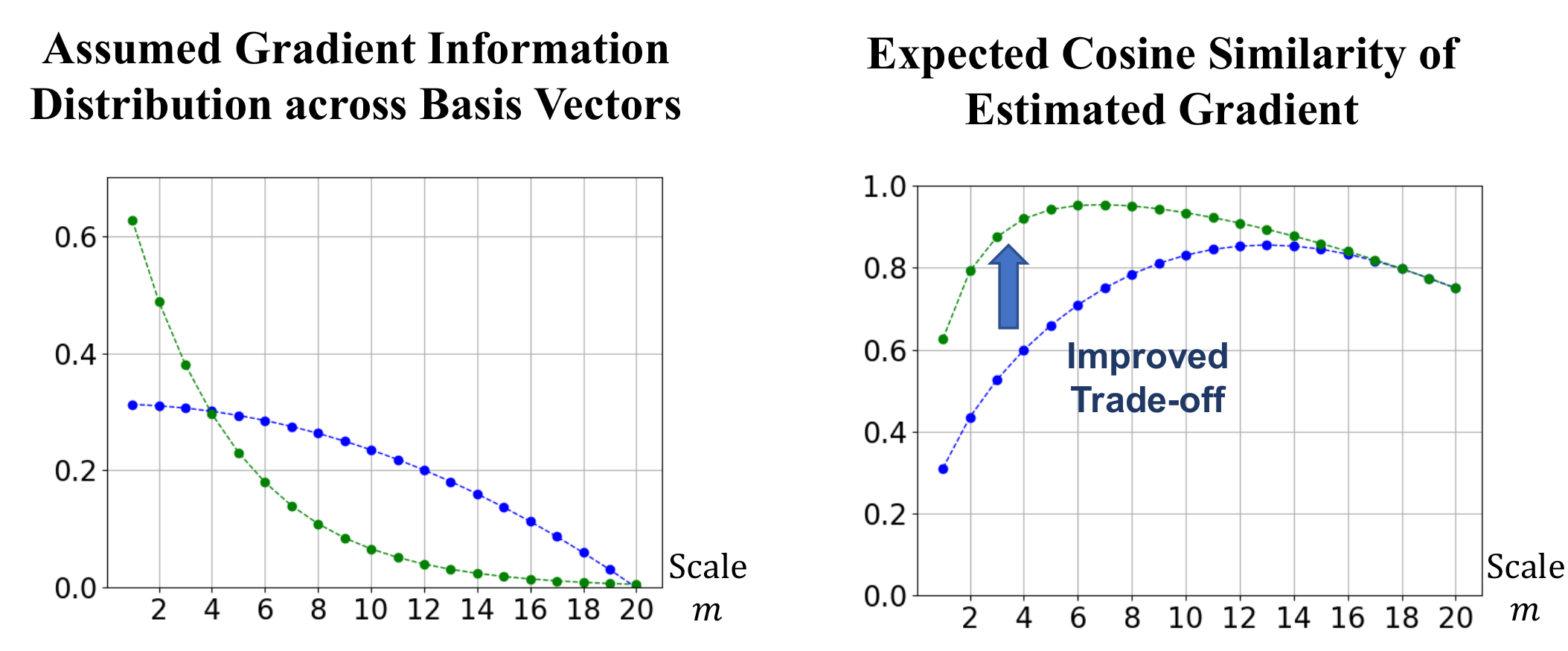}
             \vspace{-3mm}
            \caption{\small An illustration of why selected subspace improves the gradient estimation.
            \textbf{Left}: we assume a quadratic distribution of gradient information for $\f$'s basis~(\textcolor{blue}{blue curve}) and a more concentrated exponential distribution for frequency basis~(\textcolor{darkgreen}{green curve}).
            \textbf{Right}: the corresponding expected cosine similarity is numerically computed with settings $n = 20$, $\beta_S = 0.5$, $\beta_{\f} = 0$, $\alpha_i = 1$, $\delta = m^{-1}$.
            The improved trade-off w.r.t. scale is revealed.
            }
            \label{fig:progressive-scaling}
        \end{figure}
        
        To improve the precision and query efficiency of the gradient estimation, next we consider how to optimize the 
        estimator 
        on the above characteristics, especially (1) and (2), given that (3) can be achieved by a PGAN-based projection.
       
        \noindent\textbf{Trade-Off on Dimensionality.}
        From \Cref{fig:order-figure},
        we observe an apparent trade-off between Key Characteristics (1) and (2): when reducing the dimensionality $m$ of the projection subspace~(goal (1)), the preserved gradient information $\|\proj_{\nabla\f(0)} \nabla S(\xt)\|_2$ in this $m$-dimensional linear subspace $\nabla\f(0)$ becomes less, which opposes goal (2). 
        To make a tradeoff between (1) and (2), based on above observation, there exists an optimal dimensionality $m$ for the projection, and this dimensionality depends on how much true gradient information the projection subspace can preserve.
        
        \subsection{Optimize via Selecting Projection Subspaces}
        \label{subsec:projection-on-selected-subspace}
        
        To circumvent the intrinsic trade-off,
        instead of hoping that the end-to-end trained $\f$ can capture much true gradient information in its linear subspace $\nabla\f(0)$, we can actively constrain the projection $\f$ on a representative subspace.
        
        Here we focus on the linear subspace which can be represented by a linear combination of a set of basis, since the small step size $\delta$ of gradient estimator implies that only the local geometry matters and the local geometry of general subspace can be sufficiently approximated in first-order by linear subspace.
        Concretely, we select an $m$-dimensional linear subspace $V \subseteq \sR^n$.
        Then, we train the projection $\f$ on $V$, i.e., $\im(\f) \subseteq V$.
        Finally, we estimate the gradient with this $\f$ as: $u \in \sR^m \overset{\f}{\mapsto} \Delta \f(\delta u) \in V \subseteq \sR^n$, where $\dim(V) = m$.
        Interchangeably, we call $m$, the dimensionality  of $V$, as \emph{scale}, since it reflects the scale of the projection subspace $V$ as we will show later.
        
        Now we can analyze the cosine similarity of this new workflow.
        Since the image of projection $\f$ is in $V$, and $\rank(\nabla\f(0)) = \dim(V) = m$,
        we have $\Span(\nabla\f(0)) = V$ and $\|\proj_{\nabla\f(0)} \nabla S(\xt)\|_2 / \|\proj_V \nabla S(\xt)\|_2 = 1$.
        Plugging this into the above theorems and deriving from simple geometry~(details in \Cref{lem:cosine-similarity-projected-subspace}), 
        we find that the cosine similarity bound for subspace-constrained $\f$ is of the same form as in \Cref{fig:order-figure}, with all $\|\proj_{\nabla\f(0)} \nabla S(\xt)\|_2^2$ replaced by $\|\proj_{V} \nabla S(\xt)\|_2^2$.
    
        \noindent\textbf{Selected Subspace Improves Gradient Estimation.}
        The formulation in  \Cref{fig:order-figure} reveals that
        as long as we select an $m$-dimensional linear subspace $V$ that preserves more gradient information $\|\proj_V \nabla S(\xt)\|_2$ than the unconstrained projection model $\|\proj_{\nabla\f(0)} \nabla S(\xt)\|_2$, the estimated gradient would have higher cosine similarity.
        We empirically find that the low-frequency subspace\footnote{Low-frequency subspace in DCT basis is a linear subspace.}
        satisfies such condition~(\Cref{subsec:analysis-verify}): for real-world images, the gradient information of classifiers is highly concentrated on low-frequency domain.
        This is also cross-validated in the literature~\cite{yin2019fourier}.
        We illustrate this analysis along with curves from numerical experiments in \Cref{fig:progressive-scaling}.
        
        Recall that in \ourapproach, 
        we train $\f$ on a smaller spatial scale and scale up its output, which is equivalent to constraining $\f$ on the low-frequency subspace.
        Since this subspace is more representative than the subspace $\nabla\f(0)$ of end-to-end trained $\f$, theoretically \ourapproach can estimate gradient with higher cosine similarity within fewer queries.
        

        \noindent\textbf{Existence of Optimal Scale.}
        From \Cref{fig:order-figure}, we find that the trade-off between Key Characteristics (1) and (2) in \Cref{subsec:key-characteristics} still exists for selected subspace in general.
        Now it transforms to the existence of an optimal scale.
        This is revealed by the green curve in \Cref{fig:progressive-scaling}.
        For spatial and frequency scales, across different images from the same dataset and model, the coefficients of gradient information on the frequency basis vector tend to be very stable, as \Cref{subsec:analysis-verify} shows.
        It implies that $\|\proj_V \nabla S(\xt)\|_2$ in \Cref{fig:order-figure} across different images tend to be stable, and the optimal scales for gradient estimation tend to be stable too.
        Therefore, we can search for the optimal scale with a validation dataset.
\section{Experimental Evaluation}
    \label{sec:exp}
    With the established general framework  of leveraging progressive scaling to improve attack efficiency, in this section, we take PGAN as an instantiation  and conduct extensive experiments to 1)~verify our theoretical analysis; 2)~show that \ourapproach outperforms existing blackbox attacks by a significantly large margin.
    We also present some additional interesting findings.
    
    \subsection{Experimental Setup}
        \noindent\textbf{Target Models.}
        We use both offline models and a commercial online API as target models. For offline models, following~\cite{Li_2020_CVPR,li2020nolinear}, pretrained ResNet-18 on MNIST, CIFAR-10, CelebA and ImageNet are utilized. We also evaluate model ResNeXt50\_32$\times$4d~\cite{xie2017aggregated} to demonstrate the generalization ability. On datasets MNIST and CIFAR-10, we scale up the input images to $224 \times 224$ by bilinear interpolation to help explore the influence of different scales. On CelebA, the target model is fine-tuned to perform the binary classification on the attribute ‘Mouth\_Slightly\_Open’. The benign performance of target models is shown in Appendix~\ref{sec:tgt_model_performance}. For the commercial online API, the `Compare' API~\cite{facepp-compare-api} from MEGVII Face++ which determines whether the faces from two images belong to the same person is used as the target model. The compared images are chosen from CelebA.
        The rationale of selecting these classification tasks and a detailed description of target models are discussed in \Cref{adxsec:target-model}.
        
        
        \noindent\textbf{Training Procedure of PGAN.}
        PGAN is trained to generate gradient images of reference models with small resolution until reaching convergence, and then new layers will be added to double the output scale. 
        The PGAN training details could be found in \Cref{sec:pgan_training_procedure} and reference models' performance are shown in \Cref{sec:ref_model_performance}. 
        For simplicity, we will denote `PGAN28' as the attack using the output of PGAN with scale $28\times28$.
        
        
        \noindent\textbf{Implementation.}
        We follow the description in \Cref{sec:approach} to implement \ourapproach.
        Compared to other common attacks,
        we additionally train PGAN and use an additional validation set of ten images to search for the optimal scale.
        More implementation details are shown in \Cref{sec:alg_pseudocode,sec:attack_process_implementation}.
        
         \noindent\textbf{Baselines.}
         We consider
         six state-of-the-art decision-based attacks as the baselines.
         Among our baselines, \textbf{QEBA}~\cite{Li_2020_CVPR} and \textbf{NLBA}~\cite{li2020nolinear} utilize dimension reduction to sample from low-dimensional space, while the  \textbf{Sign-OPT}~\cite{cheng2020signopt} and \textbf{HSJA}~\cite{chen2020hopskipjumpattack} apply direct  Monte-Carlo sampling for gradient estimation. \textbf{EA}~\cite{dong2019efficient} adopts evolution algorithm to perform the attack. Note that we directly select the optimal scale for EA to compare under its optimal case.
         In \Cref{sec:rays_attack} we compare with \textbf{RayS} attack~\cite{chen2020rays}.
         
        \noindent\textbf{Evaluation Metrics.}
         We adopt the standard evaluation metrics: 1) the Mean Squared Error~(MSE) between the optimized adversarial examples and  target image under different queries (this process will guarantee 100\% attack success rate); 2) attack success rate at a specific MSE perturbation bound when the query number is constrained. Note that the conversion between MSE and $\ell_2$ metrics are straightforward and order-preserving. 
        
        \begin{figure}[t]
            \centering
            \includegraphics[width=\linewidth]{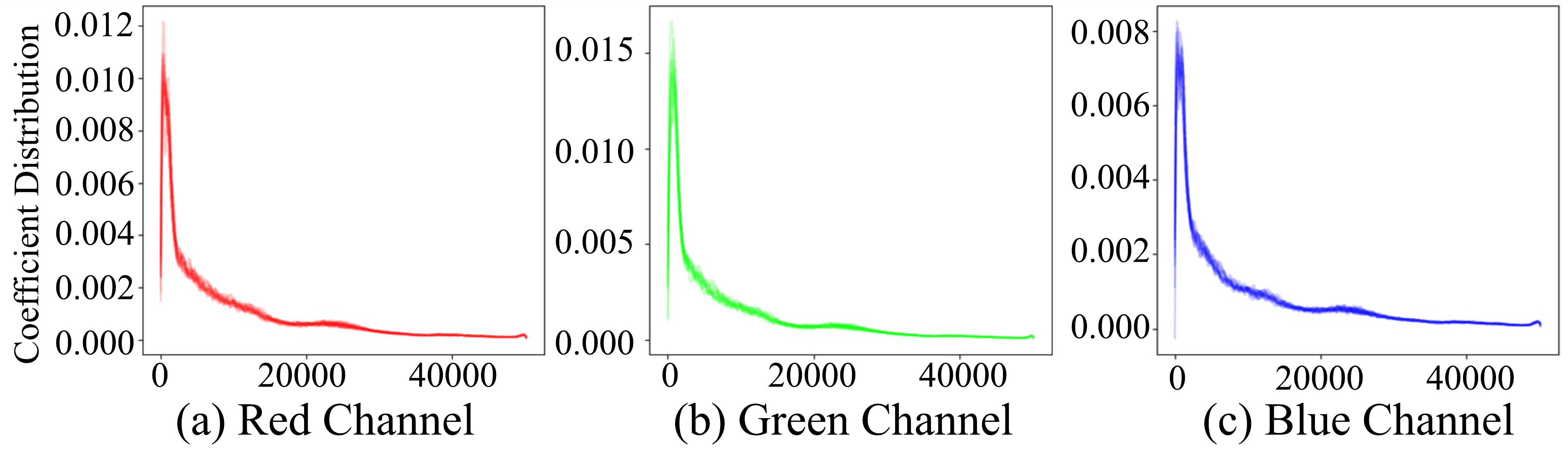}
            \vspace{-7mm}
            \caption{\small The long tail distribution of the coefficients of gradients generated from 10 images on the validation set of CIFAR-10 and represented on DCT basis for each channel.}
            \label{fig:dct_distribution}
        \end{figure}
        
        \begin{figure}[t]
            \centering
            \vspace{-3mm}
            \includegraphics[width=0.45\linewidth]{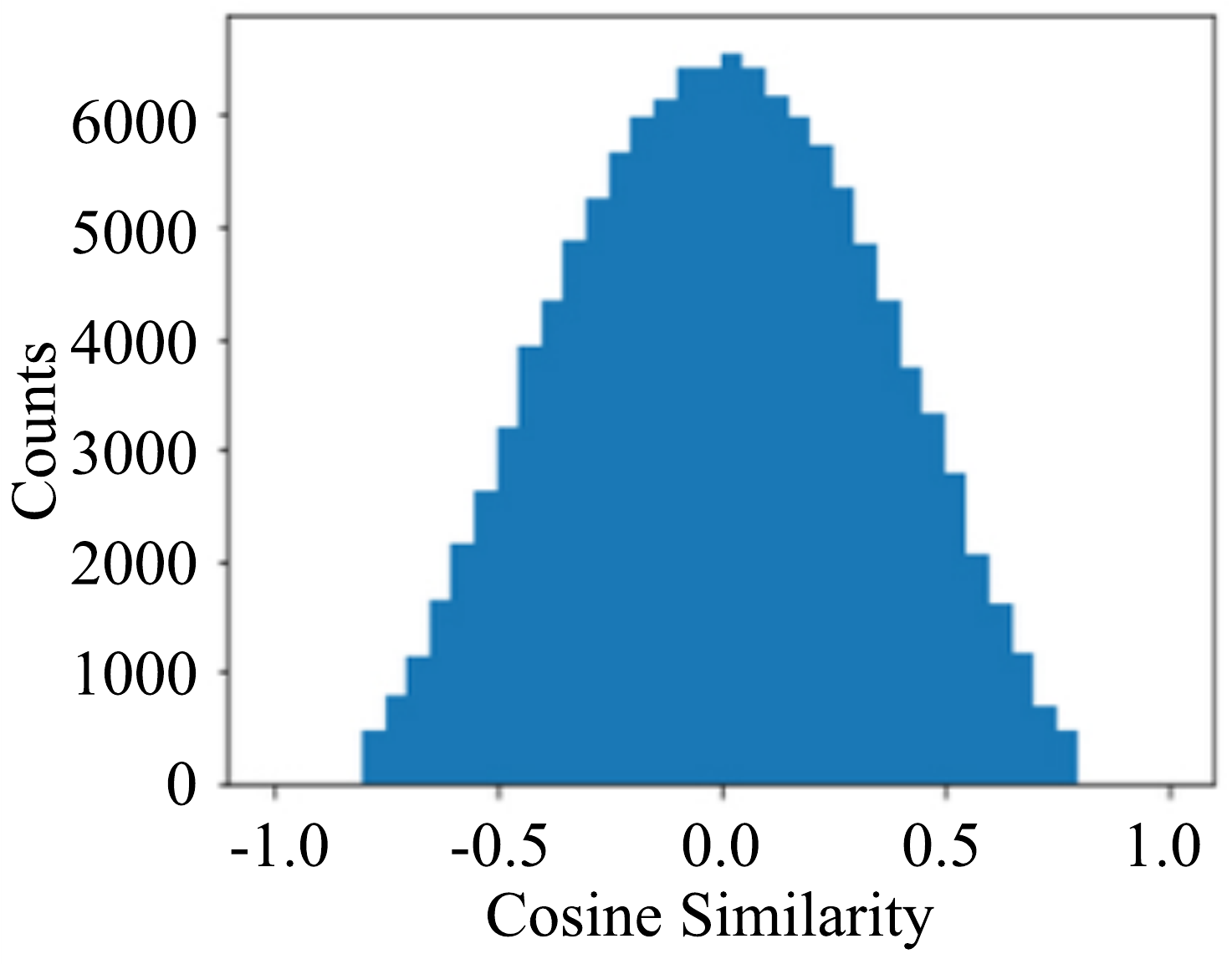}
            \vspace{-4mm}
            \caption{\small Pairwise cosine similarity of $\nabla \f(0)$ column vectors.}
            \label{fig:s3_08}
            \vspace{-0.6cm}
        \end{figure}
        
         \begin{figure*}[!ht]
            \centering
            \includegraphics[width=0.9\textwidth]{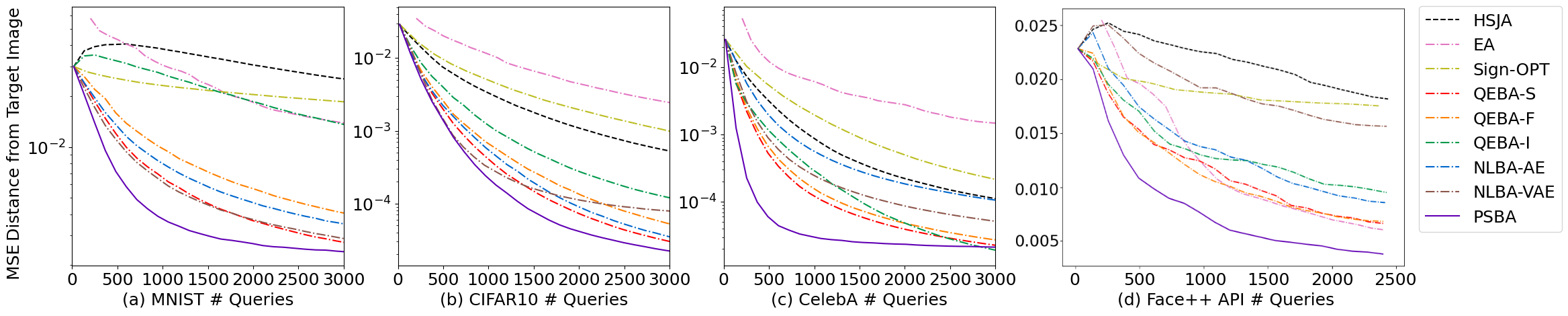}
            \vspace{-0.2cm}
            \includegraphics[width=0.9\textwidth]{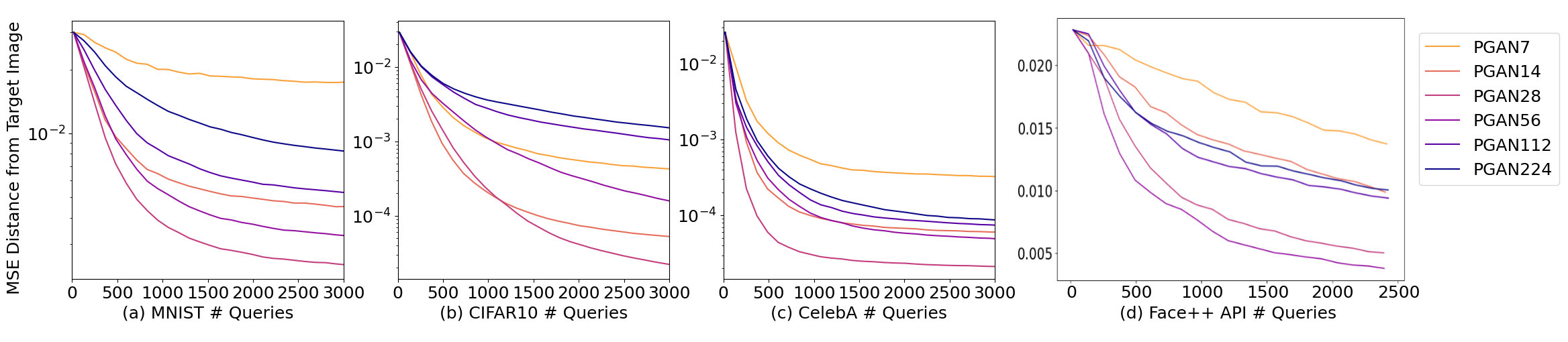}
            \vspace{-4mm}
            \caption{\small \textbf{Row 1}: Perturbation magnitude (MSE) w.r.t. query numbers for attacks on diverse datasets/models. \textbf{Row 2}: Perturbation magnitude when choosing different scales as the projection subspaces. The target model in (a)-(c) is ResNet-18, and an online commercial API in (d).}
            \label{fig:spatial_mean}
        \end{figure*}

         \begin{table*}[!ht]
            \vspace{-1.0em}
            \renewcommand\arraystretch{1.05}   
            \centering
    		\caption{\small Comparison of the attack success rate  for different attacks at query number 2K (the perturbation magnitude under MSE for each dataset are: MNIST: $5e-3$; CIFAR10: $5e-4$; CelebA: $1e-4$; ImageNet: $1e-2$).}
            \label{tab:numerical_attack_performance}
            \resizebox{0.9\linewidth}{!}{
            \begin{tabular}{c|c|c|c|c|c|c|c|c|c|c}
            \toprule
            \hline
    		\multirow{2}{*}{Data}     & \multirow{2}{*}{Model} & \multicolumn{9}{c}{\# Queries = 2K}                                               \\ \cline{3-11} 
    		&                        & HSJA & EA & Sign-OPT & QEBA-S & QEBA-F & QEBA-I & NLBA-AE & NLBA-VAE & PSBA        \\ \hline
    		\multirow{2}{*}{MNIST}    & ResNet                 & 2\%    & 6\%  & 2\%        & 60\%     & 42\%     & 4\%      & 46\%      & 58\%       & \textbf{78\%} \\ \cline{2-11} 
    		& ResNeXt                & 4\%    & 4\%  & 6\%        & 76\%     & 66\%     & 16\%     & 70\%      & 80\%       & \textbf{88\%} \\ \hline
    		\multirow{2}{*}{CIFAR10}  & ResNet                 & 26\%   & 10\% & 10\%       & 82\%     & 70\%     & 58\%     & 76\%      & 82\%       & \textbf{94\%} \\ \cline{2-11} 
    		& ResNeXt                & 32\%   & 0\%  & 18\%       & 88\%     & 72\%     & 64\%     & \textbf{90}\%      & \textbf{90}\%       & \textbf{90\%} \\ \hline
    		\multirow{2}{*}{CelebA}   & ResNet                 & 20\%   & 8\%  & 2\%        & 80\%     & 70\%     & 72\%     & 20\%      & 46\%       & \textbf{90\%} \\ \cline{2-11} 
    		& ResNeXt                & 24\%   & 6\%  & 6\%        & 60\%     & 56\%     & 72\%     & 20\%      & 38\%       & \textbf{88\%} \\ \hline
    		\multirow{2}{*}{ImageNet} & ResNet                 & 24\%   & 24\% & 6\%        & \textbf{54}\%     & 52\%     & 46\%     & 44\%      & 28\%       & \textbf{54\%} \\ \cline{2-11} 
    		& ResNeXt                & 20\%   & 22\% & 16\%       & 40\%     & 38\%     & 36\%     & 28\%      & 26\%       & \textbf{42\%} \\ \hline
    		\bottomrule
            \end{tabular}
            }
            \vspace{-0.3cm}
        \end{table*}
        
        \begin{figure}[!ht]
            \centering
            \includegraphics[width=0.80\linewidth]{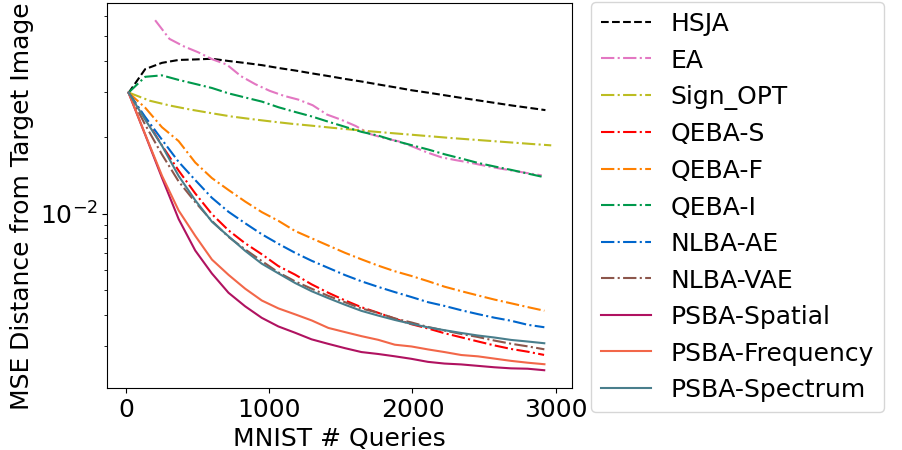}
            \vspace{-0.1cm}
            \caption{\small Perturbation magnitude (MSE) under different number of queries on MNIST for \ourapproach sampling from spatial, frequency, and spectrum domains. 
            }
            \label{fig:ppba_sfs}
            \vspace{-0.6cm}
        \end{figure}   
        

        \subsection{Verification of Theoretical Findings}  
        \label{subsec:analysis-verify}
        \noindent\textbf{Low Frequency Concentration.}
        In order to verify the hypothesis of low frequency concentration, we randomly sample 10 gradient vectors calculated on ResNet-18 model for CIFAR-10 dataset. The gradients are normalized and transformed with DCT basis for each channel.
        As shown in ~\Cref{fig:dct_distribution}, the $x$-axis shows the DCT basis from low to high frequency, and $y$-axis represents the corresponding coefficients denoised by the Savitzky-Golay filter.
        The three curves denote the three color channels respectively. The figures show stable long tail distribution across various images, and this implies the benefit of selecting the low-frequency subspace as the projection subspace in gradient estimation, as well as the existence of a stable optimal scale for the same dataset and target model, providing strong evidence for the discussion in \Cref{subsec:projection-on-selected-subspace}. In addition, we draw a graph in a more statistical sense  in~\Cref{sec:long_tailed_distribution}.   
        
        \noindent\textbf{High Sensitivity on Gradient Direction.}
        As shown in \Cref{sec:high_sensitivity}, for trained PGAN, we compare the sensitivity~(\Cref{lem:nabla-f-decomposition}) for the projected true gradient direction, $\alpha_1^2$, and the averaged sensitivity for other orthogonal directions, $\sum_{i=2}^m \alpha_i^2 / (m-1)$.
        On all datasets, $\alpha_1^2$ is significantly larger than $\sum_{i=2}^m \alpha_i^2 / (m-1)$, which implies that trained PGAN achieves higher sensitivity on the true gradient direction.
        This is exactly the goal (3) in \Cref{subsec:result-interpretation}.
        In constrast, the identical~\cite{chen2020hopskipjumpattack} and othogonal projection~\cite{Li_2020_CVPR} have identical directional sensitivity.
            
        \noindent\textbf{To what extent does orthogonality assumption hold.} Here, we compute $\nabla\f(0)$ of PSBA on ImageNet at the optimal scale and then cluster the similar column vectors~(those with cosine similarity $> 0.8$ or $<-0.8$) since they contribute to the sensitivity of one direction. Next, we compute the pairwise cosine similarity between clusters. 
        The histogram of clusters based on their cosine similarity is shown in~\cref{fig:s3_08}.
        As we can see, the histogram concentrates at $0$, i.e., orthogonal pairs are most frequent.
        We also remark that recent orthogonal training can also enforce the assumption~\cite{huang2020controllable}.
        
    
    \subsection{Attack Performance Evaluation}
        In this section, we show that the optimal scale indeed exists and our method \ourapproach\ourvariant outperforms other state-of-the-art baselines in terms of attack effectiveness and efficiency. In addition, by selecting the optimal scale, \ourapproach\ourvariant can also successfully and efficiently attack the online commercial face recognition API. Here, we randomly select $50$ pairs of source and target images from validation set that are predicted by the target model as different classes for both offline attack and online attack and the selections of other hyperparameters are shown in~\Cref{sec:attack_setup}. 
        
        \noindent\textbf{Offline Attack.}
        The attack performance of different approaches in terms of the perturbation magnitude (MSE) between the adversarial  and target image is shown in~\Cref{fig:spatial_mean} (a)-(c). As we can see from Row 1, \ourapproach\ourvariant  effectively decrease the MSE when the number of queries is small and outperforms all baselines.
        Detailed comparisons on the gradient cosine similarity are in~\Cref{sec:cosine_similarity}. 
%
        From Row 2 we can see that, interestingly, the optimal scale found by \ourapproach\ourvariant across four datasets is consistently $28\times 28$. 
        
        In \Cref{tab:numerical_attack_performance}, we show the attack success rate when the query number is constrained by 2K.
        This is because we can not easily generate a large number of queries~(e.g., exceeding 2K) for attacking one image, and our \ourapproach is designed to be a query efficient attack with fast convergence speed.
        %
        We can see that PSBA indeed significantly outperforms other methods when attacking Face++ API under small query budgets. On the other hand, when the query budgets get bigger, most of the methods would converge and attack successfully, then the comparison under this circumstance is not quite useful. 
        We leave the results with large query number constraints in \Cref{sec:different_tgt_model}. Besides, we also compare our method with RayS attack~\cite{chen2020rays} in~\Cref{sec:rays_attack}. Detailed discussions on the computation time and resource consumption are in~\Cref{sec:time_consumption}, and  the visualized results for other target models and ImageNet are in~\Cref{sec:different_tgt_model}.


        \noindent\textbf{Online Attack.} 
        To demonstrate the generalization and practicality of \ourapproach\ourvariant, we perform it against a real-world online commercial API as shown in~\Cref{fig:spatial_mean} (d). 
        Although the PGAN used here is trained on ImageNet, \ourapproach\ourvariant still outperforms other baselines and interestingly, it adopts the optimal scale as $56 \times 56$.

        
    
        
        
    \noindent\textbf{Frequency and Spectrum Domains.}
        In addition to the spatial domain, we also evaluate \ourapproach on frequency and spectrum domains.
        The results are shown in~\Cref{fig:ppba_sfs}.
        \ourapproach outperforms other baselines.
        Detailed implementations and other ablation studies are included in~\Cref{sec:ppba_freq_spectrum}. 
      
      \noindent\textbf{Additional Findings.} 
      (1)~In \Cref{sec:high_sensitivity}, we deliberately adjust the sensitivity on different directions $\{\alpha_i\}_{i=1}^m$ for given projection $\f$ to study the correlation between sensitivity and empirical attack performance.
      The results conform to our theoretical findings~(goal (3) in \Cref{subsec:key-characteristics}).
      (2)~We empirically study the optimal scale across model structures and show that different models have their own preference, which we believe will lead to interesting future directions.
      More details can be found in~\Cref{sec:optimal_scale_dif_model}.
      
        
\section{Conclusion}
    In this paper, we propose \ourapproach, a progressive-scale blackbox attack via projective gradient estimation.
    We propose a general theoretical framework to analyze existing projective gradient estimators, show key characteristics for improvement, and justify why \ourapproach outperforms other blackbox attacks.
    Extensive experiments verify our theoretical findings and show that \ourapproach outperforms existing blackbox attacks significantly against various target models including a real-world face recognition API.

\section*{Acknowledgements}
    We thank the anonymous reviewers for valuable feedback.
    This work is partially supported by NSF grant No.1910100, NSF CNS 20-46726 CAR, and Amazon Research Award.

\bibliography{bib}
\bibliographystyle{icml2021}

\onecolumn

\appendix

\newpage

 

\icmltitle{Supplementary Material for\texorpdfstring{\\}{} Progressive-Scale Blackbox Attack via Projective Gradient Estimation}

\begin{icmlauthorlist}
\icmlauthor{Jiawei Zhang}{equal,zju}
\icmlauthor{Linyi Li}{equal,uiuc}
\icmlauthor{Huichen Li}{uiuc}
\icmlauthor{Xiaolu Zhang}{antfin}
\icmlauthor{Shuang Yang}{ali}
\icmlauthor{Bo Li}{uiuc}
\end{icmlauthorlist}

In \Cref{adxsec:main-theorem-proof}, we show a summary of our theoretical results, compare it with related work, and demonstrate the complete proofs.
In \Cref{adxsec:key-characteristics}, we visualize the key characteristics for improving the gradient estimator, and formally justify the existence of the optimal scale.
In \Cref{adxsec:target-model}, we show how the target models, i.e., the models we attacked in the experiments, are prepared.
In \Cref{adxsec:ppba-pgan-detail}, we include more details about our \ourapproach-PGAN, such as the architecture and training of the projection model, the detailed algorithmic description of the progressive scaling procedure, and the implementation details.
In \Cref{adxsec:quantitative}, we show additional quantitative experimental results and ablation studies.
Finally, in \Cref{adxsec:qualitative}, we randomly sample a few original and attacked image pairs to demonstrate the efficiency of our attack compared with other baselines.

\section{Theorems and Proofs}
    \label{adxsec:main-theorem-proof}
    This appendix contains a discussion and comparison of theoretical results and all omitted mathematical proofs.
    
    \subsection{An outline of Main Theoretical Results}
        \label{adxsec:outline-theoretical-results}
        
        We summarize our main theoretical results---the lower bound of cosine similarity between the estimated gradient and the true gradient, in \Cref{tab:theory-result-summary}.
        
        In the table:
        \begin{itemize}[itemsep=-0.5mm]
            \item ``Expectation'' indicates the bound of the expected cosine similarity;
            \item ``Concentration'' indicates the bound of the cosine similarity that holds with probability at least $1-p$;
            \item ``At Boundary'' indicates the case where the estimated point is an exact boundary point, i.e., $S_{x^*}(x) = 0$;
            \item ``Approaching Boundary'' indicates the general case where the estimated point is away from the decision boundary within a small distance $\theta$~(measured along the true gradient direction).
        \end{itemize}
        
        ``At Boundary'' is actually a special case of ``Approaching Boudary'' with $\theta = 0$.
        In the main text, we only present \Cref{thm:approach-boundary-expectation,thm:approach-boundary-concentration} that are for the general case, i.e., ``Approaching Boundary'' case.
        
        \begin{table}[!h]
            \caption{A brief summary of the cosine similarity bounds for the boundary gradient estimator in \Cref{subsec:general-bound}.}
            \centering
            \small
            \begin{tabular}{c|c|c}
                \toprule
                & Expectation & Concentration \\
                \midrule
                At Boundary & \Cref{thm:at-boundary-expectation} & \Cref{thm:at-boundary-tail} \\
                Approaching Boundary & \Cref{thm:approach-boundary-expectation} & \Cref{thm:approach-boundary-concentration} \\
                \bottomrule
            \end{tabular}
            \label{tab:theory-result-summary}
        \end{table}
        
    \subsection{Comparison of Theoretical Results}
    
        \label{adxsec:theory-compare}
        We compare our theoretical results with existing work in \Cref{tab:theory-result-compare}.
        Note that it is better to have fewer assumptions and be applicable to more scenarios.
        As one can observe, our theoretical result is among the most general ones.
        Furthermore, as discussed in \Cref{subsec:general-bound}, ours is also among the tightest ones.
        From these tightest bounds, under general assumptions, we are able to discover the key characteristics and the existence of the optimal scale.
        The coarse bounds from the previous work cannot reflect these properties.
        
        \begin{table}[!h]
            \caption{A brief comparison of the cosine similarity bounds for the boundary gradient estimator in our work with existing work.}
            \centering
            \resizebox{0.98\textwidth}{!}{
            \begin{tabular}{c||c|c|c|c||c|c|c|c|c}
                \toprule
                & \multicolumn{4}{c||}{Scenario}
                & \multicolumn{5}{c}{Assumption} \\
                \cline{2-10}
                & \multicolumn{2}{c|}{At Boundary} & \multicolumn{2}{c||}{Approaching Boundary} & Sampling & \multicolumn{4}{c}{Projection} \\ 
                \cline{2-10}
                & Expectation & Concentration & Expectation & Concentration & Orthogonal & Identical & Linear & Orthogonal & No Bias \\
                \hline \hline
                HSJA~\cite{chen2020hopskipjumpattack} & \checkmark &  & \checkmark &  & & \checkmark & \checkmark & \checkmark & \checkmark \\
                \hline
                QEBA~\cite{Li_2020_CVPR} & \checkmark & & & & \checkmark & & \checkmark & \checkmark & \checkmark \\
                \hline
                NonLinear-BA~\cite{li2020nolinear} & \checkmark & & & & \checkmark & & & \\
                \hline
                Ours & \checkmark & \checkmark & \checkmark & \checkmark & & & & & \checkmark \\
                \bottomrule
            \end{tabular}
            }
            \label{tab:theory-result-compare}
        \end{table}

\subsection{Proof of \texorpdfstring{\Cref{lem:nabla-f-decomposition}}{Lemma 4.1}}

    \label{adxsec:proof}
    
    \allowdisplaybreaks
    
    \label{adxsubsec:proof-lemma-3-1}
    \begin{customlem}{\ref{lem:nabla-f-decomposition}}[$\nabla \f$ Decomposition]
        Under the assumption in \Cref{sec:assumption},
        there exists a singular value decomposition of $\nabla\f(0) = \mU \mSigma \mV^\T$ such that
        $$
            \mU_{:,1} = \proj_{\nabla \f(0)} \nabla S(\xt) / \|\nabla S(\xt)\|_2
            \text{ or }
            \mU_{:,1} = -\proj_{\nabla \f(0)} \nabla S(\xt) / \|\nabla S(\xt)\|_2
        $$
        where $\mU \in \sR^{n\times n}$ and $\mV \in \sR^{m\times m}$ are orthogonal matrices;
        $\mSigma = \diag(\alpha_1,\alpha_2,\dots,\alpha_m) \in \sR^{n\times m}_{\ge 0}$ is a rectangular diagonal matrix with $\alpha_1 > 0$.
    \end{customlem}
    
    \begin{proof}[Proof of \Cref{lem:nabla-f-decomposition}]
        For simplicity, we define $\mM := \nabla \f(0)$.
        According to the assumption, there exists a column vector $\mM_{:,c}$ that is not orthogonal with the gradient direction $\nabla S(\xt)$.
        If $c \neq 1$, we define $\mT := \left[ e_c \, e_2 \, \cdots \, e_{c-1} \, e_1 \, e_{c+1} \, \cdots \, e_m \right]$; otherwise, we let $\mT := \mI_m$.
        Here, $e_i \in \sR^m$ is a standard basis vector, i.e., it satisfies $(e_i)_i = 1$ and $(e_i)_j = 0$ for any $j \neq i$.
        As the result, $$
            \mM = [\mM_{:,c}\, \mM_{:,2}\, \mM_{:,3}\, \cdots\, \mM_{:,c-1}\, \mM_{:,1}\, \mM_{:,c+1}\, \cdots \mM_{:,m}] \mT := \mM' \mT.
        $$
        Here $\mM_{:,i}$ stands for the $i$-th column vector of $\mM$.
        $\mM'$ just exchanges the first column vector with the $c$-th column vector of $\mM$.
        According to the assumption in \Cref{sec:assumption}, $\mM'_{:,1}$ is aligned with $\proj_{\mM} \nabla(\xt)$, and for $i \ge 2$, we have $\langle \mM'_{:,i}, \mM'_{:,1} \rangle = 0$.
        
        Now, we apply QR decomposition to $\mM'$ via Gram-Schmidt Process, which yields $\mM' = \mU' \mR$, where $\mU' \in \sR^{n \times n}$ is an orthogonal matrix, and $\mR \in \sR^{n \times m}$ is an upper-triangular matrix.
        We are going to show two interesting properties of $\mU'$ and $\mR$:
        1)~$\mU'_{:,1} = \frac{\proj_{\mM} \nabla S(\xt)}{\|\proj_{\mM} \nabla S(\xt)\|_2}$ or $\mU'_{:,1} = -\frac{\proj_{\mM}\nabla S(\xt)}{\|\proj_{\mM}\nabla S(\xt)\|_2}$;
        and 2)~$\mR$ can be written as 
        $
            \left[ \begin{matrix}
                \alpha_1 & 0 \\
                0 & \mR' \\
            \end{matrix} \right]
        $
        where $\alpha_1>0$ and $\mR'$ is an upper-triangular matrix.
        The first property is apparent, since in Gram-Schmidt Process, we always have $\mU'_{:,1} = \mM'_{:,1} / \|\mM'_{:,1}\|_2$.
        Thus, it is equal to $\pm\nabla \proj_{\mM}S(\xt) / \|\proj_{\mM}\nabla S(\xt)\|_2$.
        For the second property, according to the definition of the process, $(\mR)_{1,i} = \langle \mM'_{:,1}, \mM'_{:,i} \rangle = 0$.
        Meanwhile, $\alpha_1 = \|\mM'_{:,1}\|_2^2 > 0$ since $\mM'_{:,1}$ aligns with $\nabla S(\xt)$ and it is non-zero.
        
        We apply SVD decomposition to the sub-matrix $\mR' \in \sR^{(n-1)\times(m-1)}$: $\mR' := \mS'\mSigma'\mW'^\T$.
        Here, $\mS' \in \R^{(n-1)\times(n-1)}$ and $\mW' \in \sR^{(m-1)\times(m-1)}$ are orthogonal matrices, while $\mSigma' \in \sR^{(n-1)\times(m-1)}$ is a triangular diagonal matrix.
        Therefore, $\mR$ can be decomposed as such:
        $$
            \mR
            =
            \left[ \begin{matrix}
                \alpha_1 & 0 \\
                0 & \mR' \\
            \end{matrix} \right]
            = 
            \overbrace{
            \left[ \begin{matrix}
                1 & 0 \\
                0 & \mS' \\
            \end{matrix} \right]}^{\mS}
            \overbrace{
            \left[ \begin{matrix}
                \alpha_1 & 0 \\
                0 & \mSigma' \\
            \end{matrix} \right]}^{\mSigma}
            \overbrace{
            \left[ \begin{matrix}
                1 & 0 \\
                0 & \mW'^\T \\
            \end{matrix} \right]}^{\mW^\T}.
        $$
        It is easy to observe that $\mSigma = \diag(\alpha_1,\alpha_2,\cdots,\alpha_m)$ is a rectangular diagonal matrix with $\alpha_1 > 0$,
        and $\mV$ is an orthogonal matrix.
        Notice that
        $$
            \begin{aligned}
                \nabla \f(0) = \mM & = \mM'\mT = \mU'\mR\mT = \mU'\mS\mSigma\mW^\T\mT \\
                & = 
                \left[
                    \begin{matrix}
                        | & | & \cdots & | \\
                        \mU'_{:,1} & \mU'_{:,2} & \cdots  & \mU'_{:,m} \\
                        | & | & \cdots & | \\
                    \end{matrix}
                \right]
                \left[
                    \begin{matrix}
                        1 & 0 \\
                        0 & \mS' \\
                    \end{matrix}
                \right]
                \mSigma \mW^\T\mT \\
                & = \underbrace{\left[ 
                    \begin{matrix}
                        \mU'_{:,1} & \mU'_{:,2:m} \mS'
                    \end{matrix} 
                \right]}_{\mU} \mSigma \underbrace{\mW^\T\mT}_{\mV^\T}.
            \end{aligned}
        $$
        We have already shown $\mU'_{:,1} = \frac{\proj_{\nabla\f(0)}\nabla S(\xt)}{\|\proj_{\nabla\f(0)}\nabla S(\xt)\|_2}$ or $\mU'_{:,1} = -\frac{\proj_{\nabla\f(0)}\nabla S(\xt)}{\|\proj_{\nabla\f(0)}\nabla S(\xt)\|_2}$.
        To finish the proof, we only need to verify that $\mU$ and $\mV$ are orthogonal matrices.
        Since $\mU'$ and $\mS'$ are both orthogonal matrices, the $\mU'_{:,2:m} \mS'$ is a semi-orthogonal matrix~(the column vectors are unitary and orthogonal).
        Furthermore, $\mU'_{:,1} \perp \Span(\mU'_{:,2:m})$ because $\mU'$ is an orthogonal matrix.
        Thus, $\mU'_{:,1} \perp \Span(\mU'_{:,2:m}\mS')$.
        As the result, $\mU$ is an orthogonal matrix.
        For $\mV$, $\mV^\T \mV = \mW^\T \mT \mT^\T \mW = \mI_m$ so it is an orthogonal matrix.
    \end{proof}
    
    \subsection{Warmup: Expectation Bound at Boundary}
    \label{adxsubsec:proof-thm-at-boundary-expectation}
    
    As a warm-up, we begin with the special case where the point is exactly the boundary point.
    
    The proof of the following theorems require the following lemma.
    \begin{lemma}[Cosine Similarity in Projected Subspace]
        \label{lem:cosine-similarity-projected-subspace}
        Let $\mW \in \sR^{n\times m}$ be a matrix.
        The vector $w \in \sR^n$ is in $\Span(\mW)$, and the vector $v \in \sR^n$ has non-zero projection in $\Span(\mW)$, i.e., $\proj_{\mW} v \neq 0$.
        Then,
        \begin{equation}
            \cos\langle w, v\rangle = \cos\langle w, \proj_{\mW} v\rangle \cdot
            \dfrac{\|\proj_{\mW} v\|_2}{\|v\|_2}.
        \end{equation}
    \end{lemma}
    
    \begin{proof}[Proof of \Cref{lem:cosine-similarity-projected-subspace}]
        The lemma can be illustrated by simple geometry.
        To be rigorous, here we give an algebraic proof. 
        $$
            \cos\langle w, v \rangle 
            = 
            \dfrac{\langle w, v \rangle}{\|w\|_2 \|v\|_2}
            = 
            \dfrac{\langle w, v \rangle}{\|w\|_2 \|\proj_{\mW} v\|_2} \cdot \dfrac{\|\proj_{\mW} v\|_2}{\|v\|_2}.
        $$
        We notice that $w \in \Span(w)$, and $v = \proj_{\mW} v + (v - \proj_{\mW} v)$ where $(v - \proj_{\mW} v)$ is orthogonal to $w$.
        Thus,
        $$
            \langle w, v\rangle = \langle w, \proj_{\mW} v\rangle.
        $$
        So
        $$
            \cos\langle w, v\rangle = \cos\langle w, \proj_{\mW} v\rangle \cdot
            \dfrac{\|\proj_{\mW} v\|_2}{\|v\|_2}.
        $$
    \end{proof}
    
    \begin{remark}
        The lemma reveals that for any vector $w$ in $\Span(\mW)$, the maximum possible cosine similarity between $w$ and $v$ is $\|\proj_{\mW} v\|_2 / \|v\|_2$.
        This is achieved by setting $w = k\cdot\proj_{\mW} v$.
    \end{remark}

    \begin{theorem}[Expected cosine similarity; at boundary]
        The difference function $S$ and the projection $\f$ are as defined before.
        For a boundary point $\xt$ such that $S(\xt) = 0$,
        let estimated gradient $\tnablaS(\xt)$ be as computed by \Cref{def:gradient-estimator} with step size $\delta$ and sampling size $B$.
        Over the randomness of the sampled vectors $\{u_b\}_{i=1}^B$,
        \begin{equation}
            \cos\langle \E\tnablaS(\xt), \nabla S(\xt) \rangle
            \ge 
            \dfrac{\|\proj_{\nabla\f(0)}\nabla S(\xt)\|_2}{\|\nabla S(\xt)\|_2} \cdot \left( 1 - \dfrac{(m-1)^2\delta^2}{8\alpha_1^2} \left( \dfrac{\delta\gamma^2}{\alpha_1} + \dfrac{\gamma}{\alpha_1} \sqrt{\dfrac{\sum_{i=2}^m \alpha_i^2}{m-1}} + \sqrt{\dfrac{1}{m-1}} 1.58  \beta_{\f} \right)^2 \right),
            \label{eq:thm1-1-adx}
        \end{equation}
        where
        \begin{equation}
            \gamma := \beta_{\f} + \dfrac{\beta_S \left(\max_{i\in [m]} \alpha_i + \half \delta \beta_{\f}\right)^2}{\|\proj_{\nabla\f(0)}\nabla S(\xt)\|_2}.
            \label{eq:thm1-2-adx}
        \end{equation}
        \label{thm:at-boundary-expectation}
    \end{theorem}
    
    \begin{proof}[Proof of \Cref{thm:at-boundary-expectation}]
        We begin with an important lemma.
        
        \begin{lemma}
            \label{lem:a-1}
            We let
            \begin{equation}
                w := 
                \dfrac{1}{2}
                \delta
                \left(
                    \beta_{\f} \|\nabla S(\xt)\|_2
                    +
                    \beta_S \left( \|\nabla \f(0)\|_2 + \dfrac{1}{2} \delta \beta_{\f} \right)^2
                \right).
                \label{eq:w-in-lem-a-1}
            \end{equation}
            On the point $\xt$ such that $S(\xt) = 0$, for any $\delta > 0$ and unit vector $u \in \sR^n$,
            $$
                \begin{aligned}
                    \langle \nabla S(\xt),\, \nabla \f(0) \cdot u \rangle > w & \Longrightarrow \phi(\xt + \Delta\f(\delta u)) = 1, \\
                    \langle \nabla S(\xt),\, \nabla \f(0) \cdot u \rangle < -w & \Longrightarrow \phi(\xt + \Delta\f(\delta u)) = -1.
                \end{aligned}
            $$
        \end{lemma}
        
        \begin{proof}[Proof of \Cref{lem:a-1}]
            We prove the lemma by Taylor expansion and the smoothness condition on $S$ and $\f$.
            First, from Taylor expansion on function $\Delta f(\delta u)$ at the origin,
            \begin{equation}
                \Delta \f(\delta u) = \f(\delta u) - \f(0) = \nabla \f(0) \cdot (\delta u) + \half (\delta u)^\T \nabla^2 \f(\xi) (\delta u),
                \label{eq:lem-a-1-pf-1-taylor-f}
            \end{equation}
            where $\xi$ is a point on the segment between the origin and $(\delta u)$.
            Since $\f$ is $\beta_{\f}$-smooth, $\|\half(\delta u)^\T\nabla^2 \f(\xi) (\delta u)\|_2 \le \half \beta_{\f} \delta^2$.
            Thus,
            $$
                \langle \nabla S(\xt), \Delta \f(\delta u) \rangle \in 
                \delta \langle \nabla S(\xt), \nabla \f(0) \cdot u \rangle \pm \half \delta^2 \|\nabla S(\xt)\|_2 \beta_{\f}.
            $$
            Also we have
            $$
                \| \Delta \f(\delta u) \|_2 \le \|\nabla \f(0) \cdot (\delta u)\|_2 + \|\half (\delta u)^\T \nabla^2 \f(\xi) (\delta u)\|_2 \le \delta \|\nabla \f(0)\|_2 + \half \beta_{\f} \delta^2.
            $$
            Easily seen, this also applies to any point $\xi'$ between the origin and $(\delta u)$.
            Now, we apply Taylor expansion on function $S\left(\xt + \Delta \f(\delta u)\right)$ at point $\xt$, and get
            $$
                \begin{aligned}
                    S(\xt + \Delta \f(\delta u)) & = \overbrace{S(\xt)}^{0} + \langle \nabla S(\xt), \Delta \f(\delta u) \rangle + \half (\Delta \f(\xi'))^\T \nabla^2 S(\xt) (\Delta \f(\xi')) \\
                    &  \in \delta \langle \nabla S(\xt), \nabla \f(0) \cdot u \rangle \pm \half \delta^2 \|\nabla S(\xt)\|_2 \beta_{\f} \pm \half \beta_S \left( \delta \|\nabla \f(0)\|_2 + \half \beta_{\f} \delta^2 \right)^2 \\
                    & = \delta \left( \langle \nabla S(\xt),\nabla \f(0) \cdot u \rangle \pm \half \delta \left( \|\nabla S(\xt)\|_2 \beta_{\f} + \beta_S \left( \|\nabla \f(0) \|_2 + \half \delta \beta_{\f} \right)^2  \right) \right) \\
                    & = \delta \left( \langle \nabla S(\xt), \nabla \f(0) \cdot u \rangle \pm w \right).
                \end{aligned}
            $$
            Since the step size $\delta > 0$, 
           $$
                \begin{aligned}
                    \langle \nabla S(\xt),\, \nabla \f(0) \cdot u \rangle > w & \Longrightarrow S(\xt + \Delta\f(\delta u)) > 0, \\
                    \langle \nabla S(\xt),\, \nabla \f(0) \cdot u \rangle < -w & \Longrightarrow S(\xt + \Delta\f(\delta u)) < 0.
                \end{aligned}
            $$
            Observing that $\phi$ is the sign function of $S$ according to \Cref{def:S-and-phi}, we conclude the proof.
        \end{proof}
        \begin{remark}
            This lemma shows the connection between the value of sign function and the direction alignment between $\nabla\f(0) \cdot u$ and the true gradient.
        \end{remark}
        
        Then, we study the distribution of $\nabla\f(0) \cdot u$.
        \begin{lemma}
            \label{lem:a-2}
            \begin{equation}
                \langle \nabla S(\xt), \nabla \f(0) \cdot u \rangle \sim 
                \alpha_1 \|\proj_{\nabla\f(0)}\nabla S(\xt)\|_2 \left(
                2 \Beta\left( \frac{m-1}{2}, \frac{m-1}{2} \right) - 1
                \right).
                \label{eq:lem-a-2-1}
            \end{equation}
            Meanwhile, for any $c\in [-\|\nabla S(\xt)\|_2, +\|\nabla S(\xt)\|_2]$,
            \begin{equation}
                \E \left[\nabla \f(0) \cdot u \, | \, \langle \nabla S(\xt), \nabla \f(0) \cdot u \rangle = c\right] = c
                \dfrac{\proj_{\nabla\f(0)}\nabla S(\xt)}{\|\proj_{\nabla\f(0)}\nabla S(\xt)\|_2^2}.
                \label{eq:lem-a-2-2}
            \end{equation}
        \end{lemma}
        
        \begin{proof}[Proof of \Cref{lem:a-2}]
            According to \Cref{lem:nabla-f-decomposition}, $\nabla\f(0) = \mU\mSigma\mV^\T$.
            Since $\mV$ is an orthongonal basis of $\sR^m$, and $u$ is uniformly sampled from the uniform sphere $S^{m-1}$, we let $v = \mV^\T u$ and $v \sim \Unif(S^{m-1})$ too.
            Then, 
            \begin{equation}
                \begin{aligned}
                    \langle \nabla S(\xt), \nabla\f(0) \cdot u \rangle & = \langle \nabla S(\xt), \mU \mSigma v \rangle = \Big\langle \nabla S(\xt), \sum_{i=1}^m \alpha_i v_i \mU_{:,i} \Big\rangle \\
                    & = \alpha_1v_1 \|\proj_{\nabla\f(0)} \nabla S(\xt)\|_2 \Big\langle \dfrac{\proj_{\nabla\f(0)}\nabla S(\xt)}{\|\proj_{\nabla\f(0)}\nabla S(\xt)\|_2} , \mU_{:,1} \Big\rangle
                    \overset{(*)}{=} 
                    \pm \alpha_1 v_1 \|\proj_{\nabla\f(0)}\nabla S(\xt)\|_2,
                \end{aligned}
                \label{eq:lem-a-2-pf-1}
            \end{equation}
            where $(*)$ follows from $\mU$ is an orthogonal basis with $\mU_{:,1} = \pm \nabla \proj_{\nabla\f(0)} S(\xt) / \|\proj_{\nabla\f(0)} \nabla S(\xt)\|_2$ as \Cref{lem:nabla-f-decomposition} shows.
            
            From \cite{yang2020randomized}~(Lemma I.23), $\frac{1 + v_1}{2} \sim \Beta\left( \frac{m-1}{2}, \frac{m-1}{2} \right)$, where $\Beta(\cdot,\cdot)$ stands for the Beta distribution.
            As the result,
            $$
                \alpha_1 v_1 \|\proj_{\nabla\f(0)}\nabla S(\xt)\|_2 \sim
                \alpha_1 \|\proj_{\nabla\f(0)}\nabla S(\xt)\|_2 \left(
                2 \Beta\left( \frac{m-1}{2}, \frac{m-1}{2} \right) - 1
                \right).
            $$
            Observing that this is a symmetric distribution centered at $0$, we have
            $$
                \langle \nabla S(\xt), \nabla \f(0)\cdot u \rangle \sim 
                \alpha_1 \|\proj_{\nabla\f(0)}\nabla S(\xt)\|_2 \left(
                2 \Beta\left( \frac{m-1}{2}, \frac{m-1}{2} \right) - 1
                \right),
            $$
            which proves the first part of the lemma.
            
            For the second part, hereinafter, we condition the distribution of $u$ on $\langle \nabla S(\xt), \nabla \f(0) \cdot u \rangle = c$.
            According to \Cref{eq:lem-a-2-pf-1}, the condition means that
            $$
                v_1 = c_1 := \dfrac{c}{\alpha_1\|\proj_{\nabla\f(0)}\nabla S(\xt)\|_2} \Big\langle \dfrac{\proj_{\nabla\f(0)}\nabla S(\xt)}{\|\proj_{\nabla\f(0)}\nabla S(\xt)\|_2} , \mU_{:,1} \Big\rangle.
            $$
            Here we define the constant $c_1$.
            Since $v \sim \Unif(S^{m-1})$, it means that under this condition,
            $v' = (v_2,v_3,\cdots,v_m)$ is uniformly sampled from the $(m-1)$-dimension hypersphere with radius
            $
              r = \sqrt{1 - c_1^2}.
            $
            Therefore,
            \begin{equation}
                \begin{aligned}
                    & \E [\nabla\f(0) \cdot u \,|\, \langle \nabla S(\xt), \nabla \f(0)\cdot u \rangle = c]
                    = \E [ \mU\mSigma v \,|\, v_1 = c_1 ] \\
                    = & \E \left[
                        \alpha_1 v_1 \mU_{:,1} + \sum_{i=2}^m \alpha_i v_i \mU_{:,i}
                        \,\Big|\,
                        v_1 = c_1
                    \right] = \alpha_1 c_1 \mU_{:,1} + \sum_{i=2}^m \E\left[ \alpha_i v_i \mU_{:,i} | v_1=c_1 \right] \\
                    = & c\dfrac{\proj_{\nabla\f(0)}\nabla S(\xt)}{\|\proj_{\nabla\f(0)}\nabla S(\xt)\|_2^2}  + \sum_{i=2}^m \alpha_i \mU_{:,i} \cdot \E[v_i | v_1 = c_1]. 
                \end{aligned}
                \label{eq:lem-a-2-pf-2}
            \end{equation}
            Since under this condition, $v' = (v_2,\cdots, v_m)$ is uniformly sampled from a hypersphere centered at the origin~(with radius $r$), we have $\E[v_i|v_1=c_1]=0$ for $i \ge 2$ by the symmetry of Beta distribution.
            Thus,
            $$
                \E [\nabla\f(0) \cdot u \,|\, \langle \nabla S(\xt), \nabla \f(0)\cdot u \rangle = c] = c\dfrac{\proj_{\nabla\f(0)}\nabla S(\xt)}{\|\proj_{\nabla\f(0)}\nabla S(\xt)\|_2^2}.
            $$
        \end{proof}
        
        \begin{remark}
            The lemma considers the distribution of sampled vector $u$ after the transformation by $\f$ approximated in the first-order.
            The first equation of the lemma, \Cref{eq:lem-a-2-1}, reveals the distribution of the projection~(dot product) onto the true gradient direction.
            The distribution is a linearly scaled Beta distribution.
            The second equation of the lemma, \Cref{eq:lem-a-2-2}, reveals that the sampled vector is \emph{unbiased} on any direction orthogonal to the true gradient,
            i.e., conditioned on the same projected length on the true gradient direction, the expectation of the sampled vector aligns with the projected true gradient direction without any directional bias.
        \end{remark}
        
        According to \Cref{lem:nabla-f-decomposition}, we write $\nabla \f(0) = \mU\mSigma\mV^\T$.
        For notation simplicity, we let $\hat{\proj_{\nabla\f(0)} \nabla S}(\xt)$ denote the normalized true gradient: $\hat{\proj_{\nabla\f(0)} \nabla S}(\xt) := \proj_{\nabla\f(0)} \nabla S(\xt) / \|\proj_{\nabla\f(0)} \nabla S(\xt)\|_2$.
        Furthermore, we let $s := \langle \hat{\proj_{\nabla\f(0)} \nabla S}(\xt), \mU_{:,1} \rangle \in \{\pm 1\}$ denote the sign between these two aligned vectors.
        
        With respect to the randomness of $u \sim \Unif(S^{m-1})$, we denote $v$ to $\mV^\T u \sim \Unif(S^{m-1})$, and we define the following three events $E^-$, $E^o$, and $E^+$:
        \begin{align}
            E^-: & \langle \nabla S(\xt), \nabla \f(0) \cdot u \rangle \in (-\infty, -w), \label{eq:thm-1-pf-e-} \\
            E^o: & \langle \nabla S(\xt), \nabla \f(0) \cdot u \rangle \in [-w, +w], \label{eq:thm-1-pf-eo} \\
            E^+: & \langle \nabla S(\xt), \nabla \f(0) \cdot u \rangle \in (+w, +\infty). \label{eq:thm-1-pf-e+}
        \end{align}
        From \Cref{lem:a-2}, we denote $p$ to $\Pr[E^o]$, and by the symmetry of Beta distribution, $\Pr[E^-] = \Pr[E^+] = (1-p)/2$.
        From \Cref{eq:lem-a-2-pf-1}, we know $\langle \nabla S(\xt), \nabla\f(0)\cdot u \rangle = \alpha_1 \|\proj_{\nabla\f(0)}\nabla S(\xt)\|_2 s v_1$.
        Therefore, with events $E^-$ and $E^+$, $|v_1| > \frac{w}{\alpha_1 \|\proj_{\nabla\f(0)} \nabla S(\xt)\|_2}$ while the signs of $v_1$ are different between the two events; and with event $E^o$, $|v_1| \le \frac{w}{\alpha_1 \|\proj_{\nabla\f(0)} \nabla S(\xt)\|_2}$.
        
        According to the definition of the gradient estimator in \Cref{def:gradient-estimator}, we have $\E \tnablaS(\xt) = \E[ \phi(\xt + \Delta \f(\delta u)) \Delta \f(\delta u) ]$.
        According to \Cref{eq:lem-a-1-pf-1-taylor-f},
        for any unit vector $u$,
        $$
            \phi(\xt + \Delta \f(\delta u)) \Delta \f(\delta u) 
            = \phi(\xt + \Delta \f(\delta u)) \left( \delta \nabla \f(0) \cdot u + \xi_{\delta u} \right)
            = \delta \phi(\xt + \Delta \f(\delta u)) \nabla \f(0) \cdot u + \xi'_{\delta u}
        $$
        where $\xi_{\delta u}$ and $\xi'_{\delta u}$ are vectors depended by $\delta u$ with length $ \le \half \beta_{\f} \delta^2$.
        Therefore,
        \begin{equation}
            \| \E\tnablaS(\xt) - \delta \E[\phi(\xt + \nabla \f(\delta u)) \nabla \f(0) \cdot u] \|_2 \le \half \beta_{\f} \delta ^2.
            \label{eq:thm-1-pf-1}
        \end{equation}
        Now, we inspect $\E[\phi(\xt + \nabla\f(\delta u))\nabla \f(0)\cdot u]$:
        \begin{equation}
            \E[\phi(\xt + \nabla\f(\delta u))\nabla \f(0)\cdot u] 
            =
            \overbrace{
            p \E\left[\phi(\xt + \nabla\f(\delta u)) \nabla \f(0) \cdot u \,|\, E^o\right]
            }^{(*)}
            + 
            \overbrace{\dfrac{1-p}{2}
            \left(
                \E\left[-\nabla \f(0) \cdot u \,|\, E^-\right]
                +
                \E\left[\nabla \f(0) \cdot u \,|\, E^+\right]
            \right)}^{(**)}.
            \label{eq:thm-1-pf-2}
        \end{equation}
        According to \Cref{lem:a-2}~(\Cref{eq:lem-a-2-2}),
        $$
            \E[\nabla \f(0)\cdot u \,|\, E^+] = \E[|v_1| \,|\, E^+] \alpha_1 \hat{\proj_{\nabla\f(0)}\nabla S}(\xt),\,
            \E[\nabla \f(0)\cdot u \,|\, E^-] = \E[-|v_1| \,|\, E^-] \alpha_1 \hat{\proj_{\nabla\f(0)}\nabla S}(\xt).
        $$
        By symmetry of Beta distribution, $\E\left[|v_1| \,|\, E^+\right] = \E\left[|v_1| \,|\, E^-\right]$, and
        $$
            (**) = \dfrac{1-p}{2}
            \left(
                \E\left[-\nabla \f(0) \cdot u \,|\, E^-\right]
                +
                \E\left[\nabla \f(0) \cdot u \,|\, E^+\right]
            \right)
            =
            (1-p) \alpha_1 \E\left[|v_1| \,|\, E^+\right] \hat{\proj_{\nabla\f(0)}\nabla S}(\xt).
        $$
        For $(*)$, we notice that
        $$
            \begin{aligned}
                \E\left[\phi(\xt + \nabla\f(\delta u)) \nabla \f(0) \cdot u \,|\, E^o\right]
                & 
                \overset{\text{Eq.~\ref{eq:lem-a-2-pf-2}}}{=}
                \E\left[
                    \phi(\xt + \nabla \f(\delta u)) \left(\alpha_1 v_1 s \cdot  \hat{\proj_{\nabla\f(0)}\nabla S}(\xt) + \sum_{i=2}^m \alpha_i v_i \mU_{:,i} \right)
                    \,\Big|\,
                    E^o
                \right] \\
                & = \alpha_1 s \E\left[\phi(\xt + \nabla\f(\delta u)) v_1 \,|\, E^o\right] \hat{\proj_{\nabla\f(0)}\nabla S}(\xt) + \sum_{i=2}^m \alpha_i \E\left[\phi(\xt + \nabla\f(\delta u))v_i \,|\, E^o\right] \mU_{:,i}.
            \end{aligned}
        $$
        Combining them with \Cref{eq:thm-1-pf-2}, we have
        $$
            \begin{aligned}
                \E[\phi(\xt + \nabla\f(\delta u)) \nabla \f(0)\cdot u]
                =
                & 
                \alpha_1 \left(
                    ps \E\left[\phi(\xt + \nabla \f(\delta u))v_1 \,|\, E^o\right]
                    + 
                    (1-p) \E\left[ |v_1| \,|\, E^+ \right]
                \right)
                \hat{\proj_{\nabla\f(0)}\nabla S}(\xt) \\
                & + p\sum_{i=2}^m \alpha_i \E\left[ \phi(\xt + \nabla\f(\delta u)) v_i \,|\, E^o \right] \mU_{:,i}.
            \end{aligned}
        $$
        We notice that $\{ \hat{\proj_{\nabla\f(0)}\nabla S}(\xt), \mU_{:,2}, \cdots, \mU_{:,m} \}$ is an orthogonal basis of $\sR^n$.
        Thus,
        \begin{align*}
            & \big\| \E[\phi(\xt + \nabla\f(\delta u)) \nabla\f(0)\cdot u] - \alpha_1 \E\left[ |v_1| \right] \hat{\proj_{\nabla\f(0)}\nabla S}(\xt) \big\|_2 \\
            = & \Big\|  
                \alpha_1
                \left(
                    ps \E\left[\phi(\xt + \nabla \f(\delta u))v_1 \,|\, E^o\right]
                    + 
                    (1-p) \E\left[ |v_1| \,|\, E^+ \right]
                    - \E\left[ |v_1| \right]
                \right) \hat{\proj_{\nabla\f(0)}\nabla S}(\xt) \\
            & \hspace{2em}
                +
                p\sum_{i=2}^m \alpha_i \E\left[ \phi(\xt + \nabla\f(\delta u)) v_i \,|\, E^o \right] \mU_{:,i}
            \Big\|_2 \\
            = & \sqrt{
                \alpha_1^2 
                \underbrace{\left(
                    ps \E\left[\phi(\xt + \nabla \f(\delta u))v_1 \,|\, E^o\right]
                    + 
                    (1-p) \E\left[ |v_1| \,|\, E^+ \right]
                    - \E\left[ |v_1| \right]
                \right)^2}_{\text{(I)}}
                + \underbrace{\sum_{i=2}^m \alpha_i^2 p^2 \E\left[ \phi(\xt + \nabla\f(\delta u)) v_i \,|\, E^o \right]^2}_{\text{(II)}}
            }.
        \end{align*}
        We bound the two terms (I) and (II) individually.
        For the first term, we notice that
        \begin{align*}
            & \big|
                ps \E\left[\phi(\xt + \nabla \f(\delta u))v_1 \,|\, E^o\right]
                + 
                (1-p) \E\left[ |v_1| \,|\, E^+ \right]
                - \E\left[ |v_1| \right]
            \big| \\
            \le & p\big| \E\left[\phi(\xt + \nabla \f(\delta u))v_1 \,|\, E^o \right] \big|
            + \big| (1-p) \E\left[|v_1| \,|\, E^+\right] - (1-p)\E\left[|v_1| \,|\, E^+\right] - p\E\left[|v_1| \,|\, E^o\right] \big| \\
            \le & p\E\left[|v_1| \,|\, E^o\right] + p \E\left[|v_1| \,|\, E^o\right] = 2p\E\left[|v_1| \,|\, E^o\right] \le \dfrac{2pw}{\alpha_1 \|\proj_{\nabla\f(0)}\nabla S(\xt)\|_2}.
        \end{align*}
        For the second term,
        we have
        $$
            \text{(II)} = p^2 \sum_{i=2}^m \alpha_i^2 \E\left[ \phi(\xt + \nabla \f(\delta u)) v_i \,|\, E^o \right]^2
            \le p^2 \sum_{i=2}^m \alpha_i^2 \E\left[ |v_i| \,|\, E^o \right]^2
            \le p^2 \sum_{i=2}^m \alpha_i^2 \E\left[ v_i^2 \,|\, E^o \right].
        $$
        Since $v = (v_1,v_2,\dots,v_m)^\T$ is uniformly sampled from $S^{m-1}$,
        conditioned on every sampled $v_1$, the vector $(v_2,\dots,v_m)^\T$ is uniformly sampled from the $\sqrt{1-v_1^2}S^{m-2}$.
        We let $v' = (v_2',\dots,v_m')^\T$ be uniformly sampled from $S^{m-2}$.
        Thus, for every sampled $v_1$, we always have $\E\left[v_i^2 \,|\, v_1\right] \le \E\left[{v'_i}^2\right]$ because $v'$ is sampled from a larger hypersphere.
        By stacking all sampled $v_1$'s that forms the event $E^o$, we have
        $
            \text{(II)} \le p^2 \sum_{i=2}^m \alpha_i^2 \E\left[{v'_i}^2\right].
        $
        According to \cite{yang2020randomized}, $\frac{1 + v_i'}{2} \sim \Beta\left( \frac{m}{2} - 1, \frac{m}{2} - 1\right)$, whose variance $\Var\left(\frac{1+v_i'}{2}\right) = \frac{1}{4(m-1)}$.
        Since $\Var\left(\frac{1+v_i'}{2}\right) = \E \left[ \left( \frac{1+v_i'}{2} \right)^2 \right] - \E \left[ \frac{1+v_i'}{2} \right]^2 = \frac{1}{4} \E\left[{v_i'}^2\right]$, we have $\E\left[{v_i'}^2\right] = \frac{1}{m-1}$.
        Thus,
        $$
            \text{(II)}  \le p^2 \sum_{i=2}^m \alpha_i^2 \E\left[ {v_i'}^2 \right] = \frac{1}{m-1} p^2 \sum_{i=2}^m \alpha_i^2.
        $$
        As a result,
        \begin{equation}
             \big\| \E[\phi(\xt + \Delta \f(\delta u)) \nabla\f(0)\cdot u] - \alpha_1 \E\left[ |v_1| \right] \hat{\proj_{\nabla\f(0)}\nabla S}(\xt) \big\|_2
            \le \sqrt{
                \dfrac{4p^2w^2}{\|\proj_{\nabla\f(0)}\nabla S(\xt)\|_2^2} + \dfrac{p^2\sum_{i=2}^m \alpha_i^2}{m-1}
            }. \label{eq:thm-1-pf-3}
        \end{equation}
        
        Combining \Cref{eq:thm-1-pf-3} with \Cref{eq:thm-1-pf-1}, we have
        \begin{equation}
            \| \E\tnablaS(\xt) - \delta \alpha_1 \E\left[|v_1|\right] \hat{\proj_{\nabla\f(0)}\nabla S}(\xt) \|_2 
            \le p \delta \sqrt{
                \dfrac{4w^2}{\|\proj_{\nabla\f(0)}\nabla S(\xt)\|_2^2} + \dfrac{\sum_{i=2}^m \alpha_i^2}{m-1}
            } + \half \beta_{\f} \delta^2.
        \end{equation}
        It means
        \begin{equation}
            \begin{aligned}
                & \cos \langle \E \tnablaS(\xt), \proj_{\nabla\f(0)}\nabla S(\xt) \rangle \\
                = & \cos \big\langle \E \tnablaS(\xt),\, \delta \alpha_1 \E\left[|v_1|\right] \hat{\proj_{\nabla\f(0)}\nabla S}(\xt) \big\rangle \\
                \ge &
                1 - \dfrac{1}{2} \left( 
                    \dfrac{
                        p \delta \sqrt{
                            \frac{4w^2}{\| \proj_{\nabla\f(0)}\nabla S(\xt)\|_2^2} + \frac{\sum_{i=2}^m \alpha_i^2}{m-1}
                        } + \frac{1}{2} \beta_{\f} \delta^2
                    }{
                        \delta \alpha_1 \E\left[|v_1|\right]
                    }
                \right)^2 \\
                \ge &
                1 - \dfrac{1}{2}\left( \dfrac{2pw}{\alpha_1 \E\left[|v_1|\right] \| \proj_{\nabla\f(0)}\nabla S(\xt)\|_2} 
                + \dfrac{p}{\alpha_1 \E\left[|v_1|\right]} \sqrt{\dfrac{\sum_{i=2}^m \alpha_i^2}{m-1}}
                + \dfrac{\delta \beta_{\f}}{2\alpha_1 \E\left[|v_1|\right]}
                \right)^2.
            \end{aligned}
            \label{eq:thm-1-pf-4}
        \end{equation}
        
        To this point, we need to unfold $p$ and $\E\left[|v_1|\right]$.
        From the definition of event $E^o$(\Cref{eq:thm-1-pf-eo}), 
        $$
            p = \Pr\left[ -w/(\alpha_1\| \proj_{\nabla\f(0)}\nabla S(\xt)\|_2) \le v_1 \le w/(\alpha_1\| \proj_{\nabla\f(0)}\nabla S(\xt)\|_2) \right] = \Pr\left[ v_1^2 \le w^2 / (\alpha_1^2 \| \proj_{\nabla\f(0)}\nabla S(\xt)\|^2)\right],
        $$ where $v_1^2 \sim \Beta(1/2, (m-1)/2)$~\cite{chen2020hopskipjumpattack}.
        Thus, let $B(\cdot,\cdot)$ be the Beta function,
        \begin{equation}
            p = \int_{0}^{\frac{w^2}{\alpha_1^2\| \proj_{\nabla\f(0)}\nabla S(\xt)\|_2^2}}
            \dfrac{x^{-1/2} (1-x)^{\frac{m-3}{2}}}{B\left(\frac{1}{2}, \frac{m-1}{2}\right)}\dif x
            \le \dfrac{2w}{B(\frac{1}{2},\frac{m-1}{2})\alpha_1\| \proj_{\nabla\f(0)}\nabla S(\xt)\|_2}.
            \label{eq:thm-1-pf-p-eq}
        \end{equation}
        Also, from \cite{li2020nolinear}~(Lemma~1), we have
        \begin{equation}
            \E\left[|v_1|\right] = 2\int_{0}^1 \dfrac{x(1-x^2)^{\frac{m-3}{2}}}{B(\frac{1}{2},\frac{m-1}{2})}\dif x = \dfrac{2}{(m-1)\cdot B(\frac{1}{2}, \frac{m-1}{2})}.
            \label{eq:thm-1-pf-ev1-eq}
        \end{equation}
        Plugging them into \Cref{eq:thm-1-pf-4}:
        \begin{equation}
            \begin{aligned}
                & \cos\langle \E\tnablaS(\xt),  \proj_{\nabla\f(0)}\nabla S(\xt) \rangle \ge \\
                & \hspace{2em}
                1 - 
                \dfrac{1}{2}
                \left(
                    \dfrac{m-1}{\alpha_1^2 \| \proj_{\nabla\f(0)}\nabla S(\xt)\|_2}
                    \left( 
                        \dfrac{2w^2}{\| \proj_{\nabla\f(0)}\nabla S(\xt)\|_2} + w \sqrt{\dfrac{\sum_{i=2}^m \alpha_i^2}{m-1}}
                    \right)
                    +
                    \dfrac{\delta \beta_{\f} (m-1) B(\frac{1}{2}, \frac{m-1}{2})}{4\alpha_1}
                \right)^2.
            \end{aligned}
        \end{equation}
        We apply Stirling's approximation with error bound to $(m-1)B(\frac{1}{2}, \frac{m-1}{2})$:
        $$
            \sqrt{2\pi(m-1)} \le (m-1)B\left(\frac{1}{2}, \frac{m-1}{2}\right) \le 1.26 \sqrt{2\pi(m-1)}
            \text{ for all $m \ge 1$},
        $$
        and plug $w$~(\Cref{eq:w-in-lem-a-1}) in~(we also replace $\|\nabla\f(0)\|_2$ by $\max_{i\in[m]} \alpha_i$):
        \begin{equation}
            \begin{aligned}
                & \cos\langle \E\tnablaS(\xt), \proj_{\nabla\f(0)}\nabla S(\xt) \rangle \\
                \ge &
                1 - 
                \dfrac{1}{2}
                \left(
                    \dfrac{m-1}{\alpha_1^2 \|\proj_{\nabla\f(0)}\nabla S(\xt)\|_2}
                    \left( 
                        \dfrac{2w^2}{\|\proj_{\nabla\f(0)}\nabla S(\xt)\|_2} + w \sqrt{\dfrac{\sum_{i=2}^m \alpha_i^2}{m-1}}
                    \right)
                    +
                    0.315 \sqrt{2\pi(m-1)} \dfrac{\delta \beta_{\f}}{\alpha_1}
                \right)^2 \\
                = &
                1 - 
                \dfrac{(m-1)^2\delta^2}{8\alpha_1^2}
                \left(
                    \dfrac{\delta}{\alpha_1}
                    \left(
                        \dfrac{\beta_{\f}\|\proj_{\nabla\f(0)}\nabla S(\xt)\|_2 + \beta_S \left(\max_{i\in [m]} \alpha_i + \half \delta \beta_{\f}\right)^2}{\|\proj_{\nabla\f(0)}\nabla S(\xt)\|_2}
                    \right)^2 \right. \\
                & \hspace{2em}
                \left.
                    +
                    \dfrac{1}{\alpha_1}
                    \sqrt{\dfrac{\sum_{i=2}^m \alpha_i^2}{m-1}}
                    \dfrac{\beta_{\f}\|\proj_{\nabla\f(0)}\nabla S(\xt)\|_2 + \beta_S \left(\max_{i\in [m]} \alpha_i + \half \delta \beta_{\f}\right)^2}{\|\proj_{\nabla\f(0)}\nabla S(\xt)\|_2}
                    +
                    0.63 \sqrt{\dfrac{2\pi}{m-1}} \beta_{\f}
                \right)^2 \\
                \ge & 1 - \dfrac{(m-1)^2\delta^2}{8\alpha_1^2} \left( \dfrac{\delta\gamma^2}{\alpha_1} + \dfrac{\gamma}{\alpha_1} \sqrt{\dfrac{\sum_{i=2}^m \alpha_i^2}{m-1}} + \sqrt{\dfrac{1}{m-1}} 1.58  \beta_{\f} \right)^2.
            \end{aligned}
        \end{equation}
        According to \Cref{lem:cosine-similarity-projected-subspace}, we have
        \begin{equation}
            \cos\langle \E\tnablaS(\xt), \nabla S(\xt) \rangle
            \ge 
            \dfrac{\|\proj_{\nabla\f(0)}\nabla S(\xt)\|_2}{\|\nabla S(\xt)\|_2} \cdot \left( 1 - \dfrac{(m-1)^2\delta^2}{8\alpha_1^2} \left( \dfrac{\delta\gamma^2}{\alpha_1} + \dfrac{\gamma}{\alpha_1} \sqrt{\dfrac{\sum_{i=2}^m \alpha_i^2}{m-1}} + \sqrt{\dfrac{1}{m-1}} 1.58  \beta_{\f} \right)^2 \right).
        \end{equation}
    \end{proof}
    
    \subsection{Warmup: Concentration Bound at Boundary}
    \label{adxsubsec:proof-thm-at-boundary-tail}
    
    Now we consider the concentration bound for the boundary point gradient estimation.
    
    \begin{theorem}[Concentration of cosine similarity; at boundary]
        Under the same setting as \Cref{thm:at-boundary-expectation}, over the randomness of the sampled vector $\{u_b\}_{i=1}^B$, with probability $1-p$,
        \begin{equation}
            \begin{aligned}
                & \cos\langle \tnablaS(\xt), \nabla S(\xt) \rangle \\
                \ge &
                \dfrac{\|\proj_{\nabla\f(0)}\nabla S(\xt)\|_2}{\|\nabla S(\xt)\|_2}
                \cdot 
                \left(
                1 - \dfrac{(m-1)^2\delta^2}{8\alpha_1^2} \left( \dfrac{\delta\gamma^2}{\alpha_1} + \dfrac{\gamma}{\alpha_1} \sqrt{\dfrac{\sum_{i=2}^m \alpha_i^2}{m-1}} + \dfrac{1.58\beta_{\f} + \frac{1}{\delta} \sqrt{\sum_{i=1}^m \alpha_i^2} \cdot \sqrt{\frac{2}{B} \ln(\frac{m}{\epsilon})}}{\sqrt{m-1}} \right)^2
                \right),
            \end{aligned}
            \label{eq:thm2-1-adx}
        \end{equation}
        where
        \begin{equation}
            \gamma := \beta_{\f} + \dfrac{\beta_S \left(\max_{i\in [m]} \alpha_i + \half \delta \beta_{\f}\right)^2}{\|\nabla S(\xt)\|_2}.
            \label{eq:thm2-2-adx}
        \end{equation}
        \label{thm:at-boundary-tail}
    \end{theorem}
    
    \begin{proof}[Proof of \Cref{thm:at-boundary-tail}]
        The key idea for the proof is to bound the $\ell_2$ distance from the estimated gradient vector to the expectation of the estimated gradient vector via concentration bounds.
        Then, the bound is combined to the proof of \Cref{thm:at-boundary-expectation}, concretely, \Cref{eq:thm-1-pf-3}, to derive the required result.
        
        In the proof, to distinguish the ``$p$'' in probability bound from the definition in \Cref{eq:thm-1-pf-eo}, we change this variable to ``$\epsilon$'', i.e., the bound reads ``with probability $1 - \epsilon$, \dots''.
        
        We note that the gradient estimator per \Cref{def:gradient-estimator} is
        $$
            \tnablaS(\xt) = \dfrac{1}{B} \sum_{b=1}^B \phi(\xt + \Delta \f(\delta u_b)) \Delta f(\delta u_b)
        $$
        where $u_b \sim \Unif(S^{m-1})$.
        Thus, let $u \sim \Unif(S^{m-1})$,
        \begin{align*}
            & \big\|\tnablaS(\xt) - \delta \E\left[ \phi(\xt + \Delta\f(\delta u)) \nabla \f(0) \cdot u \right] \big\|_2 \\
            \overset{(i.)}{=} & \Big\|
                \dfrac{1}{B} \sum_{b=1}^B \phi(\xt + \Delta \f(\delta u_b)) \nabla \f(0) \cdot (\delta u_b) + \dfrac{1}{B} \sum_{b=1}^B \xi_{\delta u_b} - \delta \E\left[ \phi(\xt + \Delta\f(\delta u)) \nabla \f(0) \cdot u \right] \Big\|_2 \\
            \le & \half \beta_{\f} \delta^2 + \delta \Big\|
                \dfrac{1}{B} \sum_{b=1}^B \phi(\xt + \Delta \f(\delta u_b)) \nabla \f(0) \cdot u_b - \E\left[ \phi(\xt + \Delta\f(\delta u)) \nabla \f(0) \cdot u \right] \Big\|_2 \\
            = & \half \beta_{\f} \delta^2 + \delta \big\|
                \dfrac{1}{B} \sum_{b=1}^B \phi(\xt + \Delta \f(\delta u_b)) \mU \mSigma v_b -
                \E\left[ \phi(\xt + \Delta\f(\delta u)) \mU \mSigma v \right]
                \Big\|_2 & \text{(\Cref{lem:nabla-f-decomposition}, $v = \mV^\T u$)} \\
            = & \half \beta_{\f} \delta^2 + \delta \Big\|
                    \dfrac{1}{B} \sum_{b=1}^B \sum_{i=1}^m \phi(\xt + \Delta \f(\delta u_b)) \mU_{:,i} \alpha_i v_{b,i}
                    -
                    \sum_{i=1}^m \E\left[ \phi(\xt + \Delta \f(\delta u)) \mU_{:,i} \alpha_i v_i \right]
                \Big\|_2 \\
            = & \half \beta_{\f} \delta^2 + \delta \Big\|
                    \sum_{i=1}^m \alpha_i \mU_{:,i} \left(
                        \dfrac{1}{B} \sum_{b=1}^B \phi(\xt + \Delta \f(\delta u_b)) v_{b,i}
                        -
                        \E[\phi(\xt + \Delta\f(\delta u))v_i]
                    \right)
                \Big\|_2 \\
            \overset{(ii.)}{=} & \half \beta_{\f} \delta^2 + \delta \sqrt{
                \sum_{i=1}^m \alpha_i^2 \left(
                    \dfrac{1}{B} \sum_{b=1}^B \phi(\xt + \Delta \f(\delta u_b)) v_{b,i}
                    -
                    \E[\phi(\xt + \Delta\f(\delta u))v_i]
                \right)^2}.
        \end{align*}
        In $(i.)$, $\|\xi_{\delta u_b}\|_2 \le \half \beta_{\f} \delta^2$ from Taylor expansion as in \Cref{eq:lem-a-2-pf-1}.
        In $(ii.)$, $\mU_{:,i}$'s are orthogonal basis vectors.
        
        For each $i$, since the $\phi(\xt + \Delta \f(\delta u_b))v_{b,i}$'s for different $b$'s are independent, and within range $[-1, 1]$, we apply Hoeffding's inequality and yield
        $$
            \Pr\left[
                \Big|
                \dfrac{1}{B} \sum_{b=1}^B 
                \phi(\xt + \Delta \f(\delta u_b)) v_{b,i}
                -
                \E[\phi(\xt + \Delta\f(\delta u))v_i]
                \Big|
                \le
                \sqrt{\dfrac{2}{B}\ln\left(\frac{m}{\epsilon}\right)}
            \right] \ge 
            1 - \dfrac{\epsilon}{m}.
        $$
        From union bound, with probability $1 - \epsilon$, for any $i \in [m]$, we have
        $$
            \Big|
                \dfrac{1}{B} \sum_{b=1}^B 
                \phi(\xt + \Delta \f(\delta u_b)) v_{b,i}
                -
                \E[\phi(\xt + \Delta\f(\delta u))v_i]
            \Big|
            \le
            \sqrt{\dfrac{2}{B}\ln\left(\frac{m}{\epsilon}\right)}.
        $$
        Under this condition, we have
        \begin{equation}
            \big\|\tnablaS(\xt) - \delta \E\left[ \phi(\xt + \Delta\f(\delta u)) \nabla \f(0) \cdot u \right] \big\|_2
            \le \half \beta_{\f} \delta^2 + \delta \sqrt{\sum_{i=1}^m \alpha_i^2} \cdot \sqrt{\dfrac{2}{B}\ln\left(\frac{m}{\epsilon}\right)}.
            \label{eq:thm-2-pf-1}
        \end{equation}
        Note that a tighter concentration may be achieved by replacing the Hoeffding's inequality by other tailored tail bounds for Beta-distributed $v_i$'s.
        But due to the uncertainty brought by the sign term $\phi(\cdot)$, it is challenging.
        On the other hand, for these i.i.d. random variable's concentration, the Hoeffding's inequality is tight in terms of orders due to central limit theorem.
        
        Now, we combine \Cref{eq:thm-2-pf-1} with \Cref{eq:thm-1-pf-3} and get
        \begin{equation}
            \| \tnablaS(\xt) - \delta \alpha_1 \E\left[|v_1|\right] \hat{\proj_{\nabla \f(0)} \nabla S}(\xt) \|_2 \le p\delta \sqrt{
                \dfrac{4w^2}{\|\proj_{\nabla \f(0)} \nabla S(\xt)\|_2^2} + \dfrac{\sum_{i=2}^m \alpha_i^2}{m-1}
            } 
            + \delta \sqrt{\sum_{i=1}^m \alpha_i^2} \cdot \sqrt{\dfrac{2}{B}\ln\left(\frac{m}{\epsilon}\right)}
            + \half \beta_{\f}\delta^2,
            \label{eq:thm-2-pf-2}
        \end{equation}
        where $\hat{\nabla S}$ is the normalized tue gradient, $p := \Pr[E^o]$, and $E^o$ is as defined in \Cref{eq:thm-1-pf-eo}.
        Similar as \Cref{eq:thm-1-pf-4}:
        \begin{align*}
            \cos\langle \tnablaS(\xt), \proj_{\nabla \f(0)}\nabla S(\xt) \rangle 
            \ge
            1 - \dfrac{1}{2}
            \left(
                \dfrac{
                    p\delta \sqrt{
                        \frac{4w^2}{\|\proj_{\nabla \f(0)}\nabla S(\xt)\|_2^2} + \frac{\sum_{i=2}^m \alpha_i^2}{m-1}
                    } 
                    + \delta \sqrt{\sum_{i=1}^m \alpha_i^2} \cdot \sqrt{\frac{2}{B}\ln\left(\frac{m}{\epsilon}\right)}
                    + \frac{1}{2} \beta_{\f}\delta^2
                }{\delta \alpha_1 \E\left[|v_1|\right]}
            \right)^2.
        \end{align*}
        Following the similar process as in the proof of \Cref{thm:at-boundary-expectation}, we get
        \begin{equation}
            \begin{aligned}
                & \cos\langle \tnablaS(\xt), \nabla S(\xt) \rangle \\
                \ge &
                \dfrac{\|\proj_{\nabla\f(0)}\nabla S(\xt)\|_2}{\|\nabla S(\xt)\|_2}
                \cdot 
                \left(
                1 - \dfrac{(m-1)^2\delta^2}{8\alpha_1^2} \left( \dfrac{\delta\gamma^2}{\alpha_1} + \dfrac{\gamma}{\alpha_1} \sqrt{\dfrac{\sum_{i=2}^m \alpha_i^2}{m-1}} + \dfrac{1.58\beta_{\f} + \frac{1}{\delta} \sqrt{\sum_{i=1}^m \alpha_i^2} \cdot \sqrt{\frac{2}{B} \ln(\frac{m}{\epsilon})}}{\sqrt{m-1}} \right)^2
                \right).
            \end{aligned}
        \end{equation}
    \end{proof}
    
    \subsection{Main Result: Expectation Bound Near Boundary~(\texorpdfstring{\Cref{thm:approach-boundary-expectation}}{Theorem 1})}
    \label{adxsubsec:proof-thm-approach-boundary-expectation}
    
    \begin{customthm}{\ref{thm:approach-boundary-expectation}}[Expected cosine similarity; approaching boundary]
        The difference function $S$ and the projection $\f$ are as defined before.
        For a point $\xt$ that is $\theta$-close to the boundary, i.e., there exists $\theta' \in [-\theta,\theta]$ such that $S(\xt + \theta' \nabla S(\xt) / \|\nabla S(\xt)\|_2) = 0$,
        let estimated gradient $\tnablaS(\xt)$ be as computed by \Cref{def:gradient-estimator} with step size $\delta$ and sampling size $B$.
        Over the randomness of the sampled vectors $\{u_b\}_{i=1}^B$,
        \begin{equation}
            \begin{aligned}
                \cos\langle \E\tnablaS(\xt), \nabla S(\xt) \rangle
                \ge &
                \dfrac{\|\proj_{\nabla\f(0)}\nabla S(\xt)\|_2}{\|\nabla S(\xt)\|_2}
                \cdot \\
                & \hspace{2em}
                \left(
                    1 - \dfrac{(m-1)^2 \delta^2}{8\alpha_1^2}
                    \left(
                        \dfrac{\delta \gamma^2}{\alpha_1}
                        +
                        \dfrac{\gamma}{\alpha_1} \sqrt{\dfrac{\sum_{i=2}^m \alpha_i^2}{m-1}}
                        +
                        \dfrac{1.58\beta_{\f}}{\sqrt{m-1}}
                        +
                        \dfrac{\gamma\theta}{\alpha_1\delta}
                        \cdot
                        \dfrac{\|\nabla S(\xt)\|_2}{\|\proj_{\nabla\f(0)} \nabla S(\xt)\|_2}
                    \right)^2
                \right),
            \end{aligned}
        \end{equation}
        where
        \begin{equation}
            \gamma := \beta_{\f} +
            \dfrac{\beta_S \left(\max_{i\in [m]} \alpha_i + \half \delta \beta_{\f}\right)^2 + \beta_S \theta^2 / \delta^2 }{\|\proj_{\nabla\f(0)}\nabla S(\xt)\|_2}.
        \end{equation}
    \end{customthm}
    
    \begin{proof}[Proof of \Cref{thm:approach-boundary-expectation}]
        The high-level idea is similar to the proof of \Cref{thm:at-boundary-expectation}:
        we build the connection between $\langle \nabla S(\xt, \nabla \f(0)\cdot u \rangle$ and $\phi(\xt + \Delta \f(\delta u))$.
        Then, due to the unbiased sampling, the expectation of estimated gradient is close to the true gradient direction with bounded error.
        
        For simplicity, from the symmetry and monotonicity, we let $\theta' = \theta$, i.e., $S(\xt + \theta \nabla S(\xt) / \|\nabla S(\xt)\|_2 ) = 0$.
        
        \begin{lemma}
            \label{lem:a-3}
            We let
            \begin{equation}
                w := 
                \dfrac{1}{2}
                \delta
                \left(
                    \beta_{\f} \|\nabla S(\xt)\|_2
                    +
                    \beta_S \left( \|\nabla \f(0)\|_2 + \dfrac{1}{2} \delta \beta_{\f} \right)^2
                \right)
                + \dfrac{\beta_S \theta^2}{2\delta}
                .
                \label{eq:w-in-lem-a-3}
            \end{equation}
            On the point $\xt$ such that $S(\xt + \theta \nabla S(\xt) / \|\nabla S(\xt)\|_2) = 0$, for any $\delta > 0$ and unit vector $u \in \sR^n$,
            $$
                \begin{aligned}
                    \langle \nabla S(\xt),\, \nabla \f(0) \cdot u \rangle > w + \dfrac{\theta \|\nabla S(\xt)\|_2}{\delta} & \Longrightarrow \phi(\xt + \Delta\f(\delta u)) = 1, \\
                    \langle \nabla S(\xt),\, \nabla \f(0) \cdot u \rangle < -w + \dfrac{\theta \|\nabla S(\xt)\|_2}{\delta} & \Longrightarrow \phi(\xt + \Delta\f(\delta u)) = -1.
                \end{aligned}
            $$
        \end{lemma}
        
        \begin{proof}[Proof of \Cref{lem:a-3}]
            We do Taylor expansion on $S(\xt + \Delta \f(\delta u))$ at point $\xt$:
            $$
                S(\xt + \Delta \f(\delta u)) \in S(\xt) + \delta \left( \langle \nabla S(\xt),\nabla \f(0) \cdot u \rangle \pm \half \delta \left( \|\nabla S(\xt)\|_2 \beta_{\f} + \beta_S \left( \|\nabla \f(0) \|_2 + \half \delta \beta_{\f} \right)^2  \right) \right).
            $$
            Notice that $S(\xt + \theta \nabla S(\xt) / \|\nabla S(\xt)\|_2)$ can also be expanded at point $\xt$:
            $$
                0 = S(\xt + \theta \nabla S(\xt) / \|\nabla S(\xt)\|_2) \in S(\xt) + \theta \|\nabla S(\xt)\|_2 \pm \frac{1}{2} \beta_S \theta^2,
            $$
            i.e.,
            $$
                S(\xt) \in -\theta \|\nabla S(\xt)\|_2 \pm \frac{1}{2} \beta_S \theta^2.
            $$
            Therefore,
            $$
                S(\xt + \Delta\f(\delta u)) \in \delta \left(
                    \langle \nabla S(\xt), \nabla \f(0)\cdot u \rangle
                    - \dfrac{\theta\|\nabla S(\xt)\|_2}{\delta}
                    \pm \frac{1}{2} \delta \left( \|\nabla S(\xt)\|_2 \beta_{\f} + \beta_S \left( \|\nabla \f(0) \|_2 + \frac{1}{2} \delta \beta_{\f} \right)^2  \right) \pm \dfrac{\beta_S\theta^2}{2\delta} 
                \right).
            $$
            Noticing that $\phi(\xt + \Delta\f(\delta u)) = \sgn\left(S(\xt + \Delta \f(\delta u)\right)$, we conclude the proof. 
        \end{proof}
        
        According to \Cref{lem:nabla-f-decomposition}, we write $\Delta \f(0) = \mU\mSigma\mV^\T$.
        We let $\hat{\nabla S}(\xt)$ denote the normalized true gradient: $\hat{\nabla S}(\xt) := \nabla S(\xt) / \|\nabla S(\xt)\|_2$.
        Furthermore, we define $s := \langle \hat{\nabla S}(\xt), \mU_{:,1} \rangle \in \{\pm 1\}$, which is the sign between these two aligned vectors.
        
        With respect to the randomness of $u \sim \Unif(S^{m-1})$, we let $v$ denote $\mV^\T u \sim \Unif(S^{m-1})$, and we define the following three events $E^-$, $E^o$, and $E^+$ with probability $p^-, p^o$ and $p^+$ respectively:
        \begin{align}
            E^-: & \langle \nabla S(\xt), \nabla \f(0) \cdot u \rangle \in (-\infty, -w + \theta \|\nabla S(\xt)\|_2 / \delta), & p^- := \Pr[E^-] \label{eq:thm-3-pf-e-} \\
            E^o: & \langle \nabla S(\xt), \nabla \f(0) \cdot u \rangle \in [-w + \theta \|\nabla S(\xt)\|_2 / \delta, +w + \theta \|\nabla S(\xt)\|_2 / \delta], & p^o := \Pr[E^o] \label{eq:thm-3-pf-eo} \\
            E^+: & \langle \nabla S(\xt), \nabla \f(0) \cdot u \rangle \in (+w + \theta \|\nabla S(\xt)\|_2 / \delta, +\infty). & p^+ := \Pr[E^+] \label{eq:thm-3-pf-e+}
        \end{align}
        We notice that \Cref{lem:a-2} still holds since $u$ is still uniformly sampled from hypersphere $S^{m-1}$ and \Cref{lem:nabla-f-decomposition} still holds for $\nabla\f(0)$.
        Thus,
        $$
            \langle \nabla S(\xt), \nabla \f(0)\cdot u \rangle = \alpha_1 \|\proj_{\nabla\f(0)}\nabla S(\xt)\|_2 sv_1.
        $$
        And with event $E^o$, we have 
        \begin{equation}
            |v_1| \le \frac{w}{\alpha_1 \|\proj_{\nabla\f(0)}\nabla S(\xt)\|_2} + \frac{\theta}{\alpha_1\delta}
            \cdot
            \frac{\|\nabla S(\xt)\|_2}{\|\proj_{\nabla\f(0)}\nabla S(\xt)\|_2}.
            \label{eq:thm-3-pf-eo-v}
        \end{equation}
        
        Now, we can start to bound the error between expectation of estimated gradient and a scaled true gradient.
        As the first step, according to \Cref{eq:thm-1-pf-1},
        we have
        \begin{equation}
            \| \E \tnablaS(\xt) - \delta \E[\phi(\xt + \Delta \f(\delta u))\nabla \f(0) \cdot u] \|_2 \le \frac{1}{2} \beta_{\f} \delta^2.
            \label{eq:thm-3-pf-1}
        \end{equation}
        Then we decompose $\E[\phi(\xt + \Delta \f(\delta u))\nabla \f(0) \cdot u]$ to connect it with the (projected) true gradient:
        \begin{align}
            & \E[\phi(\xt + \Delta \f(\delta u))\nabla \f(0) \cdot u] \nonumber \\
            = & p^o \E[\phi(\xt + \Delta \f(\delta u)) \nabla \f(0) \cdot u \,|\, E^o]
            + p^+ \E[\nabla \f(0) \cdot u \,|\, E^+]
            + p^- \E[-\nabla \f(0) \cdot u \,|\, E^-] \nonumber \\
            = & p^o \sum_{i=1}^m \alpha_i \E\left[ \phi(\xt + \Delta\f(\delta u)) v_i \,|\, E^o\right] \mU_{:,i}
            + p^+ \alpha_1 \hat{\proj_{\nabla\f(0)}\nabla S}(\xt) s \E[v_1 \,|\, E^+]
            + p^- \alpha_1 \hat{\proj_{\nabla\f(0)}\nabla S}(\xt) s \E[-v_1 \,|\, E^-] \nonumber \\
            = & \alpha_1 s \left(
                p^o \E[\phi(\xt + \Delta \f(\delta u)) v_1 \,|\, E^o]
                + p^+ \E[v_1 \,|\, E^+]
                + p^- \E[-v_1 \,|\, E^-]
            \right) \hat{\proj_{\nabla\f(0)}\nabla S}(\xt) \nonumber \\
            & \hspace{2em} + p^o \sum_{i=2}^m \alpha_i \E[\phi(\xt + \Delta \f(\delta u))v_i \,|\, E^o] \mU_{:,i} \label{eq:thm-3-pf-2}.
        \end{align}
        We notice that
        \begin{align*}
            & \left| 
                s \left(
                    p^o \E[\phi(\xt + \Delta \f(\delta u)) v_1 \,|\, E^o]
                    + p^+ \E[v_1 \,|\, E^+]
                    + p^- \E[-v_1 \,|\, E^-]
                \right)
                -
                \E[|v_1|]
            \right| \\
            = & p^o \left| \E[\phi(\xt + \Delta \f(\delta u)) v_1 - |v_1| \,|\, E^o] \right| \le 2p^o \E[|v_1| \,|\, E^o] \\
            \le & 2p^o \left( \dfrac{w}{\alpha_1\|\proj_{\nabla\f(0)}\nabla S(\xt)\|_2} + \dfrac{\theta}{\alpha_1\delta} \cdot \dfrac{\|\nabla S(\xt)\|_2}{\|\proj_{\nabla\f(0)} \nabla S(\xt)\|_2} \right). & \text{(\Cref{eq:thm-3-pf-eo-v})}
        \end{align*}
        For any $i \ge 2$, we define a new vector $v' = (v_2', v_3', \cdots, v_m') \sim \Unif(S^{m-2})$.
        Thus,
        $$
            \E[\phi(\xt + \Delta\f(\delta u))v_i \,|\, E^o] \le \E[|v_i| \,|\, E^o] \le \E[|v_i'|].
        $$
        Combining the above equations to \Cref{eq:thm-3-pf-2}, we get
        \begin{align}
            & \big\| \E[\phi(\xt + \Delta \f(\delta u))\nabla \f(0) \cdot u] - \alpha_1\E\left[|v_1|\right]\hat{\proj_{\nabla\f(0)}\nabla S}(\xt) \big\|_2 \nonumber \\
            \le & \sqrt{
                4\alpha_1^2(p^o)^2 \left( \dfrac{w}{\alpha_1\|\proj_{\nabla\f(0)}\nabla S(\xt)\|_2} + \dfrac{\theta}{\alpha_1\delta} \cdot \dfrac{\|\nabla S(\xt)\|_2}{\|\proj_{\nabla\f(0)} \nabla S(\xt)\|_2} \right)^2 + (p^o)^2 \sum_{i=2}^m \alpha_i^2 \E[|v_i'|]^2
            } \nonumber \\
            \le & p^o \sqrt{
                4\alpha_1^2 \left( \dfrac{w}{\alpha_1\|\proj_{\nabla\f(0)}\nabla S(\xt)\|_2} + \dfrac{\theta}{\alpha_1\delta} \cdot \dfrac{\|\nabla S(\xt)\|_2}{\|\proj_{\nabla\f(0)} \nabla S(\xt)\|_2} \right)^2
                + \sum_{i=2}^{m} \dfrac{\alpha_i^2}{m-1}
            }. \label{eq:thm-3-pf-tobeusedinthm4-1}
        \end{align}
        Combining with \Cref{eq:thm-3-pf-1}:
        \begin{equation}
            \begin{aligned}
                & \big\| \E \tnablaS(\xt) - \delta \alpha_1 \E\left[ |v_1| \right] \hat{\proj_{\nabla\f(0)}\nabla S}(\xt) \big\|_2 \\
                \le & 
                \delta p^o \sqrt{
                    4\alpha_1^2 \left( \dfrac{w}{\alpha_1\|\proj_{\nabla\f(0)}\nabla S(\xt)\|_2} + \dfrac{\theta}{\alpha_1\delta} \cdot \dfrac{\|\nabla S(\xt)\|_2}{\|\proj_{\nabla\f(0)} \nabla S(\xt)\|_2} \right)^2
                    + \sum_{i=2}^{m} \dfrac{\alpha_i^2}{m-1}
                } 
                + \frac{1}{2} \beta_{\f} \delta^2.
            \end{aligned}
            \label{eq:thm-3-pf-3}
        \end{equation}
        Thus,
        \begin{equation}
            \begin{aligned}
                & \cos\langle \E\tnablaS(\xt), \proj_{\nabla\f(0)}\nabla S(\xt) \rangle \\
                \ge & 1 - \dfrac{1}{2}\left(
                    \dfrac{2p^o w}{\alpha_1 \E\left[|v_1|\right] \|\proj_{\nabla\f(0)}\nabla S(\xt)\|_2}
                    + \dfrac{2p^o \theta}{\alpha_1\delta \E\left[|v_1|\right]}
                     \cdot \dfrac{\|\nabla S(\xt)\|_2}{\|\proj_{\nabla\f(0)} \nabla S(\xt)\|_2}
                    + \dfrac{p^o}{\alpha_1 \E\left[|v_1|\right]} \sqrt{\dfrac{\sum_{i=2}^m \alpha_i^2}{m-1}}
                    + \dfrac{\delta \beta_{\f}}{2\alpha_1 \E\left[|v_1|\right]}
                \right)^2.
            \end{aligned}
            \label{eq:thm-3-pf-4}
        \end{equation}
        We notice that $v_1$ is distributed around $0$ with symmetry and concentration, so
        $$
            \begin{aligned}
                p^o & = \Pr[E^o] = \Pr\left[ sv_1 \in \left[
                -\frac{w}{\alpha_1 \|\proj_{\nabla\f(0)}\nabla S(\xt)\|_2} + \frac{\theta}{\alpha_1\delta} \cdot \dfrac{\|\nabla S(\xt)\|_2}{\|\proj_{\nabla\f(0)} \nabla S(\xt)\|_2}, \right. \right. \\
                & \hspace{3em}
                \left. \left. \frac{w}{\alpha_1 \|\proj_{\nabla\f(0)}\nabla S(\xt)\|_2} + \frac{\theta}{\alpha_1\delta} \cdot \dfrac{\|\nabla S(\xt)\|_2}{\|\proj_{\nabla\f(0)} \nabla S(\xt)\|_2}
                \right]
                \right] \\
                \le & \Pr\left[
                    v_1^2 \le \frac{w^2}{\alpha_1^2 \|\proj_{\nabla\f(0)}\nabla S(\xt)\|_2^2} 
                \right] 
                \overset{\text{Eq.\ref{eq:thm-1-pf-p-eq}}}{\le} \dfrac{2w}{B(\frac 1 2, \frac{m-1}{2}) \alpha_1 \|\proj_{\nabla\f(0)}\nabla S(\xt)\|_2}.
            \end{aligned}
        $$
        And from \Cref{eq:thm-1-pf-ev1-eq},
        $$
            \E\left[|v_1|\right] = \dfrac{2}{(m-1)\cdot B(\frac 1 2, \frac{m-1}{2})}.
        $$
        Insert them into \Cref{eq:thm-3-pf-4}, we get
        \begin{equation}
            \begin{aligned}
                \cos\langle \E\tnablaS(\xt), \nabla S(\xt) \rangle
                \ge &
                \dfrac{\|\proj_{\nabla\f(0)}\nabla S(\xt)\|_2}{\|\nabla S(\xt)\|_2}
                \cdot \\
                & \hspace{2em}
                \left(
                    1 - \dfrac{(m-1)^2 \delta^2}{8\alpha_1^2}
                    \left(
                        \dfrac{\delta \gamma^2}{\alpha_1}
                        +
                        \dfrac{\gamma}{\alpha_1} \sqrt{\dfrac{\sum_{i=2}^m \alpha_i^2}{m-1}}
                        +
                        \dfrac{\gamma\theta}{\alpha_1\delta}
                        \cdot
                        \dfrac{\|\nabla S(\xt)\|_2}{\|\proj_{\nabla\f(0)} \nabla S(\xt)\|_2}
                        +
                        \dfrac{1.58\beta_{\f}}{\sqrt{m-1}}
                    \right)^2
                \right),
            \end{aligned}
            \label{eq:thm-3-pf-5}
        \end{equation}
    \end{proof}
    
    \subsection{Main Result: Concentration Bound Near Boundary~(\texorpdfstring{\Cref{thm:approach-boundary-concentration}}{Theorem 4})}
    \label{adxsubsec:proof-thm-approach-boundary-concentration}
    
    \begin{customthm}{\ref{thm:approach-boundary-concentration}}[Concentration of cosine similarity; approaching boundary]
        Under the same setting as \Cref{thm:approach-boundary-expectation}, over the randomness of the sampled vector $\{u_b\}_{i=1}^B$, with probability $1-p$,
        Under the same setting as \Cref{thm:approach-boundary-expectation}, over the randomness of the sampled vector $\{u_b\}_{i=1}^B$, with probability $1-p$,
        \begin{equation}
            \begin{aligned}
                & \cos\langle \E\tnablaS(\xt), \nabla S(\xt) \rangle
                \ge 
                \dfrac{\|\proj_{\nabla\f(0)}\nabla S(\xt)\|_2}{\|\nabla S(\xt)\|_2}
                \cdot \\
                &
                \left(
                    1 - \dfrac{(m-1)^2 \delta^2}{8\alpha_1^2}
                    \left(
                        \dfrac{\delta \gamma^2}{\alpha_1}
                        +
                        \dfrac{\gamma}{\alpha_1} \sqrt{\dfrac{\sum_{i=2}^m \alpha_i^2}{m-1}}
                        +
                        \dfrac{1.58\beta_{\f}}{\sqrt{m-1}}
                        +
                        \dfrac{\gamma\theta}{\alpha_1\delta}
                        \cdot
                        \dfrac{\|\nabla S(\xt)\|_2}{\|\proj_{\nabla\f(0)} \nabla S(\xt)\|_2}
                        +
                        \dfrac{\frac{1}{\delta} \sqrt{\sum_{i=1}^m \alpha_i^2} \cdot \sqrt{\frac{2}{B} \ln(\frac{m}{p})}}{\sqrt{m-1}}
                    \right)^2
                \right),
            \end{aligned}
        \end{equation}
        where
        \begin{equation}
            \gamma := \beta_{\f} +
            \dfrac{\beta_S \left(\max_{i\in [m]} \alpha_i + \half \delta \beta_{\f}\right)^2 + \beta_S \theta^2 / \delta^2 }{\|\proj_{\nabla\f(0)}\nabla S(\xt)\|_2}.
        \end{equation}
    \end{customthm}
    
    \begin{proof}[Proof of \Cref{thm:approach-boundary-concentration}]
        The proof follows the similar way as how we extend \Cref{thm:at-boundary-expectation} to \Cref{thm:at-boundary-tail}.
        
        Similar as the proof in \Cref{thm:at-boundary-tail}, by applying Hoefdding bound, with probability $1 - \epsilon$, 
        $$
            \big\| \tnablaS(\xt) - \delta \E\left[ \phi(\xt + \Delta\f(\delta u)) \nabla \f(0) \cdot u\right] \big\|_2 \le 
            \half \beta_{\f} \delta^2
            +
            \delta\sqrt{\sum_{i=1}^m \alpha_i^2} \cdot \sqrt{\dfrac{2}{B} \ln\left(\frac{m}{\epsilon}\right)}.
        $$
        When this holds, combining it with \Cref{eq:thm-3-pf-tobeusedinthm4-1}, we get
        $$
            \begin{aligned}
                & \big\| \tnablaS(\xt) - \delta \E\left[ \phi(\xt + \Delta\f(\delta u)) \nabla \f(0) \cdot u\right] \big\|_2 \\
                \le &
                \frac 1 2 \beta_{\f} \delta^2
                +
                \delta\sqrt{\sum_{i=1}^m \alpha_i^2} \cdot \sqrt{\dfrac{2}{B} \ln\left(\frac{m}{\epsilon}\right)}
                +
                \delta p^o
                \sqrt{
                    4\alpha_1^2 \left( \dfrac{w}{\alpha_1\|\proj_{\nabla\f(0)}\nabla S(\xt)\|_2} + \dfrac{\theta}{\alpha_1\delta} 
                        \cdot
                        \dfrac{\|\nabla S(\xt)\|_2}{\|\proj_{\nabla\f(0)} \nabla S(\xt)\|_2} \right)^2
                    + \sum_{i=2}^{m} \dfrac{\alpha_i^2}{m-1}
                },
            \end{aligned}
        $$
        where $p^o$ is as defined in \Cref{eq:thm-3-pf-eo}.
        Thus,
        \begin{align*}
            & \cos\langle \tnablaS(\xt), \proj_{\nabla \f(0)} \nabla S(\xt) \rangle \\
            \ge & 1 - \dfrac{1}{2}\left(
                    \dfrac{2p^o w}{\alpha_1 \E\left[|v_1|\right] \|\proj_{\nabla\f(0)}\nabla S(\xt)\|_2}
                    + \dfrac{2p^o \theta}{\alpha_1\delta \E\left[|v_1|\right]}
                    \cdot
                        \dfrac{\|\nabla S(\xt)\|_2}{\|\proj_{\nabla\f(0)} \nabla S(\xt)\|_2}
                    \right. \\
            & \hspace{2em} \left.
                    + \dfrac{p^o}{\alpha_1 \E\left[|v_1|\right]} \sqrt{\dfrac{\sum_{i=2}^m \alpha_i^2}{m-1}}
                    + \dfrac{\delta \beta_{\f}}{2\alpha_1 \E\left[|v_1|\right]}
                    + \dfrac{\sqrt{\sum_{i=1}^m \alpha_i^2} \cdot \sqrt{\frac{2}{B}\ln\left(\frac{m}{\epsilon}\right)}}{\alpha_1 \E\left[ |v_1| \right]}
                \right)^2 \\
            \ge & 1 - \dfrac{(m-1)^2\delta^2}{8\alpha_1^2} \left( \dfrac{\delta\gamma^2}{\alpha_1} + \dfrac{\gamma}{\alpha_1} \sqrt{\dfrac{\sum_{i=2}^m \alpha_i^2}{m-1}}
                 + \dfrac{1.58\beta_{\f}}{\sqrt{m-1}} + \dfrac{\gamma\theta}{\alpha_1\delta} \cdot
                    \dfrac{\|\nabla S(\xt)\|_2}{\|\proj_{\nabla\f(0)} \nabla S(\xt)\|_2}
                + \dfrac{\frac{1}{\delta} \sqrt{\sum_{i=1}^m \alpha_i^2} \cdot \sqrt{\frac{2}{B} \ln(\frac{m}{\epsilon})}}{\sqrt{m-1}}
            \right)^2.
        \end{align*}
        We conclude the proof by observing that
        $$
            \cos\langle \tnablaS(\xt), \nabla S(\xt) \rangle = \dfrac{\|\proj_{\nabla\f(0)} \nabla S(\xt)\|_2}{\|\nabla S(\xt)\|_2} \cos\langle \tnablaS(\xt), \proj_{\nabla\f(0)} \nabla S(\xt) \rangle
        $$
        from \Cref{lem:cosine-similarity-projected-subspace}.
    \end{proof}
    
    \subsection{Bound in Big-$\gO$ Notation}
        We mainly simplify the bound in \Cref{thm:approach-boundary-expectation,thm:approach-boundary-concentration} by omitting the terms with smaller orders of $m$.
        In boundary attack, following HSJA~\cite{chen2017zoo}, we set binary search precision $\theta = (m\sqrt m)^{-1}$ and step size $\delta_t = \|\xt - x^*\|_2 / m = \Theta(1/m)$.
        Therefore, $\theta / \delta = \Theta(1/\sqrt{m})$.
        We first simplify $\gamma$ in \Cref{thm:approach-boundary-expectation,thm:approach-boundary-concentration}:
        $$
        \begin{aligned}
            \gamma & = \beta_{\f} +
            \dfrac{\beta_S \left(\max_{i\in [m]} \alpha_i + \half \delta \beta_{\f}\right)^2 + \beta_S \theta^2 / \delta^2 }{\|\proj_{\nabla\f(0)}\nabla S(\xt)\|_2} \\
            & = \beta_{\f} + 
            \dfrac{\beta_S \left(\max_{i\in [m]} \alpha_i + \Theta(\beta_{\f}/m)\right)^2 + \beta_S \Theta(1/m)
            }{\|\proj_{\nabla\f(0)}\nabla S(\xt)\|_2}
            = \beta_{\f} + \gO\left( \dfrac{\beta_S \max_{i\in [m]} \alpha_i^2}{\|\proj_{\nabla\f(0)}\nabla S(\xt)\|_2} \right).
        \end{aligned}
        $$
        We plug in the $\gamma$ into \Cref{thm:approach-boundary-concentration}:
        \begin{align*}
            & \dfrac{\delta \gamma^2}{\alpha_1}
            +
            \dfrac{\gamma}{\alpha_1} \sqrt{\dfrac{\sum_{i=2}^m \alpha_i^2}{m-1}}
            +
            \dfrac{1.58\beta_{\f}}{\sqrt{m-1}}
            +
            \dfrac{\gamma\theta}{\alpha_1\delta}
            \cdot
            \dfrac{\|\nabla S(\xt)\|_2}{\|\proj_{\nabla\f(0)} \nabla S(\xt)\|_2}
            +
            \dfrac{\frac{1}{\delta} \sqrt{\sum_{i=1}^m \alpha_i^2} \cdot \sqrt{\frac{2}{B} \ln(\frac{m}{p})}}{\sqrt{m-1}} \\
            = & 
            \gO\left( \dfrac{\beta_{\f}^2}{m\alpha_1} \right)
            + \gO\left( \dfrac{\beta_S^2 \max_{i\in [m]}\alpha_i^4}{m\alpha_1 \|\proj_{\nabla\f(0)} \nabla S(\xt)\|_2^2} \right)
            + \gO\left(\dfrac{\beta_{\f}}{\alpha_1} \sqrt{\dfrac{\sum_{i=2}^m \alpha_i^2}{m-1}}\right)
            + \gO\left( \dfrac{\beta_S \max_{i\in [m]} \alpha_i^2}{\alpha_1\|\proj_{\nabla\f(0)}\nabla S(\xt)\|_2} \sqrt{\dfrac{\sum_{i=2}^m \alpha_i^2}{m-1}} \right) \\
            & + \gO\left(\dfrac{\beta_{\f}}{\sqrt{m}}\right)
            + \gO\left(  \dfrac{\beta_{\f} \|\nabla S(\xt)\|_2}{m\alpha_1} \right)
            + \gO\left( \dfrac{\beta_S \max_{i\in [m]} \alpha_i^2 \|\nabla S(\xt)\|_2}{m\alpha_1\|\proj_{\nabla\f(0)}\nabla S(\xt)\|_2}  \right)
            + \gO\left( \dfrac{1}{\delta}  \dfrac{\alpha_1 \sqrt{\frac{2}{B} \ln(\frac{m}{p})}}{\sqrt{m-1}}\right) \\
            & 
            + \gO\left( \dfrac{1}{\delta} \sqrt{\frac{2}{B} \ln\left(\frac{m}{p}\right)} \cdot\sqrt{\dfrac{\sum_{i=2}^m \alpha_i^2}{m-1}}\right).
        \end{align*}
        We can discard all terms with negative order of $m$~(since they will be negligible with respect to other terms when $m$ is not too small) and get
        \begin{align*}
            & \dfrac{\delta \gamma^2}{\alpha_1}
            +
            \dfrac{\gamma}{\alpha_1} \sqrt{\dfrac{\sum_{i=2}^m \alpha_i^2}{m-1}}
            +
            \dfrac{1.58\beta_{\f}}{\sqrt{m-1}}
            +
            \dfrac{\gamma\theta}{\alpha_1\delta}
            \cdot
            \dfrac{\|\nabla S(\xt)\|_2}{\|\proj_{\nabla\f(0)} \nabla S(\xt)\|_2}
            +
            \dfrac{\frac{1}{\delta} \sqrt{\sum_{i=1}^m \alpha_i^2} \cdot \sqrt{\frac{2}{B} \ln(\frac{m}{p})}}{\sqrt{m-1}} \\
            = & \gO\left( \dfrac{\beta_{\f}}{\alpha_1} \sqrt{\dfrac{\sum_{i=2}^m \alpha_i^2}{m-1}} \right)
            + \gO \left( \dfrac{\beta_S \max_{i\in[m]} \alpha_i^2}{\alpha_1 \|\proj_{\nabla\f(0)}\nabla S(\xt)\|_2} \sqrt{\dfrac{\sum_{i=2}^m \alpha_i^2}{m-1}} \right) 
            + \gO\left(\frac{1}{\delta} \sqrt{\dfrac{2}{B} \ln\left(\frac m p \right)} \cdot \sqrt{\dfrac{\sum_{i=2}^m \alpha_i^2}{m-1}}\right).
        \end{align*}
        Therefore, the bound in \Cref{thm:approach-boundary-concentration} becomes:
        \begin{align}
            & \cos\langle \E\tnablaS(\xt), \nabla S(\xt) \rangle \nonumber \\
            \ge & 
                \dfrac{\|\proj_{\nabla\f(0)}\nabla S(\xt)\|_2}{\|\nabla S(\xt)\|_2} 
                \cdot \nonumber \\
            &
                \left(
                    1 - \dfrac{(m-1)^2 \delta^2}{8\alpha_1^2}
                    \left(
                        \gO\left(
                            \dfrac{\beta_{\f}^2}{\alpha_1^2} \cdot \dfrac{\sum_{i=2}^m \alpha_i^2}{m-1}
                        \right)
                        + 
                        \gO\left(
                            \dfrac{\beta_S^2 \max_{i\in[m]} \alpha_i^4}{\alpha_1^2 \|\proj_{\nabla\f(0)}\nabla S(\xt)\|_2^2} \cdot 
                            \dfrac{\sum_{i=2}^m \alpha_i^2}{m-1}
                        \right)
                        +
                        \gO\left(
                            \dfrac{1}{\delta^2}
                            \cdot
                            \dfrac{2}{B}\ln\left(\frac m p\right)
                            \cdot
                            \dfrac{\sum_{i=2}^m \alpha_i^2}{m-1}
                        \right)
                    \right)
                \right) \nonumber \\
            = & \dfrac{\|\proj_{\nabla\f(0)}\nabla S(\xt)\|_2}{\|\nabla S(\xt)\|_2} \cdot \left(
            1 - \gO\left(
                m^2 \cdot \dfrac{\sum_{i=2}^m \alpha_i^2}{m-1}
                \left(
                    \dfrac{\delta^2\beta_{\f}^2}{\alpha_1^4}
                    +
                    \dfrac{\alpha_{\max}^4}{\alpha_1^4} \cdot
                    \dfrac{\delta^2\beta_S^2}{\|\proj_{\nabla\f(0)} \nabla S(\xt)\|_2^2}
                    +
                    \dfrac{\ln(\frac m p)}{B\alpha_1}
                \right)
            \right)
            \right). \label{adxeq:big-o-bound}
        \end{align}
        It recovers the simplified bound in \Cref{fig:order-figure}.
        Note the the last term $\frac{\ln(m/p)}{B\alpha_1}$ is the extra term of \Cref{thm:approach-boundary-concentration} that guarantees the $1-p$ holding probability, and discarding this term yields the version bound for the expectation bound~(\Cref{thm:approach-boundary-expectation}).

\section{Discussion on Key Characteristics and Optimal Scale}
    \label{adxsec:key-characteristics}
    
    This section illustrates the key characteristics for improving the gradient estimator, and discusses the existence of the optimal scale.
    
    \subsection{Illustration of Key Characteristics}
        \begin{figure}[!h]
            \centering
            \includegraphics[width=0.80\textwidth]{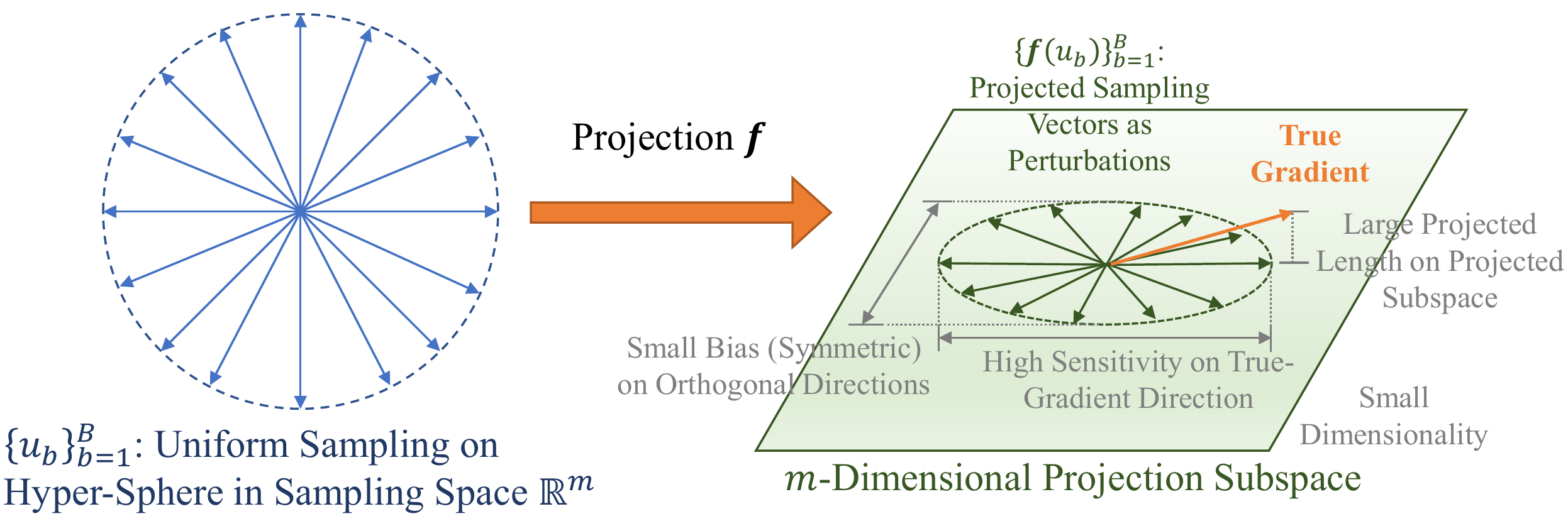}
            \caption{An illustration of key characteristics for a good projection-based gradient estimator.}
            \label{fig:key-characteristics}
        \end{figure}
        The figure illustrates the key characteristics, or the optimization goals, for improving the projection-based gradient estimator as discussed in \Cref{subsec:key-characteristics}.
    
    \subsection{Existence of Optimal Scale}
    
        As discussed in \Cref{subsec:projection-on-selected-subspace}, the scale is mapped to the dimensionality of the projection subspace---$m$.
        
        Due to the trade-off between large $\|\proj_{\nabla\f(0)} \nabla S(\xt)\|_2$ and small $m$, from \Cref{adxeq:big-o-bound}, we can intuitively learn that for a given projection $\f$, the optimal scale $m_{\opt}$ always exists.
        Now we define this formally.
        We first explicitly show that $\f$ relies on the dimensionality of the projection subspace.
        To do so, we use $\f_m: \sR^m \to \sR^n$ instead of the general notion $\f$.
        $\f_m$ can be viewed as being drawn from a pre-defined projection function family $\gF = \{\f_i: i\in [n]\}$.
        Then, the optimal scale $m_\opt$ can be then explicit expressed as such:
        $$
            m_\opt = \arg\max_{m\in [n]} 
                \dfrac{\|\proj_{\nabla\f_m(0)}\nabla S(\xt)\|_2}{\|\nabla S(\xt)\|_2} \cdot \left(
                1 - \gO\left(
                    m^2 \cdot \dfrac{\sum_{i=2}^m \alpha_i^2}{m-1}
                    \left(
                        \dfrac{\delta^2\beta_{\f_m}^2}{\alpha_1^4}
                        +
                        \dfrac{\alpha_{\max}^4}{\alpha_1^4} \cdot
                        \dfrac{\delta^2\beta_S^2}{\|\proj_{\nabla\f_m(0)} \nabla S(\xt)\|_2^2}
                        +
                        \dfrac{\ln(\frac m p)}{B\alpha_1}
                    \right)
                \right)
            \right).
        $$
        The objective function in above $\argmax$ encodes the precise bound in \Cref{thm:approach-boundary-concentration}.
        
        \subsubsection*{A Simplified Form for Linear Case}
        When both the projection function $\f_m$ and the difference function $S$ are locally linear, i.e., $\beta_{\f_m} = \beta_S = 0$, we can simplify the above equation as such:
        $$
            m_\opt = \arg\max_{m\in [n]}
                \dfrac{\|\proj_{\nabla\f_m(0)}\nabla S(\xt)\|_2}{\|\nabla S(\xt)\|_2} \left( 1
                -
                C
                m^2\ln \left(\frac m p\right)\right),
        $$
        where $0<C<1$ is a constant.
        
        Now, the existence of optimal scale becomes more apparent.
        While increasing $m$ can increase $\frac{\|\proj_{\nabla\f_m(0)}\nabla S(\xt)\|_2}{\|\nabla S(\xt)\|_2}$, this term has its upper bound $1$.
        On the other hand, the $m^2\ln (m/p)$ in the second term will also be increased, and it is unbounded.
        Therefore, an optimal $m$ should be non-zero but not large, i.e., an optimal scale $m_\opt$ usually exists.
        
        The optimal scale depends on the actual function family $\f_m$ and the difference function $S$.
        For common and practical cases, as shown in \Cref{fig:progressive-scaling}, the objective function for $\argmax$ is usually unimodal so that the progressive scaling is guaranteed to find the optimal scale.
        We leave it as our future work to theoretically analyze on what cases this objective function is strictly unimodal.

\section{Target Models} \label{adxsec:target-model}
    In this section, we introduce the target models used in the experiments including the implementation details and the target model performance.
    \subsection{Implementation Details} 
    \label{sec:tgt_details}
    \paragraph{Offline Models.}
    Following~\cite{Li_2020_CVPR,li2020nolinear}, the pretrained ResNet-18 models are used here as the target models. We also evaluate model ResNeXt50\_32$\times$4d~\cite{xie2017aggregated} to demonstrate the generalization ability. For models that are finetuned, cross entropy error is employed as the loss function and is implemented as `\texttt{torch.nn.CrossEntropyLoss}' in PyTorch.
    
    For ImageNet, no finetuning is performed as the pretrained target model is just trained exactly on ImageNet. The model is loaded with PyTorch command `\texttt{torchvision.models.resnet18(pretrained=True)}' or `\texttt{torchvision.models.resnext50\_32x4d(pretrained=True)}' following the documentation~\cite{pytorch_pretrained_models}.
    
    For CelebA, the target model is fine-tuned to do binary classification on image attributes. Among the 40 binary attributes associated with each image, the top-5 balanced attributes are `Attractive', `Mouth\_Slightly\_Open', `Smiling', `Wearing\_Lipstick', `High\_Cheekbones'. Though the `Attractive' attribute is the most balanced one, however, it is more subjective than objective, thus we instead choose the second attribute `Mouth\_Slightly\_Open'. 
    
    For MNIST and CIFAR-10 datasets, we first resize the original images to $224\times 224$ by linear interpolation, then the target model is finetuned to do 10-way classification. One reason for doing interpolation is that it can provide us more spatial scales to explore. The another reason is that the linear interpolation step also makes image sizes consistent among all the tasks and experiments.

    \paragraph{Commercial Online API.}
    Among all the APIs provided by the Face++ platform~\cite{facepp_main}, the `Compare' API~\cite{facepp-compare-api} which takes two images as input and returns a confidence score 
    indicating whether they contain the faces from the same person.
    This is also consistent with the same online attacking in~\cite{Li_2020_CVPR,li2020nolinear}.
    In implementation during the attack process, the two image arrays with floating number values are first converted to integers and stored as jpg images on disk. Then they are encoded as base64 binary data and sent as POST request to the request URL~\cite{facepp-compare-url}. We set the similarity threshold as $50\%$ in the experiments following~\cite{Li_2020_CVPR,li2020nolinear}: when the confidence score is equal to or larger than $50\%$, we consider the two faces to belong to the `same person', vice versa.
    
    For source-target images that are from two different persons, the goal of the attack is to get an adversarial image that looks like the target image (has low MSE between the adversarial image and target image), but is predicted as `same person' with the source image.
    We randomly sample 50 source-target image pairs from the CelebA dataset that are predicted as different persons by the `Compare' API. Then we apply the \ourapproach pipeline with various perturbation vector generators for comparison. 
    
    \subsection{Performance of Target Models} 
    \label{sec:tgt_model_performance}
    \label{sec:target_model_performance}
    The benign accuracy of the target models finetuned on different datasets is shown in Table~\ref{tab:target_model_performance}.
    \begin{table}[h]
        \centering
        \caption{The benign model accuracy of the target models.}
        \setlength{\tabcolsep}{7mm}{
        \begin{tabular}{c|c|c|c|c} 
        \toprule
             Model & MNIST & CelebA & CIFAR-10 & ImageNet \\
            \hline
            ResNet-18 & 99.55\% & 93.77\% & 88.15\% &  69.76\% \\
            \hline
            ResNeXt50\_32$\times$4d & 99.33\% & 94.00\% & 90.26\% &  77.62\% \\
            \bottomrule

        \end{tabular}}
        \label{tab:target_model_performance}
    \end{table}
\section{Details on \ourapproach-PGAN} \label{adxsec:ppba-pgan-detail}
    In this section, we introduce the details of Progressive-Scale based projection models including the architecture of Progressive GAN, the training procedure, the algorithm description for progressive scaling and gradient estimation, and the implementation details.
    \subsection{The Architecture of Progressive GAN} 
    \label{sec:pgan_structure}
    Progressive GAN is a method developed by Karras et. al.~\cite{progressive_gan} allowing gradually generating the image from low resolution images to high resolution images. Here, we adopt the implementation of PGAN from pytorch\_GAN\_zoo~\cite{facebookresearch} to help us explore the influence of different scales on attacking performance. \\
    The Conv2d(n\_kernel, n\_stride, n\_pad) here applies He's constant~\cite{he2015delving} at runtime, and for simplicity, the LeakyReLU(negative\_slope = 0.2) is denoted as LReLU. Besides, for the generator, we actually utilize bilinear interpolation to implement ‘Unsample’ and utilize average pool to implement ‘Downsample’. Then, the detailed model network structures for the generator and discriminator with the maximum scale $224\times224$ are listed in Table~\ref{tab:pgan_generator_structure} and Table~\ref{tab:pgan_discriminator_structure}.
    \begin{table}[!h]
        \centering
        \caption{The detailed model structure for generator in PGAN.}
        \label{tab:pgan_generator_structure}
        \setlength{\tabcolsep}{8mm}{
        \begin{tabular}{c|c|c}
        \toprule
            Generator & Act & Output shape \\
            \hline\hline
            Latent vector & - & $9408\times 1 \times 1$ \\
            Fully-connected & LReLU & $8192\times 1 \times 1$ \\
            Resize & - & $512\times 4 \times 4$ \\
            Conv(3, 1, 1) & LReLU & $512\times 4 \times 4$ \\
            \hline
            Upsample & - & $512\times 7 \times 7$ \\
            Conv(3, 1, 1) & LReLU & $256\times 7 \times 7$ \\
            Conv(3, 1, 1) & LReLU & $256\times 7 \times 7$ \\
            \hline
            Upsample & - & $256\times 14 \times 14$ \\
            Conv(3, 1, 1) & LReLU & $256\times 14 \times 14$ \\
            Conv(3, 1, 1) & LReLU & $256\times 14 \times 14$ \\
            \hline
            Upsample & - & $256\times 28 \times 28$ \\
            Conv(3, 1, 1) & LReLU & $128\times 28 \times 28$ \\
            Conv(3, 1, 1) & LReLU & $128\times 28 \times 28$ \\
            \hline
            Upsample & - & $128\times 56 \times 56$ \\
            Conv(3, 1, 1) & LReLU & $64\times 56 \times 56$ \\
            Conv(3, 1, 1) & LReLU & $64\times 56 \times 56$ \\
            \hline
            Upsample & - & $64\times 56 \times 56$ \\
            Conv(3, 1, 1) & LReLU & $32\times 112 \times 112$ \\
            Conv(3, 1, 1) & LReLU & $32\times 112 \times 112$ \\
            \hline
            Upsample & - & $32\times 224 \times 224$ \\
            Conv(3, 1, 1) & LReLU & $16\times 224 \times 224$ \\
            Conv(3, 1, 1) & LReLU & $16\times 224 \times 224$ \\
            Conv(1, 1, 0) & Tanh & $3\times 224 \times 224$ \\ 
            \bottomrule
        \end{tabular}}
    \end{table}
    \begin{table}[t]
        \centering
        \caption{The detailed model structure for discriminator in PGAN.}
        \label{tab:pgan_discriminator_structure}
        \setlength{\tabcolsep}{8mm}{
        \begin{tabular}{c|c|c}
        \toprule
            Discriminator & Act & Output shape \\
            \hline\hline
            Input image & - & $3\times 224 \times 224$ \\
            Conv(1, 1, 0) & - & $16\times 224 \times 224$ \\
            Conv(3, 1, 1) & LReLU & $16\times 224 \times 224$ \\
            Conv(3, 1, 1) & LReLU & $32\times 224 \times 224$ \\
            Downsample & - & $32\times 112 \times 112$ \\
            \hline
            Conv(3, 1, 1) & LReLU & $32\times 112 \times 112$ \\
            Conv(3, 1, 1) & LReLU & $64\times 112 \times 112$ \\
            Downsample & - & $64\times 56 \times 56$ \\
            \hline
            Conv(3, 1, 1) & LReLU & $64\times 56 \times 56$ \\
            Conv(3, 1, 1) & LReLU & $128\times 56 \times 56$ \\
            Downsample & - & $128\times 28 \times 28$ \\
            \hline
            Conv(3, 1, 1) & LReLU & $128\times 28 \times 28$ \\
            Conv(3, 1, 1) & LReLU & $256\times 28 \times 28$ \\
            Downsample & - & $256\times 14 \times 14$ \\
            \hline
            Conv(3, 1, 1) & LReLU & $256\times 7 \times 7$ \\
            Conv(3, 1, 1) & LReLU & $512\times 7 \times 7$ \\
            Downsample & - & $512\times 4 \times 4$ \\
            \hline
            Minibatch stddev & - & $513\times 4 \times 4$ \\
            Conv(3, 1, 1) & LReLU & $512\times 4 \times 4$ \\
            Fully-connected & LReLU & $512\times 1 \times 1$ \\
            Fully-connected & Linear & $1\times 1 \times 1$ \\
            \bottomrule
        \end{tabular}}
    \end{table}
    \subsection{Projection Model Training Procedure} 
    \label{sec:pgan_training_procedure}
    First, we need to prepare the datasets for PGAN training, which comprise the gradient images generated from a set of \emph{reference models}. 
    Generally, the reference models are assumed to have \emph{different structures} compared with the blackbox target model. Nonetheless, attacker-trained reference models can generate accessible gradients and provide valuable information on the distribution of the target model gradients.
    
    In our case, with the same setting as in ~\cite{li2020nolinear}, there are five reference models (i.e., DenseNet-121~\cite{huang2018densely}, ResNet-50~\cite{he2015deep}, VGG16~\cite{simonyan2015deep}, GoogleNet~\cite{szegedy2014going} and WideResNet~\cite{zagoruyko2017wide}) with different backbones compared with the target model, while the implementation and training details are similar with the target model in Section~\ref{sec:tgt_details}. The benign test accuracy results of these five reference models for MNIST, CIFAR-10 and CelebA datasets are shown in Table~\ref{tab:diff_benign_acc}.
    After the reference models are trained, their gradients with respect to the training data points are generated with PyTorch automatic differentiation function with command `loss.backward()'. The loss is the cross entropy between the prediction scores and the ground truth labels. 
    
    For ImageNet and CelebA, we randomly sample $500,000$ gradient images ($100,000$ per reference model) for each of ImageNet and CelebA and fix them throughout the experiments for fair comparison.
    
    For CIFAR-10 and MNIST, there are fewer images and so we use the whole dataset and generate $250,000$ gradient images for CIFAR-10 ($50,000$ per reference model) and $300,000$ ($60,000$ per reference model) gradient images for MNIST.

    For AE and VAE, they are directly trained on the original gradient datasets, and the training details can be found in~\cite{Li_2020_CVPR}. However, for PGAN, since the size of the images from original gradient datasets is $224\times224$, we down-scale them by average pool first and then as the new training datasets when the PGAN is trained to build the low resolution image. Actually, this training procedure is just the same as in~\cite{progressive_gan}, the only difference here is the so-called images are gradient images generated from reference models instead of the real-world pictures.
    
    \begin{table}[!hb]
        \centering
        \caption{The benign model accuracy of the reference models on four datasets. For the dataset MNIST and CIFAR10, the images are linearly interpolated to size $224\times 224$; for the dataset CelebA, the attribute is chosen as `mouth\_slightly\_open'}
        \setlength{\tabcolsep}{6mm}{
        \begin{tabular}{c|c|c|c|c|c}
        \toprule
            \hline
            Dataset & DenseNet-121 & ResNet-50 & VGG16 & GoogleNet & WideResNet \\
            \hline
            MNIST & 98.99\% & 99.43\% & 99.16\% & 99.46\% & 98.59\%\\
            \hline
            CIFAR10 & 92.73\% & 88.47\% & 92.67\% & 92.26\% & 85.19\% \\
            \hline
            CelebA & 93.81\% & 94.02\% & 94.13\% & 91.77\% & 93.79\% \\
            \hline
            ImageNet & 74.65\% & 76.15\% & 71.59\% & 69.78\% & 78.51\% \\
            \hline
            \bottomrule
        \end{tabular}}
        \label{tab:diff_benign_acc}
    \end{table}
    
    \subsection{Reference Model Performance} 
    \label{sec:ref_model_performance}
    Intuitively, with well-trained reference models that perform comparatively with the target models, the attacker can get gradient images that are in a more similar distribution with the target model's gradients for training, thus increasing the chance of an attack with higher quality.
    The reference model performance in terms of prediction accuracy for MNIST, CIFAR-10, CelebA and ImageNet datasets are shown in Table~\ref{tab:diff_benign_acc}. The model performance is comparable to that of the target models.

    \subsection{Algorithm Description}
    \label{sec:alg_pseudocode}
    We provide the pseudocode for the progressive-scale process with the \ourapproach-PGAN in Algorithm~\ref{alg:optimal_scale_search} (once the optimal scale determined, it will be used across all the pairs of source and target images) and for frequency reduction gradient estimation with the PGAN224 in Algorithm~\ref{alg:dct_transformation}.
    \begin{algorithm}
    \caption{The Process for Searching the Optimal Scale for \ourapproach-PGAN.}
    \label{alg:optimal_scale_search}
    \begin{algorithmic}[1]
    \renewcommand{\algorithmicrequire}{\textbf{Input:}}
     \renewcommand{\algorithmicensure}{\textbf{Output:}}
     \REQUIRE a validation set which comprises ten pairs of source-target images, the PGANs with different output scales, access to query the decision of target model.
     \ENSURE the optimal scale for attacking the target model.
     \STATE $optimal\_scale \gets 7\times 7$ 
     \STATE $lowest\_distance \gets \infty$ 
     \FOR{$s = 7\times7$ to $224\times224$}
        \STATE Take the PGAN generator with output scale $s$ as the gradient estimator to attack the target model.
        \STATE $current\_distance \gets $ the MSE of the the ten adversarial images to the corresponding target images after 10 step attack. The number of sampled perturbation vectors per step is set to $100$.
        \IF{$current\_distance \le lowest\_distance$}
        \STATE $lowest\_distance \gets current\_distance$
        \STATE $optimal\_scale \gets s$
        \ELSE
        \STATE \textbf{return} $optimal\_scale$
        \ENDIF
     \ENDFOR
    \STATE \textbf{return} $optimal\_scale$
     \end{algorithmic}
     \end{algorithm}
    
    \begin{algorithm}
    \caption{Frequency Reduction Gradient Estimation}
    \label{alg:dct_transformation}
    \begin{algorithmic}[1]
    \renewcommand{\algorithmicrequire}{\textbf{Input:}}
     \renewcommand{\algorithmicensure}{\textbf{Output:}}
     \REQUIRE a data point on the decision boundary $x \in \mathbb{R}^m$, nonlinear projection function $\f$, number of random sampling $B$, access to query the decision of target model $\phi(\cdot) = \sgn(S(\cdot))$.
     \ENSURE the approximated gradient $\widetilde{\nabla S}(x_{adv}^{(t)})$.
     \STATE Sample $B$ random Gaussian vectors of the lower dimension: $v_b \in \mathbb{R}^{n}$.
     \STATE Use PGAN224 to project the random vectors to the gradient space: $u_b = \f(v_b)\in \mathbb{R}^m$.
     \STATE Do DCT transformation on each channel of $u_b$ and get the frequency representation: $d_b = \text{DCT}(u_b)$
     \STATE Save the $k\times k$ signals on the upper left corner and set other signals to zero : $d_b' = \text{Filter}(d_b)$
     \STATE Map the signals back to the original space by Inverse DCT transformation: $u_b' = \text{IDCT}(d_b')$
     \STATE Get query points by adding perturbation vectors with the original point on the decision boundary $x_{adv}^{(t)} + \delta u_b'$.
     \STATE Monte Carlo approximation for the gradient: \\ 
     $\widetilde{\nabla S}(x_{adv}^{(t)}) = \frac1B \sum_{b=1}^B \phi\left(x_{adv}^{(t)}+\delta u_b'\right) u_b' = \frac1B \sum_{b=1}^B \sgn \left(S\left(x_{adv}^{(t)}+\delta u_b'\right) \right) u_b'$
     \STATE \textbf{return}  $\widetilde{\nabla S}(x_{adv}^{(t)})$
     \end{algorithmic}
     \end{algorithm}
    
    \subsection{Attack Implementation}
    \label{sec:attack_process_implementation}
    The goal is to generate an adversarial image that looks similar as the target image, i.e., as close as to the target image, but is predicted as the label of the target image. 
    We fix the random seed to $0$ so that the samples are consistent across different runs and various methods to ensure reproducibility and to facilitate fair comparison. 
    
    \paragraph{Gradient Estimation. }
    For convenience and precision concern, we just use $\delta_t\f(u_b)$ instead of the theoretical representation $\Delta \f(\delta_t u_b)$, i.e., $\f(\delta_t u_b) - \f(0)$ in the actual calculation of estimated gradient $\tnablaS(x_t)$ and the $\f(u_b)$ is normalized here. Besides, the variance reduction balancing adopted in~\cite{chen2020hopskipjumpattack} is also applied in our gradient estimation out of the concern for the accuracy of estimation.
    
    \paragraph{Offline Models. }
    During the attack, we randomly sample source-target pairs of images from each of the corresponding datasets. We query the offline models with the sampled images to make sure both source image and target image are predicted as their ground truth labels and the labels are different so that the attack is nontrivial.
    For the same dataset, the results of different attack methods are reported as the average of the same $50$ randomly sampled pairs.
    \paragraph{Online API. }
    For the online API attacks, the source-target pairs are sampled from the dataset CelebA.  
    The results of different attack methods are also reported as the average of the same $50$ randomly sampled pairs.
    
\section{Quantitative Results} \label{adxsec:quantitative}
    \label{sec:quantitative_results}
    
    \subsection{Attack Setup}
    \label{sec:attack_setup}    
    We randomly select $50$ pairs of source and target images from test set that are predicted by the target model as different classes for both offline attack and online attack. The goal here is to move the source image gradually to the target image under the measure of MSE while maintaining being predicted as source label by the target model. In this process, the number of sampled perturbation vectors at each step~($B$ in \Cref{def:gradient-estimator}) is controlled as $100$ for every gradient generator in the Monte Carlo algorithm to estimate the gradient for fairness (except EA, in which the $B$ is set to 1) . The optimal dimensions chosen on the search space for EA are shown in~\Cref{tab:ea_best_dimension}, and other hyper-parameters are the same with the setting in~\cite{dong2019efficient}.
    
    \begin{table}[!ht]
        \centering
        \caption{The optimal dimension of the search space for EA on different datasets and target models.}
        \label{tab:ea_best_dimension}
        \setlength{\tabcolsep}{7mm}{
        \begin{tabular}{c|c|c|c|c}
        \toprule
        \hline
            Model & MNIST & CIFAR-10 & CelebA & ImageNet  \\
            \hline
            ResNet-18 & $30\times30\times1$ & $30\times30\times3$ & $112\times112\times3$ & $30\times30\times3$\\
            \hline
             ResNeXt50\_32$\times$4d & $30\times30\times1$ & $45\times45\times3$ & $45\times45\times3$ & $45\times45\times3$\\
             \hline
             \bottomrule
        \end{tabular}}
    \end{table}
    
    \subsection{Time and Resource Consumption}
    \label{sec:time_consumption}    
    The optimal scale is usually small and relatively stable for an assigned dataset as shown in~\Cref{sec:optimal_scale_dif_model}. Indeed, from our experimental observation, it is enough to just use 10 scr-tgt image pairs to determine the optimal scale within 10 minutes on one 2080 Ti GPU. Besides, the model is trained before we start to attack, and the optimal scale is also determined before the attack. So in fact, no matter in the training or attacking stage, the time and resource consumption are almost the same with other generative model-based attacks~\cite{li2020nolinear, tu2020autozoom}. The PGAN training time on scale $28\times28$ for ImageNet  is about one day with two RTX 2080 Ti GPUs and that for the scale $224\times224$ is about two days. We note that the PGAN training can be done offline and is one-time training for attacking different models. Besides, the averaged attack time with $10,000$ queries on ImageNet of HSJA is $114.2$s, and that of PSBA is $136.4$s. 
    We remark that the bulk of PSBA attack time is on the resize operation similar to the baseline QEBA-S~(see \citep{Li_2020_CVPR}).
   

    \subsection{Attack Performance for Different Datasets and Target Models}
    \label{sec:different_tgt_model}    
\begin{figure*}[!ht]
        \centering
        \vspace{-0.2cm}
        \includegraphics[width=0.9\textwidth]{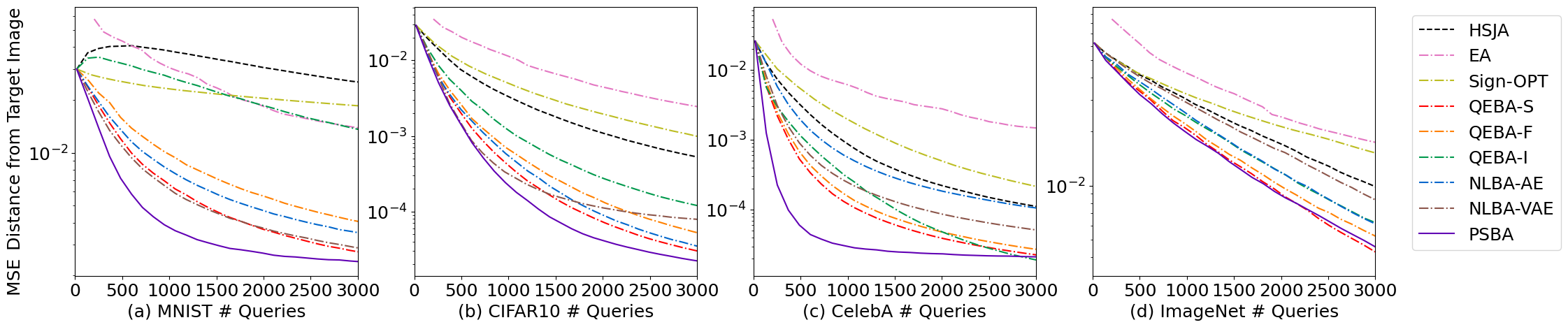}
        \vspace{-0.2cm}
        \includegraphics[width=0.9\textwidth]{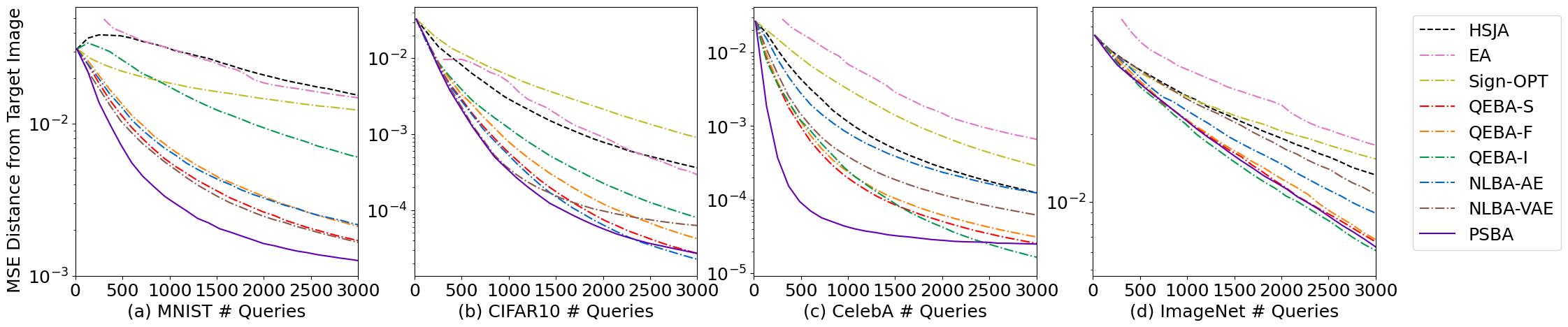}
        \vspace{-2mm}
        \caption{\small Perturbation magnitude (MSE) w.r.t. query numbers. Row 1: For attacks on ResNet-18; Row 2: For attacks on ResNeXt50\_32$\times$4d.}
        \vspace{-3mm}
        \label{fig:addition_dataset_tgt_model}
    \end{figure*}

    The complete attack performance results for different datasets and target models are shown in~\cref{fig:addition_dataset_tgt_model}. On complex datasets like ImageNet, the major challenge is the excessive number of categories (1,000 on ImageNet). Thus, we suspect that the geometry of model’s decision boundary is more non-smooth and complex, and therefore the general boundary attacks should be harder to improve.
    
     \begin{figure*}[!ht]
        \centering
        \vspace{-0.2cm}
        \includegraphics[width=0.9\textwidth]{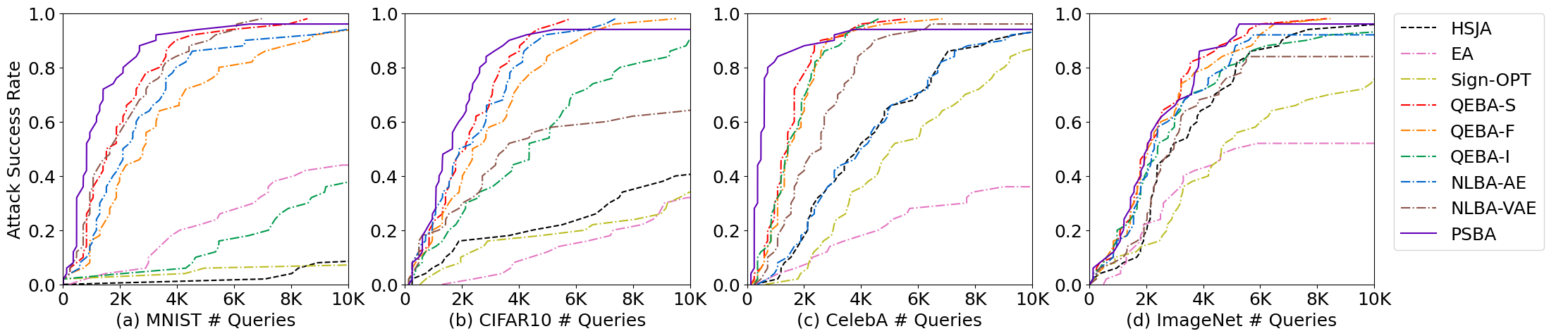}
        \vspace{-0.2cm}
        \includegraphics[width=0.9\textwidth]{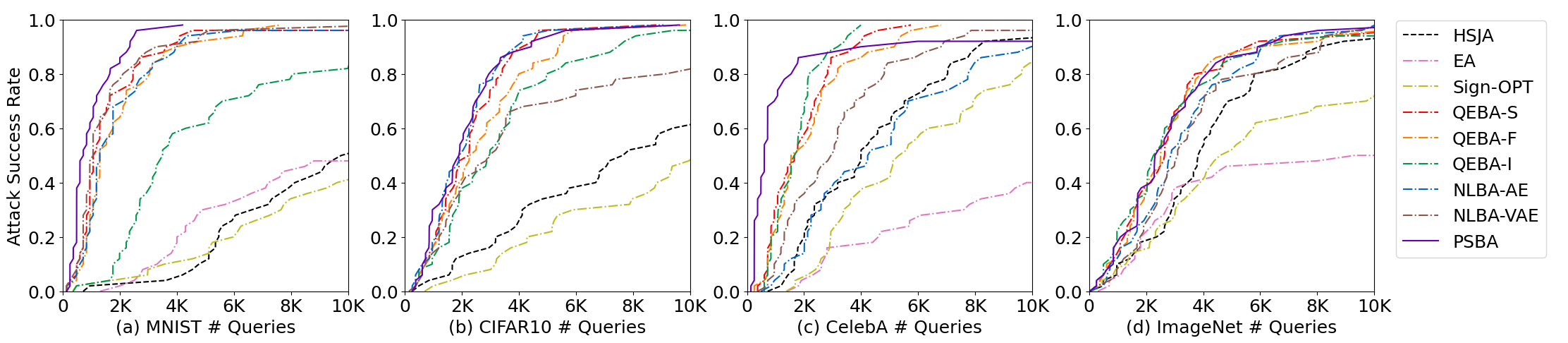}
        \vspace{-2mm}
        \caption{\small The attack success rate w.r.t. number of queries on four different datasets. Row 1: For attacks on ResNet-18; Row 2: For attacks on ResNeXt50\_32$\times$4d.}
        \vspace{-3mm}
        \label{fig:spatial_attack_success}
    \end{figure*}
    
         \begin{table*}[!ht]
         \renewcommand\arraystretch{1.15}
        \centering
        \caption{Comparison of the attack success rate  for different attacks at query number 1K (the perturbation magnitude under MSE for each dataset are: MNIST: $5e-3$; CIFAR10: $5e-4$; CelebA: $1e-4$; ImageNet: $1e-2$).}
        \label{tab:1k_success_rate}
        \begin{small}
        \begin{tabular}{c|c|c|c|c|c|c|c|c|c|c}
        \toprule
        \hline
        \multirow{2}{*}{Data}     & \multirow{2}{*}{Model} & \multicolumn{9}{c}{\# Queries = 1K}                                               \\ \cline{3-11} 
                                  &                        & HSJA & EA & Sign-OPT & QEBA-S & QEBA-F & QEBA-I & NLBA-AE & NLBA-VAE & PSBA        \\ \hline
        \multirow{2}{*}{MNIST}    & ResNet                 & 2\%    & 4\%  & 4\%        & 40\%     & 18\%     & 4\%      & 26\%      & 44\%       & \textbf{64\%} \\ \cline{2-11} 
                                  & ResNeXt                & 4\%    & 0\%  & 4\%        & 44\%     & 30\%     & 4\%     & 28\%      & 50\%       & \textbf{64\%} \\ \hline
        \multirow{2}{*}{CIFAR10}  & ResNet                 & 16\%    & 2\%  & 14\%        & 52\%     & 42\%     & 30\%     & 50\%      & 50\%       & \textbf{62\%} \\ \cline{2-11} 
                                  & ResNeXt                & 12\%    & 0\%  & 8\%        & 44\%     & 38\%     & 30\%     &48\%      &  \textbf{52\%}       & 48\%  \\ \hline
        \multirow{2}{*}{CelebA}   & ResNet                 & 8\%   & 6\%  & 2\%        & 48\%     & 36\%     & 40\%     & 10\%      & 22\%       & \textbf{86\%} \\ \cline{2-11} 
                                  & ResNeXt                & 2\%   & 0\%  & 0\%        & 32\%     & 24\%     & 22\%     & 6\%      & 12\%       & \textbf{72\%} \\ \hline
        \multirow{2}{*}{ImageNet} & ResNet                 & 10\%   & 8\% & 12\%        & \textbf{22\%}     & 18\%     & \textbf{22\%}     & 18\%      & 10\%       & 20\% \\ \cline{2-11} 
                                  & ResNeXt                & 12\%   & 6\% & 10\%       & 16\%     & 14\%     & \textbf{24\%}\     & 14\%      & 10\%       & 20\% \\ \hline
        \bottomrule
        \end{tabular}
        \end{small}
    \end{table*}

     \begin{table*}[!ht]
     \renewcommand\arraystretch{1.15}
        \centering
        \caption{Comparison of the attack success rate  for different attacks at query number 3K (the perturbation magnitude under MSE for each dataset are: MNIST: $5e-3$; CIFAR10: $5e-4$; CelebA: $1e-4$; ImageNet: $1e-2$).}
        \label{tab:3k_success_rate}
        \begin{small}
        \begin{tabular}{c|c|c|c|c|c|c|c|c|c|c}
        \toprule
        \hline
        \multirow{2}{*}{Data}     & \multirow{2}{*}{Model} & \multicolumn{9}{c}{\# Queries = 3K}                                               \\ \cline{3-11} 
                                  &                        & HSJA & EA & Sign-OPT & QEBA-S & QEBA-F & QEBA-I & NLBA-AE & NLBA-VAE & PSBA        \\ \hline
        \multirow{2}{*}{MNIST}    & ResNet                 & 2\%    & 12\%  & 4\%        & 80\%     & 58\%     & 6\%      & 66\%      & 76\%       & \textbf{90\%} \\ \cline{2-11} 
                                  & ResNeXt                & 4\%    & 10\%  & 10\%        & 88\%     & 80\%     & 38\%     & 82\%      & 90\%       & \textbf{98\%} \\ \hline
        \multirow{2}{*}{CIFAR10}  & ResNet                 & 40\%   & 34\% & 24\%       & 96\%     & 90\%     & 74\%     & 90\%      & \textbf{96\%}       & \textbf{96\%} \\ \cline{2-11} 
                                  & ResNeXt                & 50\%   & 2\%  & 30\%       & 96\%     & 90\%     & 82\%     & 96\%      & \textbf{100\%}       & 96\% \\ \hline
        \multirow{2}{*}{CelebA}   & ResNet                 & 40\%   & 16\%  & 24\%        & 92\%     & 92\%     & 88\%     & 44\%      & 68\%       & \textbf{94\%} \\ \cline{2-11} 
                                  & ResNeXt                & 38\%   & 18\%  & 24\%        & 88\%     & 78\%     & 90\%     & 40\%      & 52\%       & \textbf{90\%} \\ \hline
        \multirow{2}{*}{ImageNet} & ResNet                 & 54\%   & 36\% & 30\%        & \textbf{68\%}     & 66\%     & 64\%     & 64\%      & 56\%       & \textbf{68\%} \\ \cline{2-11} 
                                  & ResNeXt                & 36\%   & 38\% & 28\%       & \textbf{68\%}     & 64\%     & 64\%     & 54\%      & 48\%       & 66\% \\ \hline
        \bottomrule
        \end{tabular}
        \end{small}
    \end{table*}
    
        \begin{table*}[!ht]
        \renewcommand\arraystretch{1.15}
        \centering
        \caption{Comparison of the attack success rate  for different attacks at query number 5K (the perturbation magnitude under MSE for each dataset are: MNIST: $5e-3$; CIFAR10: $5e-4$; CelebA: $1e-4$; ImageNet: $1e-2$).}
        \label{tab:5k_success_rate}
        \begin{small}
        \begin{tabular}{c|c|c|c|c|c|c|c|c|c|c}
        \toprule
        \hline
        \multirow{2}{*}{Data}     & \multirow{2}{*}{Model} & \multicolumn{9}{c}{\# Queries = 5K}                                               \\ \cline{3-11} 
                                  &                        & HSJA & EA & Sign-OPT & QEBA-S & QEBA-F & QEBA-I & NLBA-AE & NLBA-VAE & PSBA        \\ \hline
        \multirow{2}{*}{MNIST}    & ResNet                 & 2\%    & 24\%  & 8\%        & 94\%     & 76\%     & 12\%      & 88\%      & 90\%       & \textbf{96}\% \\ \cline{2-11} 
                                  & ResNeXt                & 12\%    & 32\%  & 14\%        & 98\%     & 94\%     & 62\%     & 96\%      & 96\%       & \textbf{100}\% \\ \hline
        \multirow{2}{*}{CIFAR10}  & ResNet                 & 22\%    & 12\%  & 20\%        & \textbf{96}\%     & 86\%     & 54\%     & 94\%      & 58\%       & 94\% \\ \cline{2-11} 
                                  & ResNeXt                & 36\%    & 2\%  & 22\%        & \textbf{98}\%     & 86\%     & 82\%     &96\%      &  70\%      & 94\%  \\ \hline
        \multirow{2}{*}{CelebA}   & ResNet                 & 66\%   & 24\%  & 50\%        & 98\%     & 98\%     & \textbf{100}\%     & 66\%      & 92\%       & 96\% \\ \cline{2-11} 
                                  & ResNeXt                & 62\%   & 24\%  & 42\%        & 98\%     & 90\%     & \textbf{100}\%     & 54\%      & 86\%       & 92\% \\ \hline
        \multirow{2}{*}{ImageNet} & ResNet                 & 74\%   & 50\% & 54\%        & 90\%     & 86\%     & 82\%     & 84\%      & 78\%       & \textbf{92}\% \\ \cline{2-11} 
                                  & ResNeXt                & 72\%   & 48\% & 54\%       & \textbf{88}\%     & \textbf{88}\%     & 86\%     & 78\%      & 80\%       & \textbf{88}\% \\ \hline
        \bottomrule
        \end{tabular}
        \end{small}
    \end{table*}
    
     \begin{table*}[!ht]
     \renewcommand\arraystretch{1.15}
        \centering
        \caption{Comparison of the attack success rate  for different attacks at query number 8K (the perturbation magnitude under MSE for each dataset are: MNIST: $5e-3$; CIFAR10: $5e-4$; CelebA: $1e-4$; ImageNet: $1e-2$).}
        \label{tab:8k_success_rate}
        \begin{small}
        \begin{tabular}{c|c|c|c|c|c|c|c|c|c|c}
        \toprule
        \hline
        \multirow{2}{*}{Data}     & \multirow{2}{*}{Model} & \multicolumn{9}{c}{\# Queries = 8K}                                               \\ \cline{3-11} 
                                  &                        & HSJA & EA & Sign-OPT & QEBA-S & QEBA-F & QEBA-I & NLBA-AE & NLBA-VAE & PSBA        \\ \hline
        \multirow{2}{*}{MNIST}    & ResNet                 & 4\%    & 40\%  & 8\%        & 98\%     & 90\%     & 30\%      & 92\%      & \textbf{100}\%       & 98\% \\ \cline{2-11} 
                                  & ResNeXt                & 40\%    & 46\%  & 36\%        & 98\%     & \textbf{100}\%     & 80\%     & 98\%      & 98\%       & \textbf{100}\% \\ \hline
        \multirow{2}{*}{CIFAR10}  & ResNet                 & 36\%    & 22\%  & 24\%        & \textbf{100}\%     & 98\%     & 82\%     & \textbf{100}\%      & 64\%       & 96\% \\ \cline{2-11} 
                                  & ResNeXt                & 54\%    & 2\%  & 36\%        & \textbf{98}\%     & \textbf{98}\%     & 94\%     &\textbf{98}\%      &  80\%      & \textbf{98}\%  \\ \hline
        \multirow{2}{*}{CelebA}   & ResNet                 & 88\%   & 36\%  & 72\%        & \textbf{100}\%     & \textbf{100}\%     & \textbf{100}\%     & 90\%      & 98\%       & 96\% \\ \cline{2-11} 
                                  & ResNeXt                & 90\%   & 34\%  & 72\%        & \textbf{100}\%     & \textbf{100}\%     & \textbf{100}\%     & 86\%      & 98\%       & 94\% \\ \hline
        \multirow{2}{*}{ImageNet} & ResNet                 & 96\%   & 54\% & 70\%        & \textbf{98}\%     & \textbf{98}\%     & 92\%     & 94\%      & 86\%       & \textbf{98}\% \\ \cline{2-11} 
                                  & ResNeXt                & 90\%   & 48\% & 70\%       & 94\%     & 92\%     & 94\%     & 96\%      & 88\%       & \textbf{96}\% \\ \hline
        \bottomrule
        \end{tabular}
        \end{small}
    \end{table*}
    
     \begin{table*}[!ht]
     \renewcommand\arraystretch{1.15}
        \centering
        \caption{Comparison of the attack success rate  for different attacks at query number 10K (the perturbation magnitude under MSE for each dataset are: MNIST: $5e-3$; CIFAR10: $5e-4$; CelebA: $1e-4$; ImageNet: $1e-2$).}
        \label{tab:10k_success_rate}
        \begin{small}
        \begin{tabular}{c|c|c|c|c|c|c|c|c|c|c}
        \toprule
        \hline
        \multirow{2}{*}{Data}     & \multirow{2}{*}{Model} & \multicolumn{9}{c}{\# Queries = 10K}                                               \\ \cline{3-11} 
                                  &                        & HSJA & EA & Sign-OPT & QEBA-S & QEBA-F & QEBA-I & NLBA-AE & NLBA-VAE & PSBA        \\ \hline
        \multirow{2}{*}{MNIST}    & ResNet                 & 10\%    & 46\%  & 8\%        & \textbf{100}\%     & 94\%     & 38\%      & 96\%      & \textbf{100}\%       & 98\% \\ \cline{2-11} 
                                  & ResNeXt                & 52\%    & 50\%  & 42\%        & 98\%     & \textbf{100}\%     & 84\%     & 98\%      & 98\%       & \textbf{100}\% \\ \hline
        \multirow{2}{*}{CIFAR10}  & ResNet                 & 42\%    & 34\%  & 36\%        & \textbf{100}\%     & \textbf{100}\%     & 92\%     & \textbf{100}\%      & 66\%       & 96\% \\ \cline{2-11} 
                                  & ResNeXt                & 62\%    & 2\%  & 50\%        & \textbf{100}\%     & \textbf{100}\%     & 98\%     &\textbf{100}\%      &  82\%      & \textbf{100}\%  \\ \hline
        \multirow{2}{*}{CelebA}   & ResNet                 & 94\%   & 38\%  & 88\%        & \textbf{100}\%     & \textbf{100}\%     & \textbf{100}\%     & 94\%      & 98\%       & 96\% \\ \cline{2-11} 
                                  & ResNeXt                & 94\%   & 42\%  & 86\%        & \textbf{100}\%     & \textbf{100}\%     & \textbf{100}\%     & 92\%      & 98\%       & 94\% \\ \hline
        \multirow{2}{*}{ImageNet} & ResNet                 & 96\%   & 54\% & 76\%        & \textbf{100}\%     & \textbf{100}\%     & 94\%     & 94\%      & 86\%       & 98\% \\ \cline{2-11} 
                                  & ResNeXt                & 94\%   & 52\% & 72\%       & 96\%     & 96\%     & 96\%     & \textbf{98}\%      & 98\%       & \textbf{98}\% \\ \hline
        \bottomrule
        \end{tabular}
        \end{small}
    \end{table*}        
    
   The `successful attack' is defined as the $x_{adv}$ reaching some pre-defined distance threshold under the metric of MSE. Note that because the complexity of tasks and images varies between datasets, we set different MSE thresholds for the datasets. For example, ImageNet images are the most complicated so the task is most difficult, thus we set larger (looser) threshold for it. The corresponding numerical results for small query number constrains are shown in~\Cref{tab:1k_success_rate} and~\Cref{tab:3k_success_rate}, the visualized attack success rates for different target models are shown in~\cref{fig:spatial_attack_success}. Since in practice, we care about the efficiency more, that is, we usually focus on the attack performance when the query number is small, like 1K or 2K.
    But we still provide the results for large query number constraints in~\Cref{tab:5k_success_rate}, ~\Cref{tab:8k_success_rate} and~\Cref{tab:10k_success_rate},
    showing that PSBA  achieves the highest or comparable ASR to other approaches. 
    Note that PSBA converges significantly faster than baselines ($\leq 3K$) (Fig.6 \& 10 in paper), which leads to its high attack success rate with small number of queries. On the other hand, when large number of queries are allowed, baselines such as QEBA will eventually converge to similar result (e.g., close to 100\% ASR), which again demonstrates the importance of evaluation under small query budget.
    

    \begin{table}[!ht]
    \renewcommand\arraystretch{1.3}
    \vspace{-0.2cm}
    \centering
    \caption{\small Comparison of Attack Success Rate (ASR) on 100 randomly chosen ImageNet images for ResNet-18 with different perturbation thresholds $\epsilon$ and attack types (the query budget = $10000$).}
    \begin{small}
    \vspace{0.2cm}
    \begin{tabular}{c|c|c|c|c|c}
    \toprule
    \hline
    Attack Type                        & $\epsilon$                     & Methods & Avg. Queries    & Med. Queries    & ASR (\%)     \\ \hline
    \multirow{6}{*}{Targeted Attack}   & \multirow{3}{*}{0.30 ($\ell_{\infty}$)} & HSJA    & 1909.9           & 853.0           & 45 \\ \cline{3-6} 
                                       &                             & RayS    & 4677.0            & 4677.0            & 2            \\ \cline{3-6} 
                                       &                             & \cellcolor[gray]{0.8}PSBA    & \cellcolor[gray]{0.8}\textbf{1437.7} & \cellcolor[gray]{0.8}\textbf{502.0}  & \cellcolor[gray]{0.8}\textbf{96} \\ \cline{2-6} 
                                       & \multirow{3}{*}{0.01 (MSE)}   & HSJA    & 3182.8          & 2726.0          & \textbf{99}           \\ \cline{3-6} 
                                       &                             & RayS    & 2479.0           & 2479.0           & 2            \\ \cline{3-6} 
                                       &                             & \cellcolor[gray]{0.8}PSBA    & \cellcolor[gray]{0.8}\textbf{2460.4} & \cellcolor[gray]{0.8}\textbf{1924.0} & \cellcolor[gray]{0.8}\textbf{99}  \\ \hline
    \multirow{6}{*}{Untargeted Attack} & \multirow{3}{*}{0.05 ($\ell_{\infty}$)} & HSJA    & 1782.8          & 595.5           & 88           \\ \cline{3-6} 
                                       &                             & RayS    & \textbf{528.7}  & \textbf{214.5}  & \textbf{100} \\ \cline{3-6} 
                                       &                             & \cellcolor[gray]{0.8}PSBA    & \cellcolor[gray]{0.8}596.9           & \cellcolor[gray]{0.8}270.0           & \cellcolor[gray]{0.8}99           \\ \cline{2-6} 
                                       & \multirow{3}{*}{0.0001 (MSE)} & HSJA    & 2208.4          & 1305.0          & 92           \\ \cline{3-6} 
                                       &                             & RayS    & 1798.1          & 625.0           & 83           \\ \cline{3-6} 
                                       &                             & \cellcolor[gray]{0.8}PSBA    & \cellcolor[gray]{0.8}\textbf{1151.5} & \cellcolor[gray]{0.8}\textbf{486.0}  & \cellcolor[gray]{0.8}\textbf{96}  \\ \hline
    \bottomrule
    \end{tabular}
    \end{small}
    \label{tab:rays_comparison}
    \end{table}

     \subsection{Comparison with \textbf{RayS} Attack}
     \label{sec:rays_attack}
     RayS attack~\cite{chen2020rays} performs blackbox attack by enumerating gradient signs at different scales, which is an efficient attack strategy for untargeted attack under $\ell_{\infty}$ norm. While PSBA can perform both untargeted and targeted attacks, the gradient generator used in our experiments is mainly designed for the targeted attack under MSE measure, which is a more practical and tougher task. To customize our \ourapproach for $\ell_{\infty}$ norm bounded attack scenario, we believe some specific design for the generator is needed. But currently, even if we directly use the original PGAN generator, our method can still compete with the RayS attack when compared under the $\ell_{\infty}$ based untargeted attack scenario. The corresponding additional experiments conducted on ImageNet dataset are shown in~\Cref{tab:rays_comparison}, demonstrating that the PSBA always outperforms RayS in targeted attacks, while RayS achieves slightly better results for untargeted attack under $\ell_\infty$.
     
     Since the RayS in targeted setting is not mentioned in the original paper, we have tried two ways to implement it: 1) initialize the perturbation using the same setting as that in untargeted attack; 2) initialize the perturbation using the images from the target class. The values recorded in the table are the better ones between these two ways and there may still exist a better way to do it. The other thing to note is that currently all the methods are not so powerful on ImageNet for targeted attack with the perturbation threshold $\epsilon$ set to $0.05$ under $\ell_{\infty}$ norm.  
     
     \subsection{Cosine Similarity Measure for Offline Models}
     \label{sec:cosine_similarity}
\begin{figure*}[!h]
        \centering
        \vspace{-0.2cm}
        \includegraphics[width=0.9\textwidth]{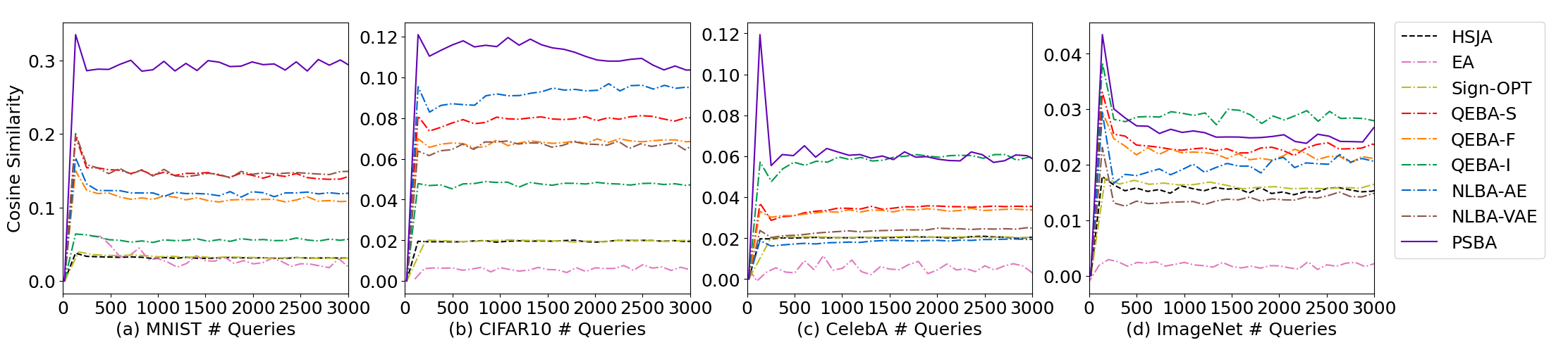}
        \vspace{-0.2cm}
        \includegraphics[width=0.9\textwidth]{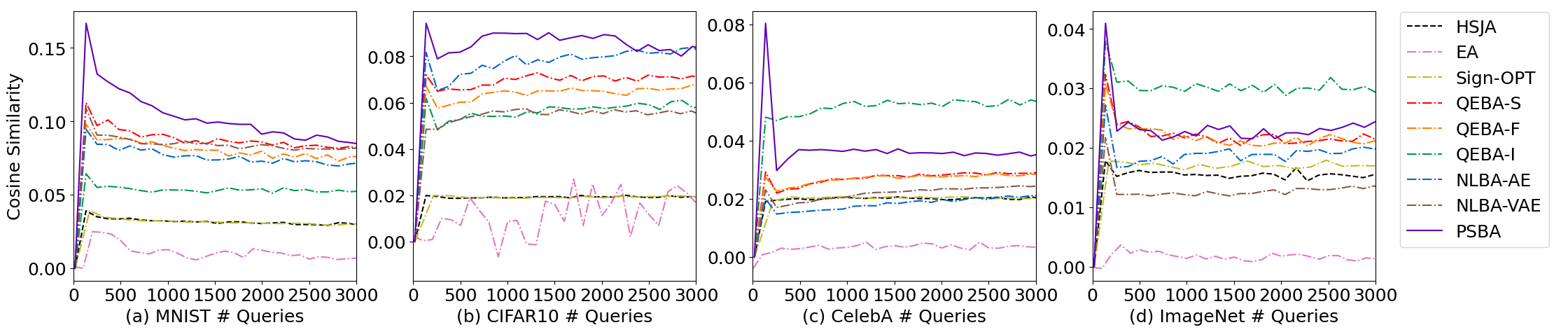}
        \vspace{-2mm}
        \caption{\small The cosine similarity between the estimated and true gradient w.r.t. the number of queries during attacking models on different datasets. Row 1: For attacks on ResNet-18; Row 2: For attacks on ResNeXt50\_32$\times$4d.}
        \vspace{-3mm}
        \label{fig:spatial_cos_sim}
    \end{figure*}

     The cosine similarity between the estimated gradient and the true gradient is a significant measure of the quality of the gradient estimation, and is highly correlated with the actual attack performance.
     The cosine similarity for the different boundary attack methods are shown in~\Cref{fig:spatial_cos_sim}.
     As we can see, our approach \ourapproach-PGAN usually achieves higher cosine similarity especially when the number of queries is limited.
    
     \subsection{Long Tail Distribution for the Gradients Generated from Different Models on Frequency Domain}
     \label{sec:long_tailed_distribution}
\begin{figure*}    
    \centering
    \includegraphics[width=\textwidth]{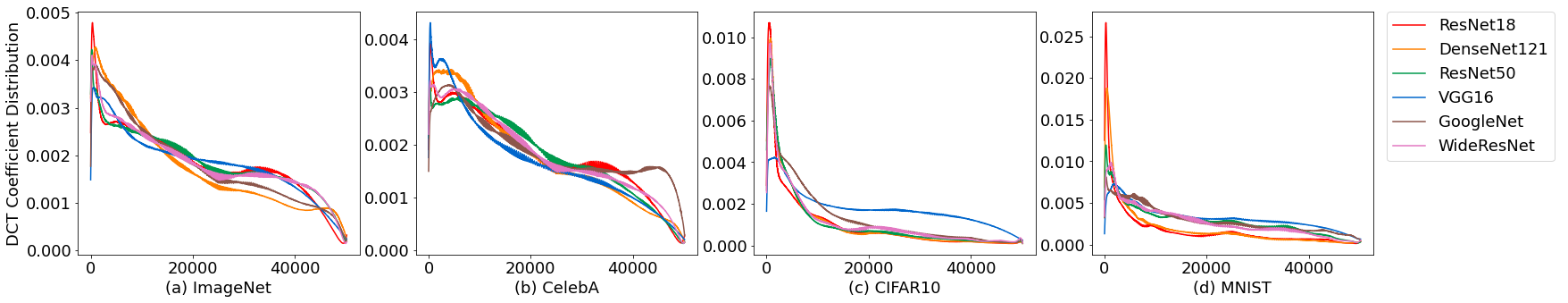}
    \vspace{-3mm}
    \caption{\small The long-tailed distribution for the coefficients of the gradients represented on DCT basis on different models and datasets.}
    \label{fig:long_tailed_distribution}
    \vspace{-4mm}
    \end{figure*}

     As mentioned in the~\Cref{subsec:analysis-verify}, the gradients generated by the target model ResNet-18 tend to focus on the low-frequency region. However, this pattern actually exists on other models as well and the experiments conducted here are in a more statistical sense: first, gradients from $1,000$ images are generated from these six models on different datasets respectively; then, by transforming them into frequency domain by DCT transformation, we average the absolute value of the coefficients on the corresponding basis components and smooth them by the Savitzky-Golay filter. As a result, if we draw these components from low-frequency to high-frequency on $x$-axis, we will see the interesting long tail distribution as shown in the~\Cref{fig:long_tailed_distribution}. This extensively existed phenomenon, as justified in \Cref{subsec:projection-on-selected-subspace}, indicates that the attack performance would be improved if we just save the low-frequency part of the generated gradient images.
     This conjecture has already been proved by our experiments in~\Cref{sec:ppba_freq_spectrum}.
     
     \subsection{High Sensitivity on Gradient Direction}
     \label{sec:high_sensitivity}
      Hereinafter, the target model is specified to   ResNet-18 for reducing the redundancy. The tendency on ResNeXt models or other models is the same, that is, the results shown below are actually decoupled with the structure of the target model.
     \vspace{-4mm}
     \paragraph{Verification of High Sensitivity.}
     We empirically verify that the trained PGAN has higher sensitivity on the projected true gradient as discussed in \Cref{subsec:analysis-verify}.
     The gradient estimator chosen here is the PGAN generator with the optimal scale $28\times28$ for each dataset. Inspired by the definition in \Cref{lem:nabla-f-decomposition}, the value $\alpha_1^2$ is approximately calculated by $\frac{1}{B} \sum_{b=1}^B \cos^2 \langle\Delta \f(\delta_t u_b),\proj_{\nabla\f(0)} \nabla S(\xt)\rangle$, where the number of queries, i.e., $B$, is set to $10,000$ instead of the original $100$ for better estimation and verification. 
     The value $\frac{\sum_{i=2}^m \alpha_i^2}{m-1}$ is then approximately calculated by $\frac{1}{B(m-1)} \sum_{b=1}^{B} (1 - \cos^2\langle\Delta \f(\delta_t u_b),\proj_{\nabla\f(0)} \nabla S(\xt)\rangle)$. Since the actual output scale of $\f$ is $28\times28$ here, the ground gradient $\nabla S(\xt)$ is resized to $28\times 28$ first~(thus denoted by $\proj_{\nabla\f(0)} \nabla S(\xt)$) and the value $m$ is equal to $n\_channel\times28\times28$ here.
     All these values are averaged on 50 pairs of source-target images with $10$-step attack on ResNet-18.
     
\begin{figure}[htbp]
    \centering
    \begin{minipage}[t]{0.48\textwidth}
    \centering
    \includegraphics[width=6cm]{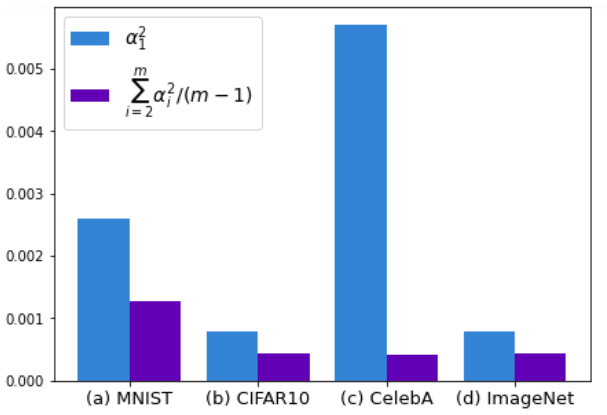}
    \caption{\small The $\alpha$ value on diverse datasets.}
    \label{fig:alpha_value}
    \end{minipage}
    \begin{minipage}[t]{0.48\textwidth}
    \centering
    \includegraphics[width=7cm]{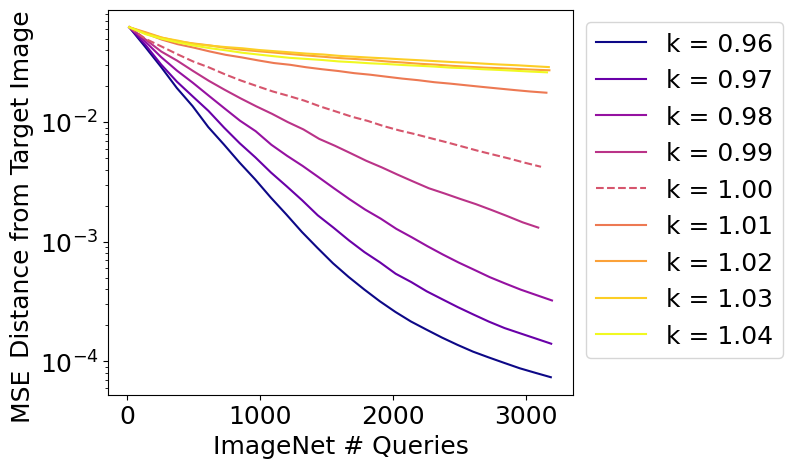}
    \caption{\small The perturbation magnitude w.r.t. different number of queries for different $k$ values on ImageNet.}
    \label{fig:k_value}
    \end{minipage}
    \end{figure}

     As shown in~\Cref{fig:alpha_value}, across all four datasets, we observe that the sensitivity on the projected true gradient direction~(blue bars) is significantly higher than the (averaged) sensitivity on other orthogonal directions~(purple bars).
     
     \paragraph{Adjust Sensitivity on Different Directions. }
     Here, we deliberately adjust the sensitivity on different directions to show the correlation between the attack performance and sensitivity by changing the weight of the components which are orthogonal to the ground gradient. In other words, we replace the $\Delta \f(\delta_t u_b)$ in the original calculation of estimated gradient $\tnablaS(x_t)$ with $ \left( \dfrac{\langle \Delta \f(\delta_t u_b),\nabla S(\xt)\rangle}{\|\nabla S(\xt)\|} \nabla S(\xt) + k \left( \Delta \f(\delta_t u_b) - \dfrac{\langle \Delta \f(\delta_t u_b), \nabla S(\xt) \rangle} {\| \nabla S(\xt)\|} \nabla S(\xt)\right) \right)$ and then repeat the attack on the target model. Lower value of $k$ means less weight is put on the orthogonal components. 
     Empirically, the range of $k$ is set between $0.96$ to $1.04$ and it is worth noting that when the value of $k$ is set to $1$, the new gradient estimation adopted here is just the same with the original gradient estimation.
     We choose the projection `PGAN28' and dataset ImageNet.
     As shown in \Cref{fig:k_value}, aligned with our theoretical analysis, lower $k$ results in better attack performance, and vise versa. 
    
    \subsection{\ourapproach with Different Domains}
    \label{sec:ppba_freq_spectrum}
    In this subsection, we demonstrate the performance of \ourapproach when applied to other domains like frequency and spectrum.Since the PGAN is originally designed for spatial expansion, it would be also beneficial for us to adopt some similar training strategy to expand progressively on both frequency and spectrum domain. However, for both convenience and effectiveness, we just take PGAN224, whose attack performance is almost the worst compared to other PGAN with smaller scale, as the gradient generator in the experiments here.  As a result, the attack performance on these other domains can be further improved by progressively expanding training strategy. 

    With this in mind, in this subsection, the main attention is concentrated on: 1) the existence of the optimal scale on both frequency and spectrum domain.
    2)~whether the original attack performance of PGAN224 would be improved a lot by just selecting the optimal scale on specific frequency or spectrum domain. Since the conclusions are consistent with any model, we just show the experiment results on ResNet-18 below as an instance.
    
\begin{figure*}[t]
    \centering \includegraphics[width=\textwidth]{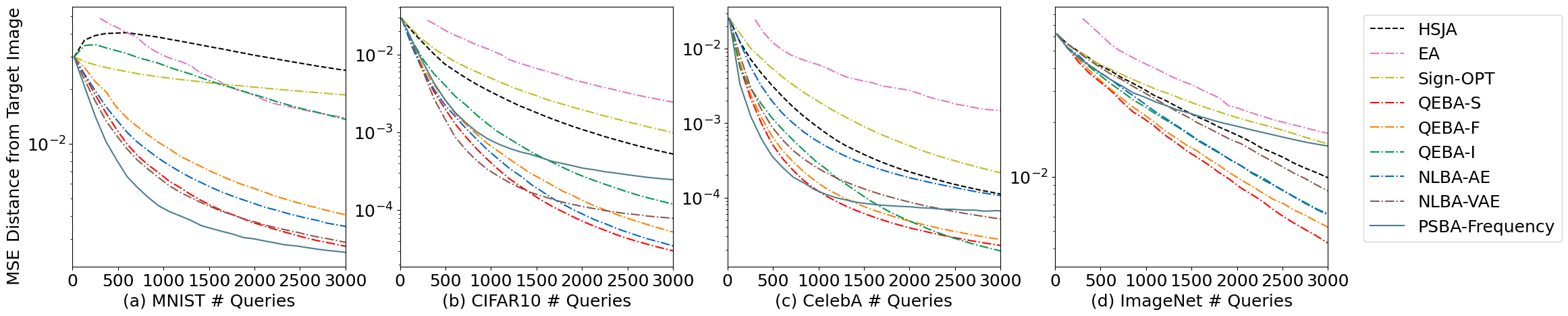}
    \vspace{-5mm}
    \caption{\small The perturbation magnitude w.r.t. different number of queries for different methods. The PSBA here is applied on frequency domain.}
    \label{fig:fre_mean}
    \end{figure*}

\begin{figure*}[t]
    \centering \includegraphics[width=\textwidth]{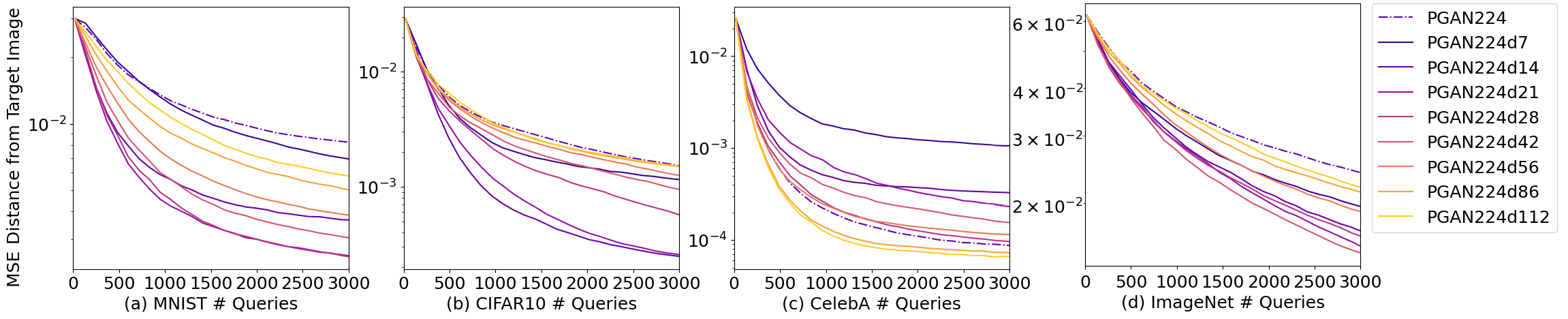}
    \caption{\small The perturbation magnitude w.r.t. different number of queries for different scales chosen on frequency domain.}
    \label{fig:fre_mean_cmp}
    \end{figure*}

\begin{figure*}[t]    
    \centering
    \includegraphics[width=\textwidth]{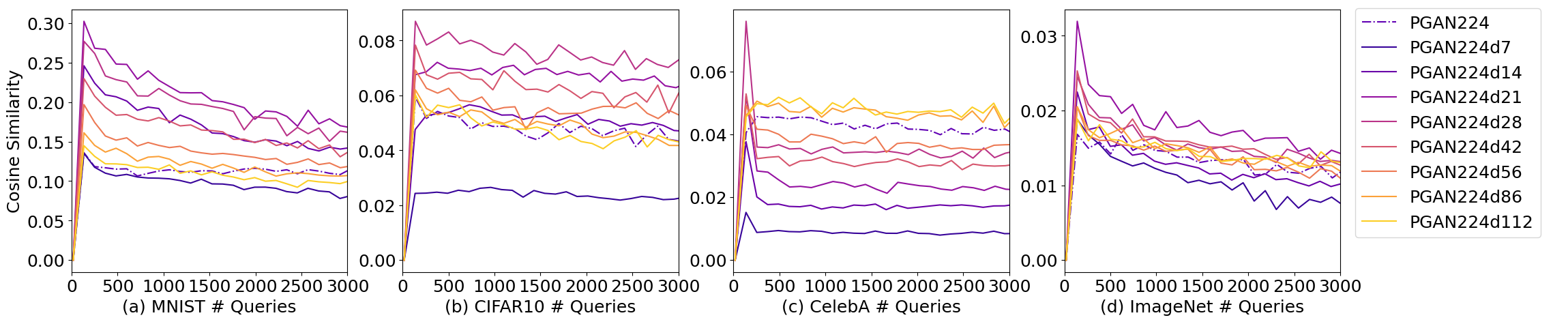}
    \caption{\small The cosine similarity between the estimated and true gradients for different scales chosen on frequency domain.}
    \label{fig:freq_cos_sim_cmp}
    \end{figure*}

\begin{figure}[htbp]
    \centering
    \begin{minipage}[t]{0.45\textwidth}
    \centering
    \includegraphics[width=8cm]{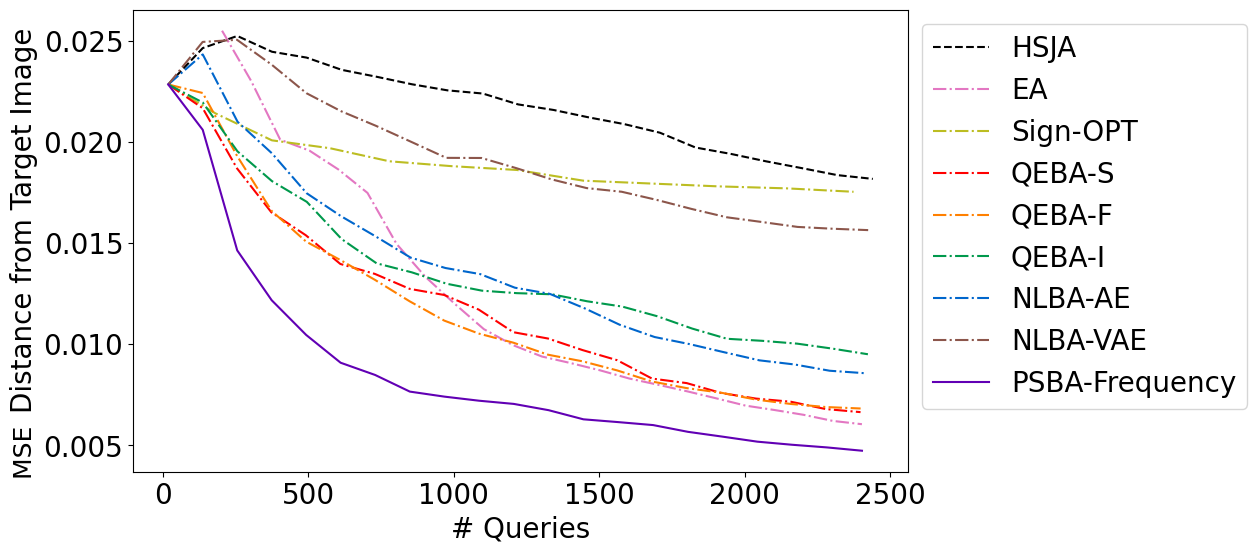}
    \vspace{-5mm}
    \caption{\small The perturbation magnitude w.r.t. different queries against Face++ `Compare' API, the PSBA here is applied on frequency domain.}
    \label{fig:freq_api_means}
    \end{minipage}
    \hspace{0.05\textwidth}
    \begin{minipage}[t]{0.45\textwidth}
    \centering
    \includegraphics[width=8cm]{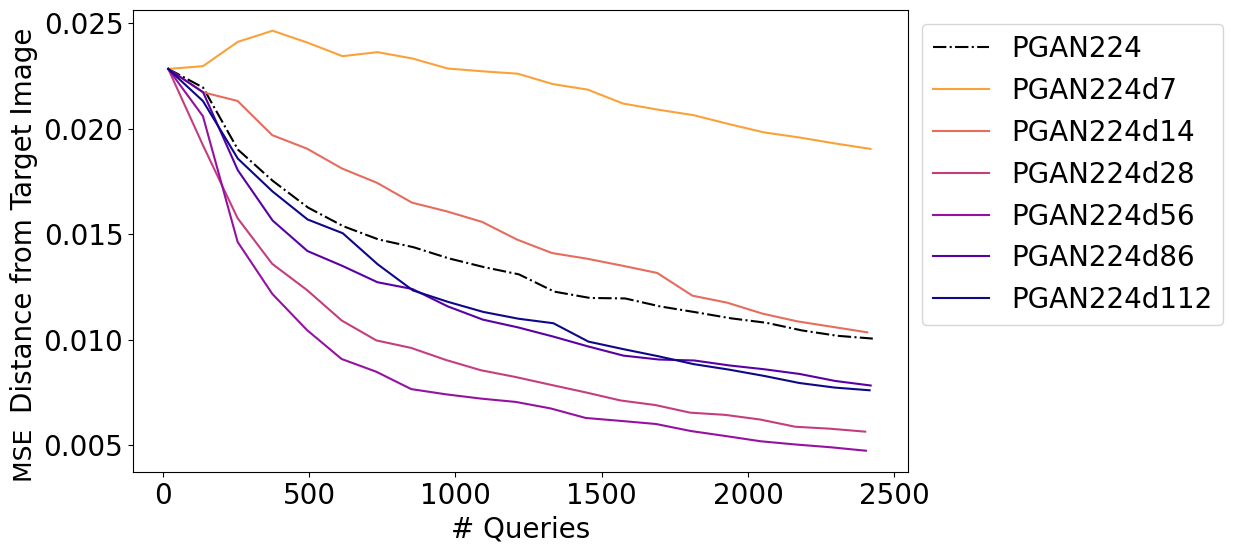}
    \vspace{-5mm}
    \caption{\small The perturbation magnitude w.r.t. different number of queries for different scales chosen on frequency domain against Face++ `Compare' API.}
    \label{fig:freq_api_means_cmp}
    \end{minipage}
    \end{figure}

    \paragraph{Frequency Domain.}
    As discussed in \Cref{sec:long_tailed_distribution}, with applying the DCT transformation on the output of PGAN in $224\times224$ scale, the low-frequency components will be concentrated at the upper left corner, i.e., low-frequency subspace.
    Then, we let `PGAN224dk' denote the adjusted attack process where we just save the $k\times k$ signals on the upper left corner of the frequency representation of the output of PGAN224 and transform it back to the original space by the Inverse DCT to continue the attack. In other words, we just use some low-frequency part of the original gradient images generated from PGAN224 to estimate the ground gradient and the pseudocode is also provided in~\Cref{alg:dct_transformation} for making it more clear. The final attack performance is shown in~\Cref{fig:fre_mean}, and as we can see, in some cases like when the src-tgt images are sampled from MNIST, \ourapproach-freq even outperforms all other baselines with a simple adjustment on the output of the inherent bad gradient estimator PGAN224. The results corresponding to different choices of frequency region and their induced changes of cosine similarity are shown in~\Cref{fig:fre_mean_cmp} and~\Cref{fig:freq_cos_sim_cmp}. Besides, this simple strategy can also work well on the attack to the online API, which is shown in~\Cref{fig:freq_api_means} and~\Cref{fig:freq_api_means_cmp}. 
    
    As we can see, even this simple ``gating'' strategy can improve the attack performance a lot compared with original PGAN224. Though it is not as competitive as the progressive scaling in the spatial domain due to lack of projection model finetuning.
    Furthermore, the existence of optimal scale is pronounced.
    
\begin{figure*}[h]
    \centering \includegraphics[width=\textwidth]{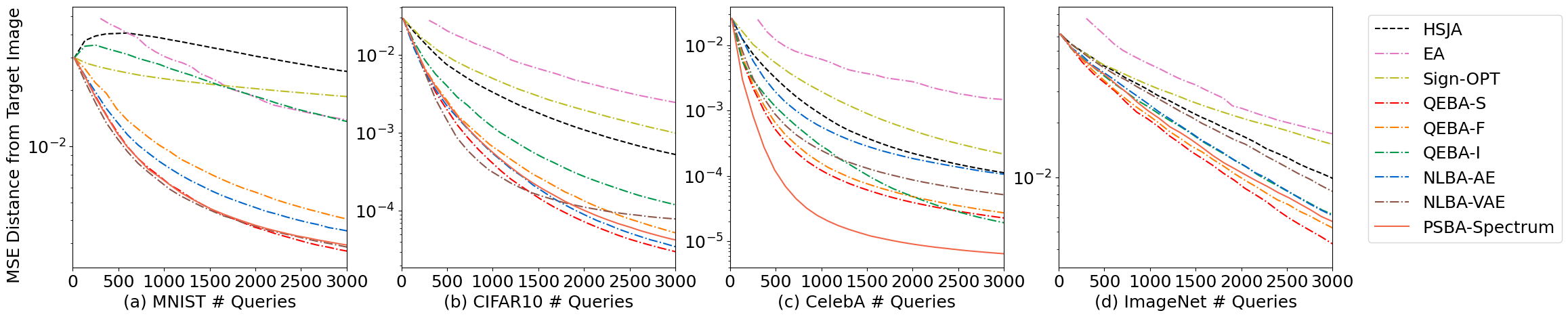}
    \caption{\small The perturbation magnitude w.r.t. different number of queries for different scales chosen on spectrum domain.}
    \label{fig:spectrum_mean}
    \end{figure*}

\begin{figure*}[h]
    \centering \includegraphics[width=\textwidth]{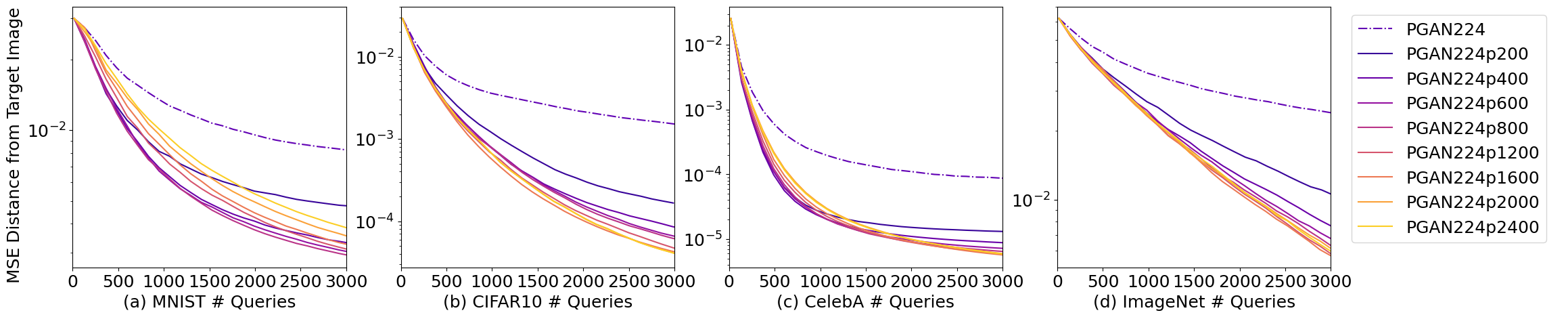}
    \caption{\small The perturbation magnitude w.r.t. different number of queries for different methods, the PSBA here is applied on spectrum domain.}
    \label{fig:spectrum_mean_cmp}
    \end{figure*}

\begin{figure*}[h]    
    \centering
    \includegraphics[width=\textwidth]{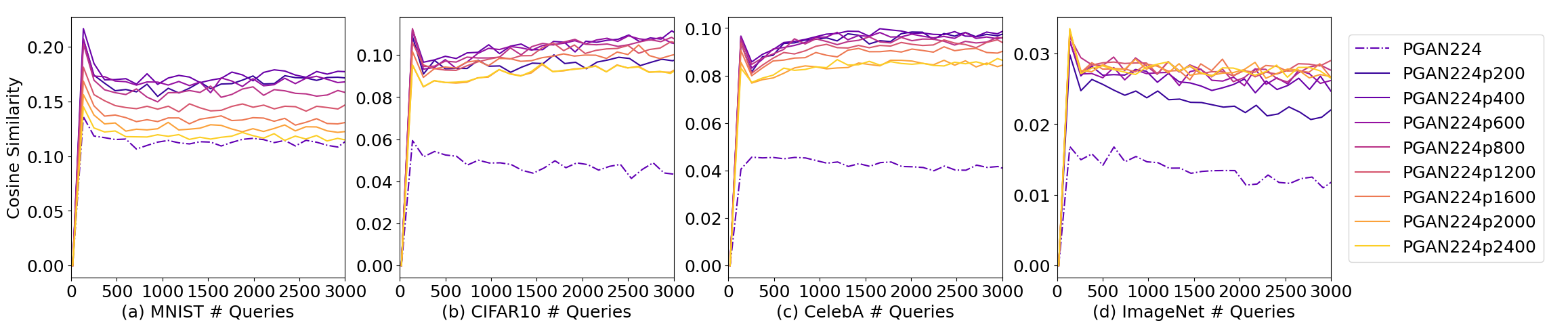}
    \caption{\small The cosine similarity between the estimated and true gradients for different scales chosen on spectrum domain.}
    \label{fig:spectrum_cos_sim_cmp}
    \end{figure*}

    \paragraph{Spectrum Domain.} 
    Here, we sample 40,000 gradient images generated from PGAN224, and then use PCA to decompose them to get $9,408$ main components. It may seem to be great if we project the original gradient images generated from PGAN224 to just part of the main components, however, the computation cost is a little unacceptable, since there are a lot of dot product operations between two $150,000$-dimensional vectors required. Therefore, for efficiency, the gradient images generated here are actually composed by the combination of the top-$k$ main components among the total $9,408$ components with the coefficients sampled from normal distribution. Thus, for simplicity, we denote `PGAN224pk' as the attack with the combination of the top-$k$ main components decomposed by PCA. By progressively increase the value of $k$, the attack performance are shown in~\Cref{fig:spectrum_mean}. The result on different spectrum scales and corresponding changes of cosine similarity are shown in~\Cref{fig:spectrum_mean_cmp} and~\Cref{fig:spectrum_cos_sim_cmp}.
    Again, we observe an apparent improvement over the original PGAN224 and the existence of the optimal scale.
    
    \subsection{Optimal Scale across Different Model Structures}
    \label{sec:optimal_scale_dif_model}
    One may think that the optimal scale may be influenced a lot by the specific structure of the target model. In other words, the depth of the model, the existence of the residual connection and batchnorm layer, and so on, would affect the optimal scale. However, the result as shown in~\Cref{tab:optimal_scale_dif_model} demonstrates that the optimal scale is actually stable, which is usually the small scale $28\times28$.
    
    It is our future work to look into this phenomenon and analyze the optimal properties such as stability of optimal scale.
    Besides, together with the improved attack performance owing to the removing of the high frequency part or the focus on the most informative spectrum of the generated images of PGAN224, it will be promising to explore the benefits brought from gradient sparsification and devise a more efficient algorithm in the future.

\section{Qualitative Results} \label{adxsec:qualitative}
In this section, we present the qualitative results for attacking both offline models and online APIs.
\subsection{Offline Models}
\label{sec:offline_qualitative}
\begin{figure}[t]
    \centering
    \includegraphics[height=0.95\textheight]{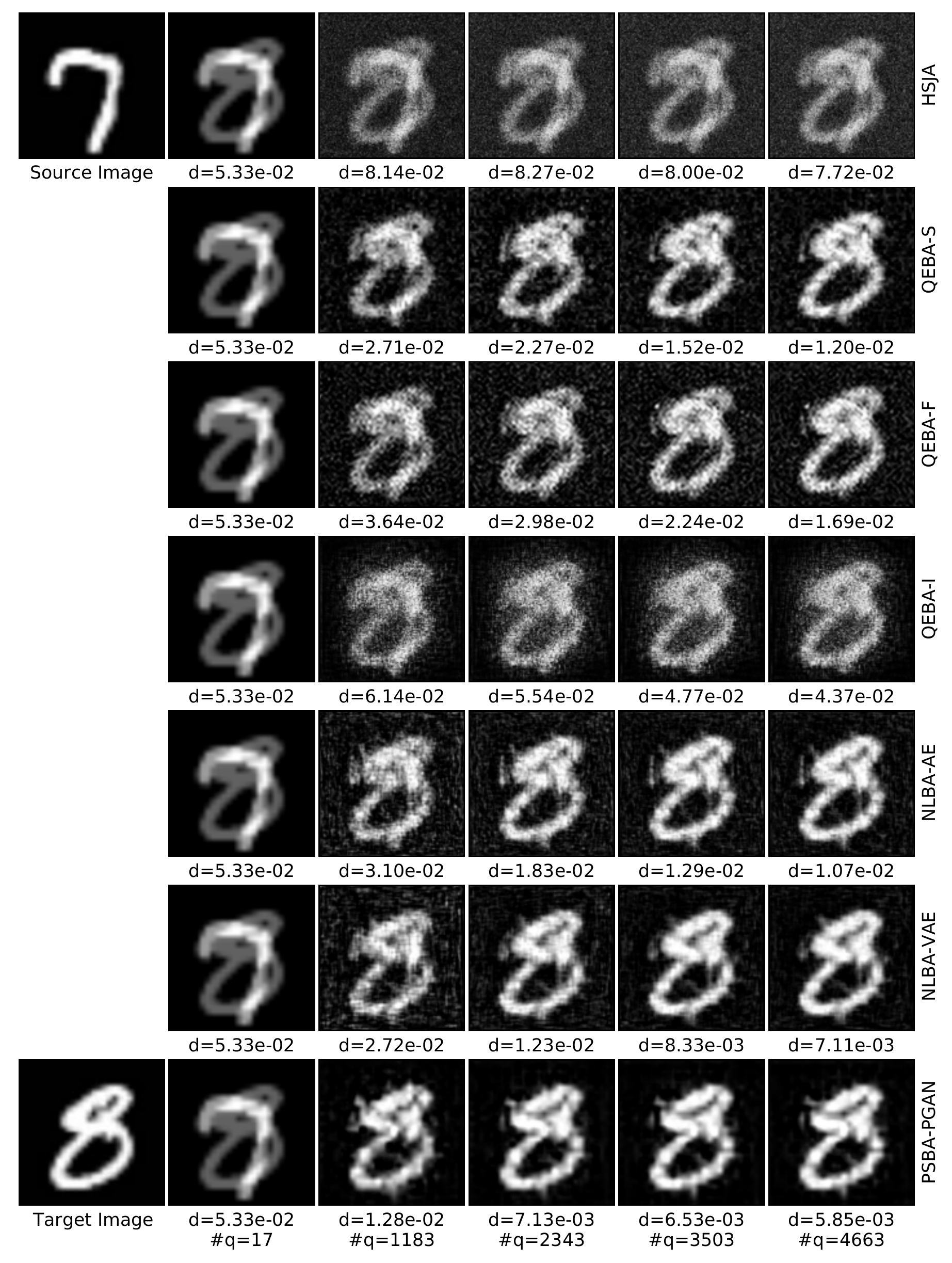}
    \caption{The qualitative case study of attacking ResNet-18 model on MNIST dataset.}
    \label{fig:mnist_attack_process}
\end{figure}

\begin{figure}[ht]
    \centering
    \includegraphics[height=0.95\textheight]{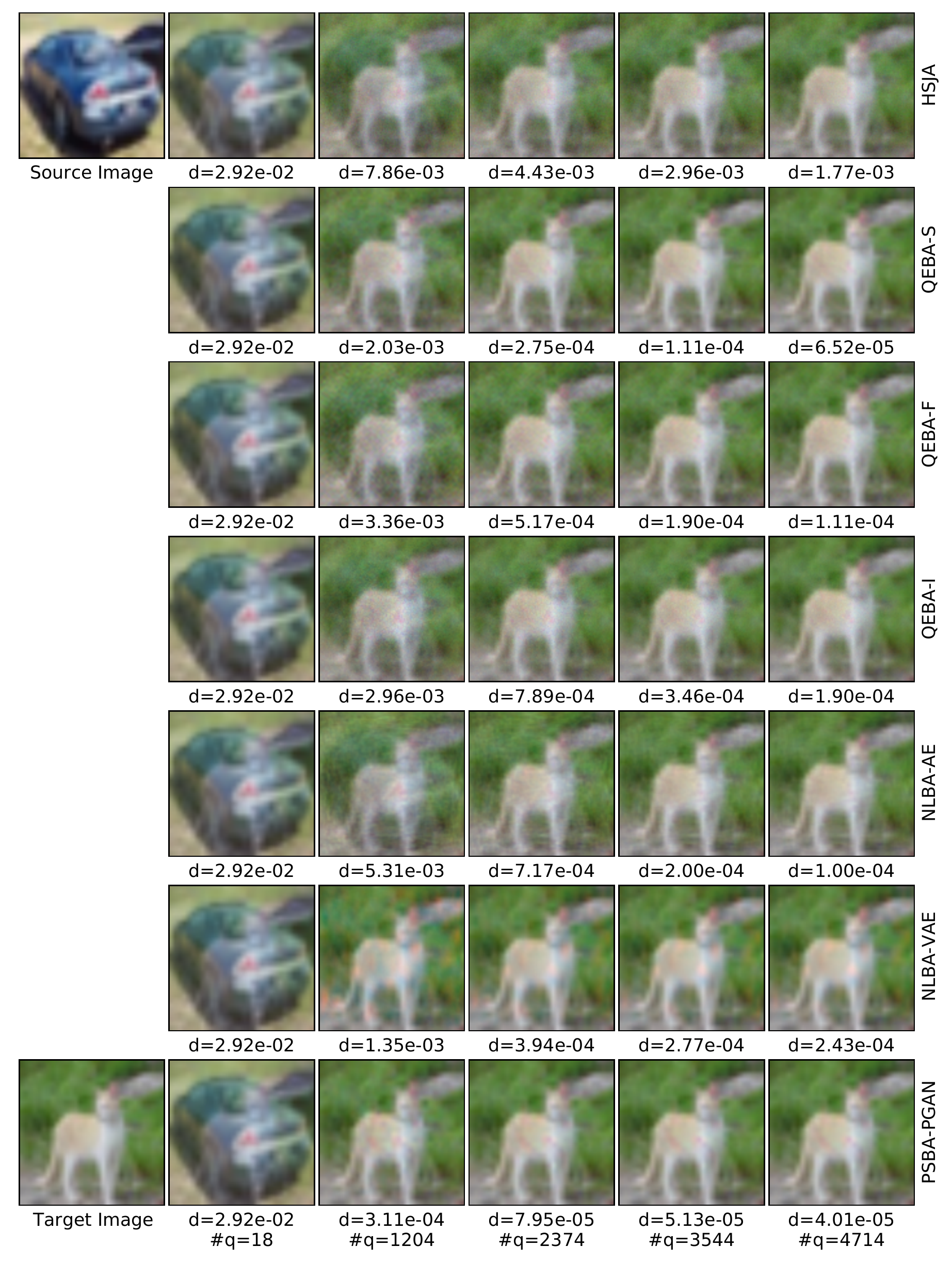}
    \caption{The qualitative case study of attacking ResNet-18 model on CIFAR-10 dataset.}
    \label{fig:cifar10_attack_process}
\end{figure}

\begin{figure}[t]
    \centering
    \includegraphics[height=0.95\textheight]{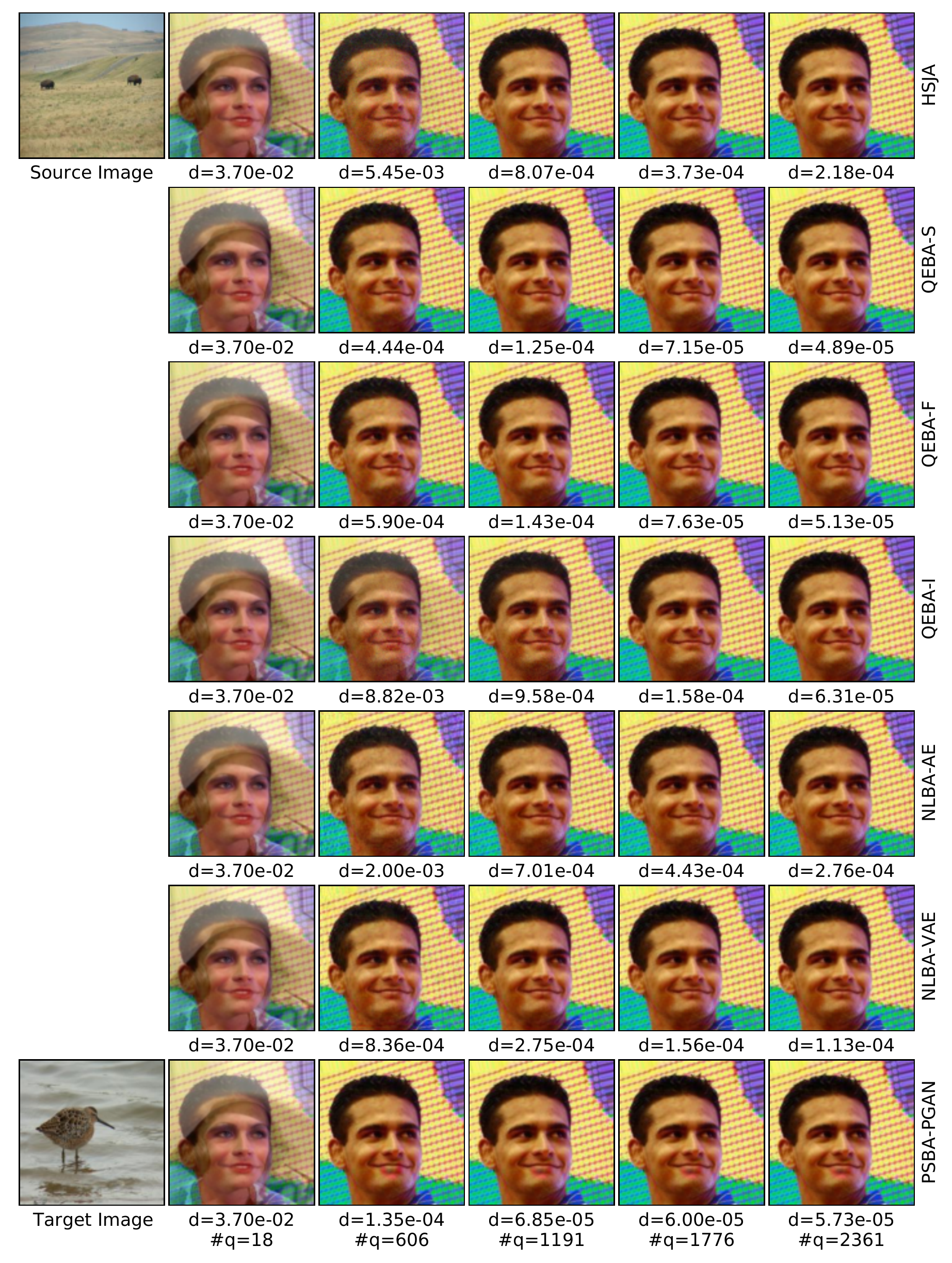}
    \caption{The qualitative case study of attacking ResNet-18 model on CelebA dataset.}
    \label{fig:celeba_attack_process}
\end{figure}

\begin{figure}[t]
    \centering
    \includegraphics[height=0.95\textheight]{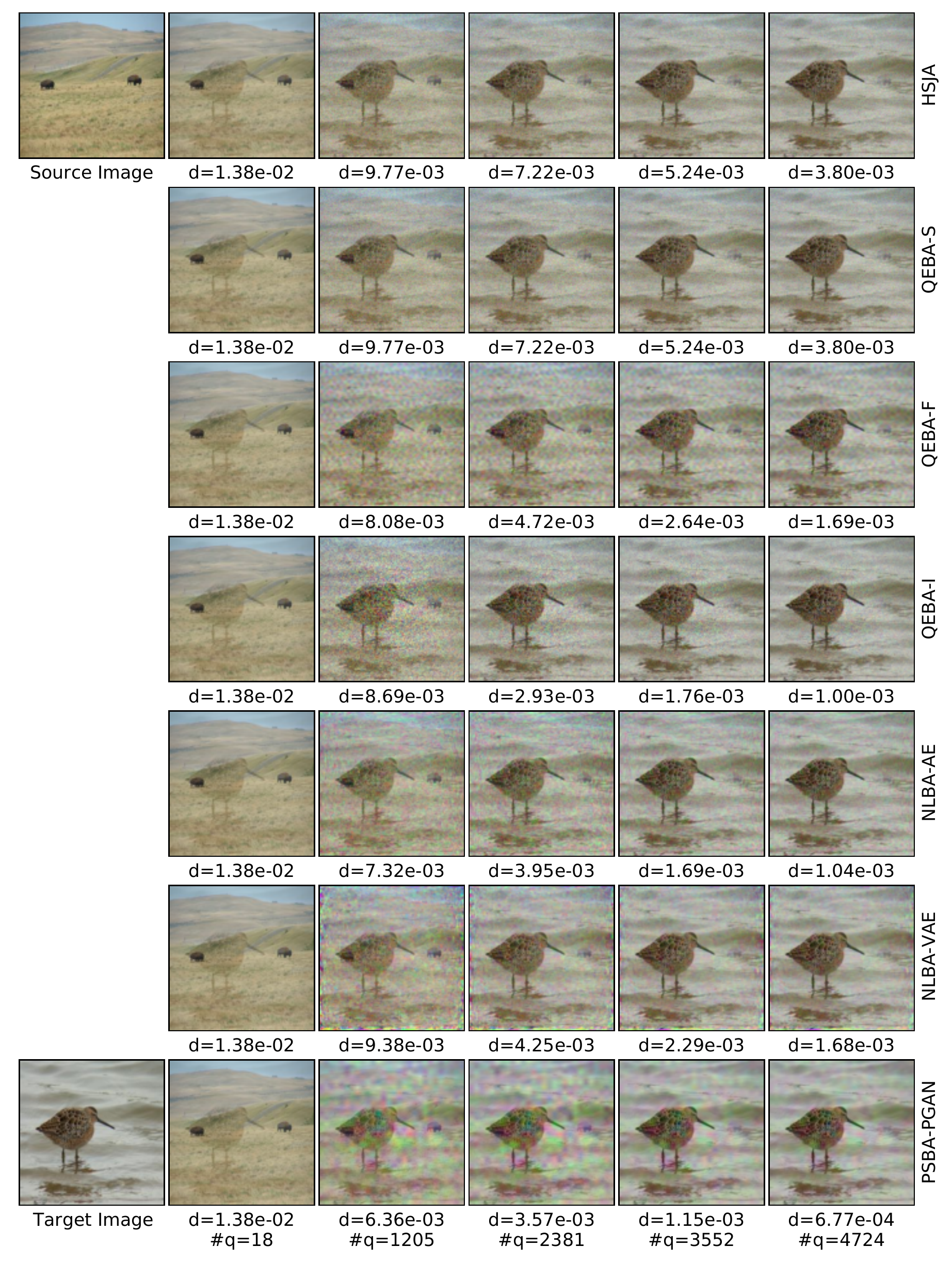}
    \caption{The qualitative case study of attacking ResNet-18 model on ImageNet dataset.}
    \label{fig:imagenet_attack_process}
\end{figure}

The goal of the attack is to generate an adversarial image that looks like the target image but has the same label with source image. 
We report qualitative results that show how the adversarial image changes during the attack process in Figure~\ref{fig:mnist_attack_process}, Figure~\ref{fig:cifar10_attack_process}, Figure~\ref{fig:celeba_attack_process}, and Figure~\ref{fig:imagenet_attack_process} for the four datasets respectively. The target model chosen here is ResNet-18. In the figures, the left-most column has two images: the source image and the target image. They are randomly sampled from the corresponding dataset. We make sure that the images in the sampled pairs have different ground truth labels (otherwise the attack is trivial).
The other five columns each represents the adversarial image at certain number of queries as indicated by $\#q$ at the bottom line. In other words, all images in these five columns can successfully attack the target model. Each row represents one method as shown on the right. The $d$ value under each image shows the MSE between the adversarial image and the target image. The smaller the $d$ is, the better the attack is.

\subsection{Commercial Online API Attack}
\begin{figure}[t]
    \centering
    \includegraphics[height=0.93\textheight]{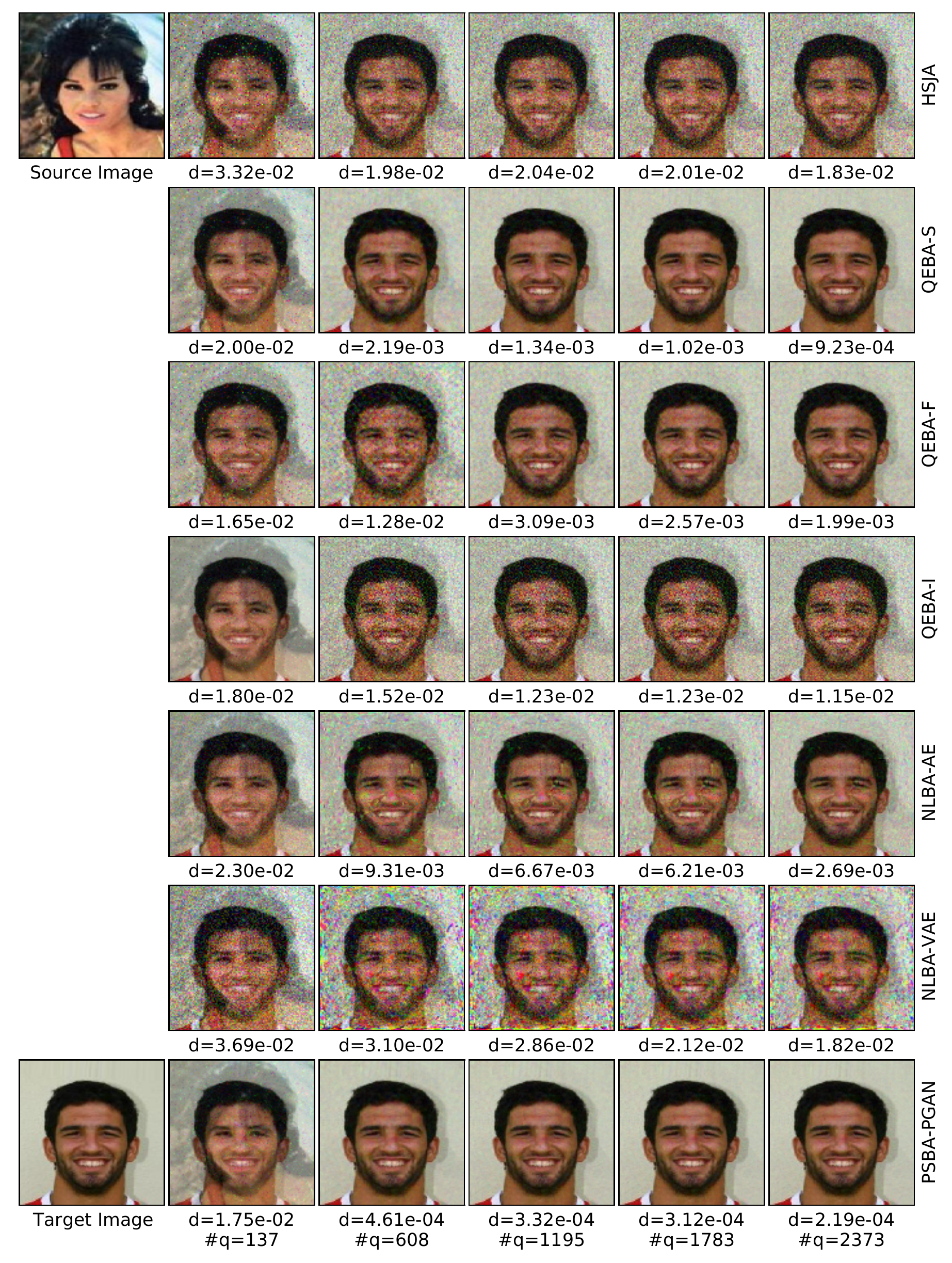}
    \caption{A case study of Face++ online API attack process. The source-target image pair is randomly sampled from CelebA dataset (ID: 019862 and 168859).}
    \label{fig:facepp_attack_process}
\end{figure}

As discussed in Section~\ref{sec:exp}, the goal is to generate an adversarial image that looks like the target image but is predicted as `same person' with the source image. 
In this case, we want to get images that looks like the man but is actually identified as the woman.
The qualitative results of attacking the online API Face++ `compare' is shown in Figure~\ref{fig:facepp_attack_process}.
In the figure, the source image and target image are shown on the left-most column.   

    \begin{table}[!h]
    \centering
    \caption{The optimal scale across different model structures.}
    \setlength{\tabcolsep}{6mm}{
    \begin{tabular}{|c|c|c|c|c|}
    \hline
    \diagbox{Model}{Dataset} & MNIST & CIFAR-10 & CelebA & Imagenet \\
    \hline
        ResNet-18 & $28\times28$ & $28\times28$ & $28\times28$ & $28\times28$ \\
        \hline
        ResNet-34 & $28\times28$ & $28\times28$ & $28\times28$ & $56\times56$ \\
        \hline
        ResNet-152 & $28\times28$ & $28\times28$ & $56\times56$ & $56\times56$ \\
        \hline
        ResNext50\_32x4d & $28\times28$ & $28\times28$ & $28\times28$ & $56\times56$ \\
        \hline
        Vgg11 & $28\times28$ & $28\times28$ & $28\times28$ & $56\times56$ \\
        \hline
        Vgg19 & $28\times28$ & $28\times28$ & $28\times28$ & $56\times56$ \\
        \hline
        Vgg11\_bn & $112\times112$ & $28\times28$ & $28\times28$ & $56\times56$ \\
        \hline
        Vgg19\_bn & $112\times112$ & $28\times28$ & $28\times28$ & $56\times56$ \\
        \hline
        DenseNet161 & $28\times28$ & $14\times14$ & $28\times28$ & $56\times56$ \\
        \hline
    \end{tabular}}
    \label{tab:optimal_scale_dif_model}
    \end{table} 
    
\end{document}